\PassOptionsToPackage{shortlabels}{enumitem}
\PassOptionsToPackage{table,xcdraw,dvipsnames}{xcolor}

\documentclass[11pt,a4paper]{include/gdm_format}

\usepackage[authoryear,sort&compress,round]{natbib}
\usepackage{graphicx}
\usepackage{subcaption}
\usepackage{adjustbox}
\usepackage{multirow}
\usepackage{amsmath,amssymb,amsfonts}
\usepackage{amsthm}
\usepackage{mathrsfs}
\usepackage{appendix}
\usepackage{textcomp}
\usepackage{booktabs}
\usepackage{algorithm}
\usepackage{algorithmicx}
\usepackage{algpseudocode}
\usepackage{listings}
\usepackage{enumitem}
\usepackage{xspace}
\usepackage{cleveref}
\usepackage[most]{tcolorbox}
\usepackage{fvextra}
\usepackage{titletoc}
\usepackage{tabularx}
\usepackage{makecell}
\usepackage{tablefootnote}
\usepackage{pdfpages}
\usepackage[normalem]{ulem}
\usepackage{xurl}
\usepackage[parfill]{parskip}

\usepackage{amsmath,amsfonts,bm}

\def\eqref#1{equation~\ref{#1}}

\def\1{\bm{1}}

\DeclareMathAlphabet{\mathsfit}{\encodingdefault}{\sfdefault}{m}{sl}
\SetMathAlphabet{\mathsfit}{bold}{\encodingdefault}{\sfdefault}{bx}{n}

\newcommand{\ouralgo}{The AI Scientist\xspace}

\newcommand{\ourreviewer}{Automated Reviewer\xspace}

\DefineVerbatimEnvironment{mintedfallback}{Verbatim}{
  breaklines,
  breaksymbol={},
  breaksymbolleft={},
  fontsize=\scriptsize
}
\newenvironment{minted}[1]{\mintedfallback}{\endmintedfallback}

\title{Towards End-to-End Automation of AI Research}

\correspondingauthor{Yutaro Yamada (yutaro.yamada@sakana.ai), Robert Tjarko Lange (robert@sakana.ai), Cong Lu (conglu97@outlook.com), Chris Lu (luchris429@gmail.com), David Ha (hadavid@sakana.ai), and Jeff Clune (jclune@gmail.com).}

\author[1,*]{Yutaro Yamada}
\author[1,*]{Robert Tjarko Lange}
\author[1,3,4,*]{Cong Lu}
\author[1,2,*]{Chris Lu}
\author[1,3,4]{Shengran Hu}
\author[2]{Jakob Foerster}
\author[1]{David Ha}
\author[3,4]{Jeff Clune}

\affil[*]{Equal contribution.}
\affil[1]{Sakana AI, Tokyo, Japan}
\affil[2]{FLAIR, University of Oxford, Oxford, UK}
\affil[3]{University of British Columbia, Vancouver, Canada}
\affil[4]{Vector Institute, Toronto, Canada}

\begin{document}

\begin{abstract}
The automation of science is a long-standing ambition in the field of AI~\citep{lenat1977automated,buchanan1981dendral}. While the community has made significant progress in automating individual components of the scientific process, a system that autonomously navigates the entire research lifecycle---from conception to publication---has remained out of reach. Here, we present the strongest demonstration to date toward automating the entire process end-to-end. We present \ouralgo, which creates research ideas, writes code, runs experiments, plots and analyzes data, writes the entire scientific manuscript and performs its own peer review. Its ideas, execution, and presentation are of sufficient quality to produce a manuscript generated by an AI system that passes the first round of peer review at a major machine learning conference workshop. The workshop has an acceptance rate of 70 percent. Our system leverages modern foundation models~\citep{gpt4,claude3,llama3} within a complex agentic system. We evaluate \ouralgo in two settings: a focused mode using human-provided code templates as an initial scaffold to conduct research on a specific topic, and a template-free, open-ended mode that leverages agentic search for wider scientific exploration~\citep{aide2025,chan2025mlebench}. Both settings produce diverse ideas and automatically test, report on, and evaluate them. This achievement demonstrates AI's growing capacity for scientific contribution and signifies a potential paradigm shift in how research is conducted. As with any impactful new technology, there could be significant risks, including taxing overwhelmed review systems and adding noise to scientific literature. However, if developed responsibly, such autonomous systems could greatly accelerate scientific discovery.
\end{abstract}

\maketitle

\section{Introduction}

AI has long been used to aid scientific discovery, an ambition with deep roots in the history of the field~\citep{waltz2009automating,langley2024integrated,langley1987scientific,lenat1977automated,lenat1984and}.
Prior to the rise of large language models (LLMs), AI was limited to helping with specific, narrow tasks, such as discovering chemical structures~\citep{buchanan1981dendral}, finding mathematical proofs~\citep{lenat1977automated}, discovering novel materials~\citep{pyzer2022alphamaterials,merchant2023scaling,szymanski2023autonomous}, and predicting the 3D shape of proteins~\citep{jumper2021alphafold, hayes2024simulating}.
Other systems focused on analyzing pre-collected datasets to find novel insights~\citep{langley1987scientific,ifargan2024autonomousllmdrivenresearchdata, falkenhainer1986integrating}.
However, with the recent advent of powerful and general foundation models, AI's role has expanded to assist with a wider array of research activities.
For example, LLMs now help with generating novel hypotheses~\citep{girotra2023ideas,lehman2022evolutionlargemodels,lu2025automated,omniepic,hu2025automated}, writing literature reviews~\citep{baek2024researchagentiterativeresearchidea,wang2024autosurveylargelanguagemodels}, and coding experiments~\citep{huang2024mlagentbench,lu2024discovering,ma2023eureka,zhang2025darwin}.
Despite these advances in automating individual components, a system that autonomously navigates the entire research lifecycle---from conception to publication---has remained out of reach until now.

This paper introduces \ouralgo, the first pipeline to achieve this vision of the full end-to-end automation of that scientific process.
\ouralgo uses existing foundation models to perform ideation, literature search, experiment planning and implementation, result analysis, manuscript writing, and peer review to produce complete, novel papers.
We focus on machine learning science, where experiments typically occur entirely on the computer.

A central challenge in developing such a system is automatically evaluating the quality of its scientific output at scale.
To address this, we created an \ourreviewer and first evaluated its performance against real, human-generated papers.
The \ourreviewer can accurately predict conference acceptance decisions, performing on par with human reviewers (Supplementary \Cref{app:reviewer_details}).
We then used the \ourreviewer to compare various configurations of \ouralgo, assessing how performance changes with the scale of test-time compute and the quality of the underlying foundation model.
We find that \ouralgo performs better with additional compute resources (\Cref{fig:tree_ablations}C).
Furthermore, the \ourreviewer shows that improvements to the base models significantly improve the quality of the generated papers, a finding which strongly implies that future versions of our system will be substantially more capable, as models continue to improve (\Cref{fig:conceptual}B).

To assess \ouralgo in the same setting in which human-authored papers are evaluated, we conducted an experiment where we submitted generated papers to a workshop at the International Conference on Learning Representations (ICLR) with the organizer's consent.
In computer science, such top-tier conferences are the primary and most prestigious venues for archival and rigorously peer-reviewed publication.
They also have workshops with a substantially lower, but still non-trivial bar for peer-reviewed acceptances.
One manuscript achieved high enough scores to exceed the average human acceptance threshold at a workshop, marking the first instance of a fully AI-generated paper successfully navigating a peer review process, albeit one with a lower bar.

\section{Generating Manuscripts}

\begin{figure}[tbp]
\centering
\includegraphics[width=0.995\textwidth, trim={40 50 10 70},clip]{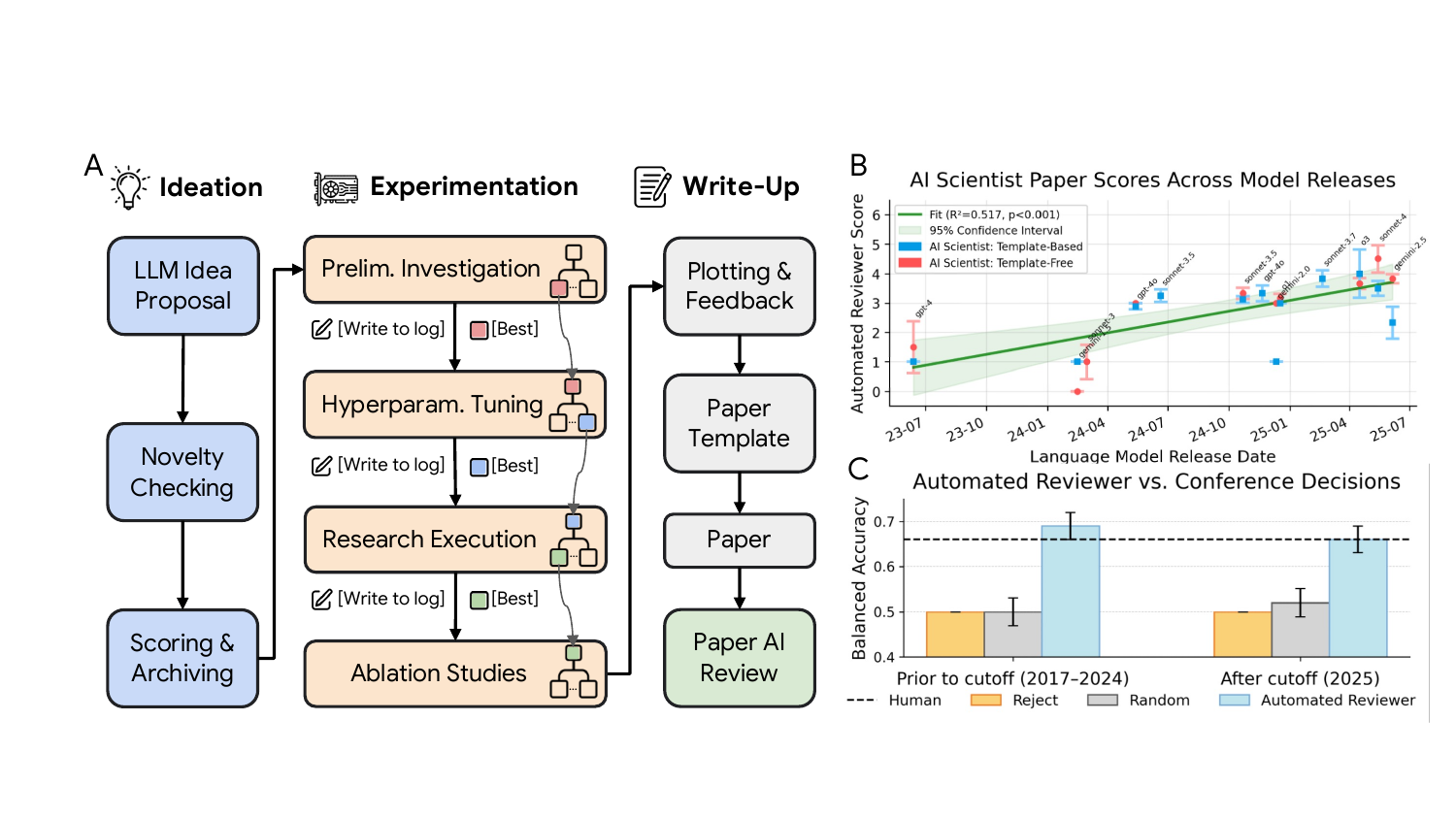}
\caption{
\textbf{\ouralgo Workflow}.
\textbf{A}. \ouralgo consists of distinct phases covering automated idea generation, tree-based experimentation, manuscript writing, and reviewing.
The experimentation phase employs agentic tree search to generate and refine code implementations.
This is structured into four stages: (1) initial investigation, (2) hyperparameter tuning, (3) research agenda execution, and (4) ablation studies.
From one experimentation stage to the next, the best-performing checkpoint is selected to seed the next stage of tree search.
\textbf{B}. Paper quality consistently improves with the underlying model release date (as judged by the \ourreviewer), suggesting consistent future improvements with improving foundation models. The observed correlation is statistically significant ($p$-value $<0.00001$).
Shaded regions represent the standard error.
Full experimental details, including model versions and replication counts, are provided in Supplementary \Cref{app:v2_hyperparams}.
\textbf{C}. The \ourreviewer achieves performance comparable to human reviewers validated by openly available decisions from past conferences (\Cref{tab:reviewers_main_paper}).
Error bars represent the 95\% bootstrapped confidence intervals.
}
\label{fig:conceptual}
\end{figure}

\ouralgo sequentially completes four main phases (\Cref{fig:conceptual}A): In the first phase, \ouralgo is prompted to iteratively grow an archive~\citep{Mouret2015IlluminatingSS} of high-level research directions and hypotheses it can explore within a user-specified machine learning research subfield (an example progression is visualized in Supplementary \Cref{app:idea_progression}).
For each direction, it generates a descriptive title, explanation of its reasoning for what the idea is and why it's interesting to pursue it, and a proposed experimental plan (Supplementary \Cref{app:v1_prompts_scientist,app:v2_prompts_scientist}).
After idea generation, \ouralgo filters ideas by connecting the language model to the Semantic Scholar API~\citep{fricke2018semantic} and web access as tools~\citep{schick2024toolformer}.
This allows \ouralgo to discard any idea that is too similar to existing literature.
The second phase of \ouralgo executes the proposed experiments and then visualizes their results for the downstream write-up.
We tested two different variants of experiment execution: (1) \underline{Template-Based}: \ouralgo is provided with a starting code template that reproduces a training run from a popular algorithm.
\ouralgo then executes the proposed experiment plan in linear order (Supplementary \Cref{app:v1_details}).
(2) \underline{Template-Free}: Alternatively, \ouralgo can generate an initial starting code script by itself.
In this case, experimentation includes additional stages for optimizing the code it writes from scratch, and experiment execution leverages additional test-time compute with tree search (see methods).
After each experiment, \ouralgo is given the results and is prompted to take notes in the style of an experimental journal for future planning and writeup.
The third phase of \ouralgo produces a concise write-up of its research in the style of a standard machine learning conference paper.
\ouralgo is prompted to fill in a blank LaTeX conference template section by section, using its notes and plots (see methods).
To construct the related work section and add citations throughout the manuscript, the system queries the Semantic Scholar \citep{fricke2018semantic} API for relevant literature, comparing its findings against the generated manuscript over 20 rounds.
For each potential citation, the system generates a textual justification for its inclusion, which informs \ouralgo on how to use the reference appropriately within the manuscript.
Finally, the paper generated by \ouralgo undergoes a review by the \ourreviewer to automatically evaluate the scientific quality of the conducted research.

\section{Automated Evaluation of Generated Papers}

The \ourreviewer provides reviews based on the top-tier Neural Information Processing Systems (NeurIPS) conference review guidelines~\citep{neurips2022reviewerguidelines}.
The output contains numerical scores (soundness, presentation, contribution, overall, and reviewer confidence), lists of weaknesses and strengths, as well as a binary decision (\textit{accept} or \textit{reject}).
The \ourreviewer's pipeline consists of an ensemble of five reviews, followed by a meta-review where the model acts as an Area Chair to make a final decision conditioned on all five reviews (Supplementary \Cref{app:reviewer_details}).
We compared \ourreviewer decisions with ground truth data for ICLR papers, extracted from the publicly available OpenReview dataset~\citep{gonzalez2024learning}.
As shown in \Cref{tab:reviewers_main_paper}, the \ourreviewer's agreement with human paper assessments is comparable to inter-human agreement measured by F1 and balanced accuracy as reported in the NeurIPS 2021 consistency study~\citep{beygelzimer2021neurips}, which measured agreement between human reviewers on a comparable set of submissions (Supplementary \Cref{app:reviewer_details}).
This demonstrates its ability to replicate the collective judgment of human reviewers with high fidelity.
These results are statistically significant (non-parametric bootstrap test~\citep{efron1993bootstrap} and two-sample z-test~\citep{lehmann1959ztest}; Supplementary \Cref{app:reviewer_details}).
Next, to investigate the effect of potential data contamination (i.e., the possibility that paper decisions were part of the LLM's training set), we evaluated the \ourreviewer on two datasets: one containing 1,000 papers from years potentially within the model's training data ($2017 - 2024$), and a second ``clean'' dataset from the year after the cutoff (2025), which could not have been seen during training.
A comparison between years before and after the knowledge cutoff suggests that data contamination may exist, as balanced decision accuracy decreases from 69\% before to 66\% in the year after the cutoff.
However, the results for the year after the cutoff remain comparable to those of human reviewers (e.g., 66\% balanced accuracy), showing that potential contamination would have at most a minimal effect.

\begin{table}[tbp]
    \centering
    \caption{
    Performance comparison of human reviewers (NeurIPS 2021 consistency experiment~\citep{beygelzimer2021neurips}) and the \ourreviewer, evaluated on papers published prior to ($2017 - 2024$) and after (2025) the knowledge cutoff.
    The \ourreviewer achieves performance superior or comparable to human reviewer consistency in key metrics such as F1 Score, AUC, and Balanced Accuracy, even for data beyond the knowledge cutoff, highlighting its robustness and reliability across different time periods.
    Error margins denote the 95\% bootstrapped confidence intervals.
    Arrows indicate if it is better for a score to be higher ($\uparrow$) or lower ($\downarrow$). Supplementary \Cref{app:v1_reviewer_details} explains each metric and comparison in detail.
    \label{tab:reviewers_main_paper}
    }

    \small
    \setlength{\tabcolsep}{4pt}
    \renewcommand{\arraystretch}{1.05}
    \begin{tabular}{@{}lcccccc@{}}
    \toprule
    \bfseries Reviewer
      & \makecell{\bfseries Balanced\\\bfseries Acc.\,($\uparrow$)}
      & \makecell{\bfseries Accuracy\\\bfseries ($\uparrow$)}
      & \makecell{\bfseries F1 Score\\\bfseries ($\uparrow$)}
      & \makecell{\bfseries AUC\\\bfseries ($\uparrow$)}
      & \makecell{\bfseries FPR\\\bfseries ($\downarrow$)}
      & \makecell{\bfseries FNR\\\bfseries ($\downarrow$)} \\
    \midrule
    Human (NeurIPS)
      & 0.66 & 0.73 & 0.49 & 0.65 & 0.17 & 0.52 \\
    \midrule
    \multicolumn{7}{@{}l}{\textit{Years prior to knowledge cutoff ($2017 - 2024$)}} \\
    Random Decision
      & 0.50 & 0.54 & 0.47 & 0.52 & 0.47 & 0.43 \\
    Always Reject
      & 0.50 & 0.65 & 0.00 & 0.50 & 0.00 & 1.00 \\
    \textbf{\ourreviewer}
      & \textbf{\underline{0.69 $\pm$ 0.04}}
      & \textbf{\underline{0.65 $\pm$ 0.10}}
      & \textbf{\underline{0.62 $\pm$ 0.09}}
      & \textbf{\underline{0.69 $\pm$ 0.09}}
      & \textbf{\underline{0.45 $\pm$ 0.10}}
      & \textbf{\underline{0.17 $\pm$ 0.08}} \\
    \midrule
    \multicolumn{7}{@{}l}{\textit{Year after knowledge cutoff (2025)}} \\
    Random Decision
      & 0.52 & 0.51 & 0.48 & 0.49 & 0.50 & 0.48 \\
    Always Reject
      & 0.50 & 0.56 & 0.00 & 0.50 & 0.00 & 1.00 \\
    \textbf{\ourreviewer}
      & \textbf{\underline{0.66 $\pm$ 0.03}}
      & \textbf{\underline{0.63 $\pm$ 0.09}}
      & \textbf{\underline{0.67 $\pm$ 0.09}}
      & \textbf{\underline{0.65 $\pm$ 0.10}}
      & \textbf{\underline{0.52 $\pm$ 0.10}}
      & \textbf{\underline{0.17 $\pm$ 0.07}} \\
    \bottomrule
    \end{tabular}
\end{table}

\begin{figure}[!t]
\centering
\includegraphics[width=0.93\textwidth]{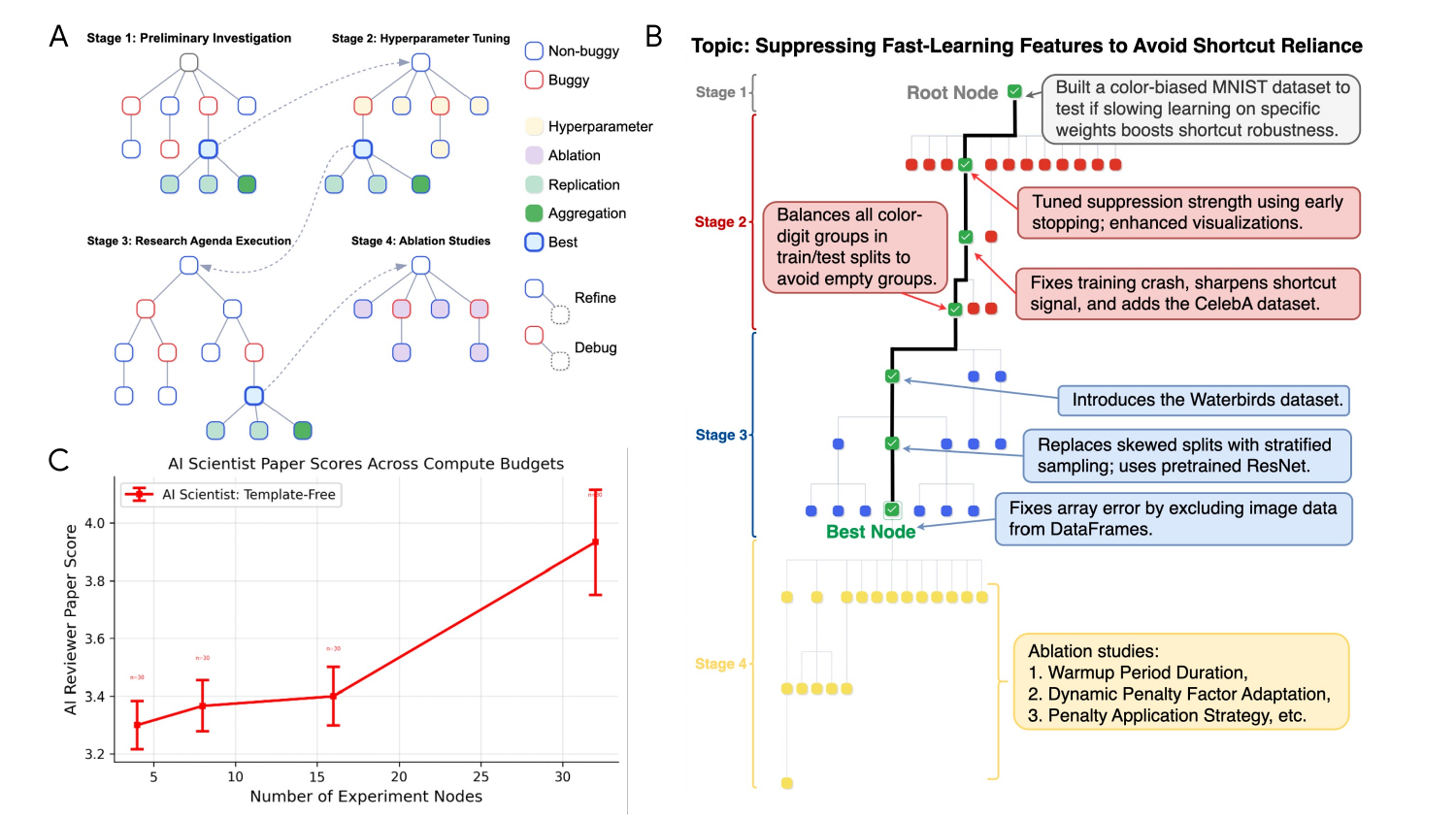}
\caption{
\textbf{A}. The research experimentation phase is visualized as a four-stage process.
A preliminary baseline code implementation is first constructed (Stage 1) and refined by tuning hyperparameters (Stage 2).
The resultant code serves as a starting point for executing the research agenda via agentic tree search (Stage 3), followed by ablation experiments (Stage 4).
Full details of the agentic tree search process are provided in the methods.
\textbf{B}. A real example of \ouralgo's tree search with node annotations outlining the conducted experiments across the four different stages.
\textbf{C}. Scaling the number of experimentation nodes in the agentic tree search shows improvements for deeper test-time search budgets.
Error bars represent the standard error.
Full experimental details are provided in Supplementary \Cref{app:v2_hyperparams}.
}
\label{fig:tree_ablations}
\end{figure}

Using the \ourreviewer, we assessed the quality of the research papers generated by a wide range of LLMs as the core model within \ouralgo.
Our analysis reveals a clear trend: as models improve over time, the quality of the papers produced by \ouralgo increases correspondingly (\Cref{fig:conceptual}B).
With recent generations of models, on average, \ouralgo produces papers that approach borderline acceptability for machine learning conference workshops, as judged by our \ourreviewer (Supplementary \Cref{fig:app_scores_ai_papers}).
Additionally, there is a strong correlation between the amount of compute allocated per paper and the resulting quality (\Cref{fig:tree_ablations}C), indicating that both model scale and inference-time investment play important roles in \ouralgo's output quality, further suggesting substantial improvements as the costs of AI systems continue to exponentially decrease and capabilities exponentially increase~\citep{index2024artificial}.

\section{Human Evaluation Results}

Perhaps the ultimate and fairest test of the quality of the work of \ouralgo is a version of what we might call an \emph{AI Scientist Turing Test}: submitting the work to the same rigorous, blind peer-review systems human science is evaluated in.
We submitted three generated manuscripts to the formal peer-review process of a workshop at a top-tier machine learning conference.
This experiment was conducted with IRB approval (Supplementary \Cref{app:limitations}) and the full cooperation of the ICLR 2025 leadership and the organizers of the ``I Can't Believe It's Not Better'' (ICBINB) workshop.
This was the only venue that we submitted to.
The template-free version of \ouralgo was readily adapted to this setting by simply prompting with the workshop's broad theme (which was investigating deep learning limitations, including where prior ideas to improve it had not worked).
The overall process was then run to generate ideas, experiments, and papers. We manually filtered the most promising outputs at each stage (Supplementary \Cref{app:filtering}). Had this filtering not occurred, the papers under analysis would still have been produced in their final form, just along with other papers and thus at a greater total cost.
This process resulted in three complete manuscripts being selected for submission.
The selection was based on three criteria: whether the idea was aligned with the workshop topic, whether the code correctly implemented the proposed idea and ran without errors, and the correctness of the manuscript formatting (Supplementary \Cref{app:filtering}).
The entire scientific workflow for each paper, from ideation and coding to manuscript writing, was performed without any human modification.
These three submissions were included among the 43 total papers reviewed at the workshop.
Reviewers were informed that some of the submissions were AI-generated, but not which ones, ensuring a blind process.

\begin{figure}[!t]
\centering
\includegraphics[width=0.94\textwidth]{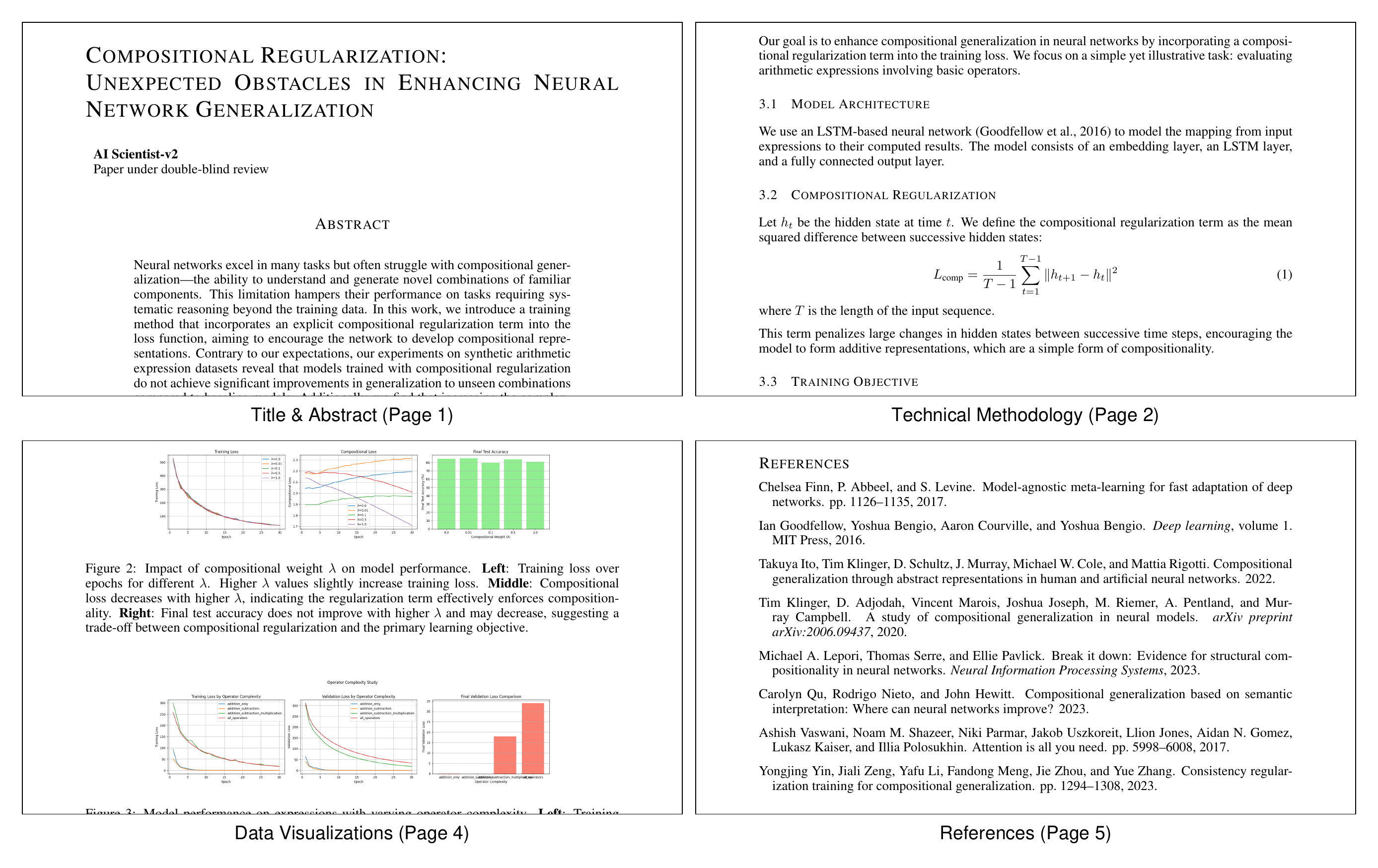}
\caption{
Selected sections from a paper generated by \ouralgo that was accepted via peer review at a top-tier machine learning conference workshop. This paper received peer review scores of 6 (weak accept), 7 (accept), and 6 (weak accept) before meta-review and ranked among the top 45\% of submitted workshop papers, marking the first instance of a fully AI-generated paper successfully navigating peer review at a top-tier conference workshop. \ouralgo wrote the entire paper, from the title through to the references and supplementary information. Shown are example sections of the paper it produced, namely portions of the abstract, methods, results (including data visualizations), and references. The full paper contains other typical paper sections, such as an introduction, a motivation, and a conclusion. The paper format follows the specific template provided by the workshop. The complete paper is available in Supplementary \Cref{app:case_study_v2}.
}
\label{fig:paper}
\end{figure}

One of the three AI-generated manuscripts received an average score of 6.33 (individual scores: 6, 7, 6) from reviewers, placing it above the workshop's average acceptance threshold (\Cref{fig:paper}).
The organizers said that the paper would have been accepted in all likelihood were it not withdrawn according to our pre-established protocol due to being AI-generated.
Notably, the accepted manuscript reported a negative result, aligning with the workshop's focus on interesting negative results.
The other two papers did not meet the bar for acceptance (Supplementary \Cref{tab:submitted_papers_v2}).
This marks the first time a fully AI-generated paper passed a standard scientific peer-review process.
We also conducted our own internal review, with human AI researchers on our team (Supplementary \Cref{app:case_study_v2}).
The team concluded that while one of the papers did meet the bar for workshop papers, none met the higher bar for a main ICLR conference publication.
A full analysis of all three submitted papers, including their strengths, weaknesses, and implementations, is provided in Supplementary \Cref{app:case_study_v2}.

\section{Limitations}

While \ouralgo generated a peer-reviewed workshop paper, there is room for improvement to match the best human-produced science.
Only one of three submissions was accepted, and workshops have much higher acceptance rates than main conferences (e.g., 70\% for the ICLR 2025 ICBINB workshop \citep{ICBINB2025} vs. 32\% for the ICLR 2025 main conference \citep{ICLR2025FactSheet}).
Therefore, \ouralgo does not yet meet the standards for top-tier publications, nor even consistently for workshops.
Common failure modes include the generation of naive or underdeveloped ideas, incorrect implementations of the main idea, a lack of deep methodological rigor, errors in experimental implementation, duplicating figures in the main text and the appendix, and many types of hallucinations such as inaccurate citations (a full analysis of failure modes is provided in Supplementary \Cref{app:case_study_v2,app:limitations,app:filtering}).
That said, often in machine learning, once something begins to work (even with clear flaws), in a few short years with scale (e.g., of compute and data), better core models, and better techniques, the capabilities of a system become surprising, and can exceed human performance levels. In assessing the impact of a technology, it is thus important to keep in mind its likely future trajectory. Crucially, this trajectory is not just about better models, but about the complexity of tasks AI systems can execute. Recent work suggests the length of tasks AI can reliably complete doubles every seven months~\citep{metr2025longtasks}, indicating that many current implementation and debugging bottlenecks may be resolved in the near term. However, some AI weaknesses have proved surprisingly difficult to solve, such as AI being easily fooled~\citep{nguyen2015deep, szegedy2013intriguing} and otherwise overconfidently wrong (i.e., hallucinations)~\citep{maynez2020faithfulness}, though progress has been made~\citep{openai2025gpt5,Huang2025HallucinationSurvey}. Such challenges could persist, preventing us from reliably trusting the outputs of systems like \ouralgo. It is also not clear to what extent AI systems can produce extremely novel, creative ideas that resemble great conceptual leaps in science. Studying and improving AI systems on these fronts are key areas for future research.

Currently, \ouralgo conducts computational experiments only. In future work, this same playbook could be applied to other scientific domains where one can automatically conduct experiments (or have humans conduct them) and get data back from them (e.g. automated chemistry labs, on which swift progress is being made~\citep{boiko2023autonomous}).

The ability to automate paper generation raises significant ethical and societal concerns, including the potential to overwhelm the peer review process, artificially inflate research credentials, repurpose the ideas of others without giving proper credit, eliminate scientist jobs, and/or conduct unethical or dangerous experiments (Supplementary \Cref{app:limitations}).
To conduct this study responsibly, we obtained explicit permission from ICLR leadership, workshop organizers, and the University of British Columbia's IRB (H24-02652).
Crucially, as part of our experimental protocol, we determined in advance that all AI-generated submissions would be withdrawn after peer review, regardless of outcome.
This decision was made to avoid setting a precedent for publishing fully automated research before the scientific community has established clear standards for disclosure and evaluation.
Developing these norms is a critical next step to ensure such systems are used to advance, not undermine, scientific integrity.
Finally, more research is needed to ensure open-ended exploratory AI proceeds safely and in alignment with human values~\citep{bommasani2021opportunities, ecoffet2020open}.

\ouralgo's generation of the first AI-authored manuscript to pass peer review at a major machine learning workshop marks a milestone in the centuries-long scientific endeavor.
While challenges remain in consistency and achieving top-tier quality, this success demonstrates AI's growing capacity for scientific reasoning and signals the dawn of a new era where the process of discovery is no longer a solely human pursuit, and where the pace at which we are able to reap the harvest of scientific discovery could accelerate dramatically.

\clearpage
\section*{Methods}
\label{sec:methods}

Our research methodology is centered around two core automated systems: an AI Scientist for generating novel scientific research and an \ourreviewer for its rigorous evaluation. These systems work in concert to explore the potential of AI in accelerating scientific discovery.

\subsection*{\ouralgo}
\ouralgo is an agentic system designed to autonomously conduct machine learning research. We present results for two modes: a template-based system that extends human-provided code, and a more open-ended template-free system that operates with much less prior guidance.
The detailed prompts used for each system are provided in Supplementary \Cref{app:v1_prompts_scientist,app:v2_prompts_scientist}. More results and analyses of the papers generated by each system are provided in \Cref{app:template_based_tables,app:case_study_v1,app:case_study_v2,appsec:template_based_papers,appsec:template_free_papers}.

\subsubsection*{Foundational Technologies}
Both versions are built upon autoregressive large language models (LLMs)~\citep{gpt4, claude3, llama3} that learn to generate text by modeling the conditional probability of a new token given preceding tokens. Through vast data and model scaling, LLMs exhibit human-like abilities, including reasoning and code generation. Agentic patterns~\citep{llm_agent_survey} such as few-shot prompting~\citep{brown2020language}
and self-reflection~\citep{shinn2024reflexion} are leveraged by \ouralgo to improve performance and reliability.
For code generation, the template-based system uses the state-of-the-art open-source coding assistant, Aider~\citep{aider}, which is designed to implement features, fix bugs, or refactor code in existing codebases.
To go further and effectively use more test-time compute, the template-free system uses LLMs to power tree search without relying on Aider.

\paragraph{Template-Based AI Scientist.}
\label{subsec:template-method}

The system is provided with a starting code template that reproduces a simple training run from a popular algorithm on a standard benchmark (e.g., training a small transformer~\citep{vaswani2017attention} on the works of Shakespeare). Its workflow unfolds in three phases:
\begin{enumerate}
    \item \textbf{Idea Generation:} The process begins with a simple experiment defined by a human-provided code template. The system then enters an iterative loop of idea generation and refinement using LLMs as a mutation operator. In each iteration, it proposes a batch of new research ideas that are variations or extensions of existing ideas in its growing archive. Each idea is a structured object containing a descriptive title, a summary of the core hypothesis, a detailed experimental plan, and self-assessed scores for interestingness (1-10 scale), novelty (1-10 scale), and feasibility (1-10 scale). This iterative growth of an idea archive is inspired by open-endedness algorithms that maintain a diverse collection of artifacts~\citep{stanley2017open, lehman2022evolutionlargemodels}. To enforce novelty, each proposed idea is automatically checked against the scientific literature via the Semantic Scholar API~\citep{fricke2018semantic}; ideas with high semantic similarity to existing work are discarded. The system is prompted to act as an ``ambitious AI PhD student who is looking to publish a paper that will contribute significantly to the field.'' For novelty assessment, the system conducts up to 10 rounds of literature search queries, with each round allowing the system to refine its search based on previous results.
    
    \item \textbf{Experiment Execution:} Once a promising idea is selected from the archive, the system devises a multi-step experimental plan with up to 5 experiments. It then executes this plan sequentially, using Aider to modify the codebase. A key feature of this phase is its robustness to runtime errors. The system automatically detects execution failures, captures the error logs, and invokes an instance of the Aider agent~\citep{aider} to perform automated debugging. The Aider agent is prompted with the failing code and the error message to generate a patch, with up to 4 re-attempt cycles per experiment. The corrected code is then used to re-run the experiment with a timeout of 7200 seconds per experiment. All experimental outcomes, including metrics, generated plots, and observations, are logged in an experimental journal. This journal serves as a form of memory, informing the subsequent steps in the experimental plan.
    
    \item \textbf{Manuscript Generation:} Upon completion of the experimental phase, the system synthesizes the findings into a full scientific paper. To do so, it uses Aider to populate 
    a standard conference \LaTeX{} template, writing sections including Introduction, Methods, Results, and Conclusion. The Results section is written by analyzing the experimental journal, summarizing key findings, and embedding the generated figures. To situate the work within the broader scientific context, the system constructs a Related Work section by querying the Semantic Scholar API for relevant literature (up to 20 search rounds) and generating summaries for each cited paper. The manuscript undergoes several passes of automated editing and refinement to improve clarity and coherence. Finally, the system compiles the \LaTeX{} source, automatically correcting any compilation errors (up to 5 correction rounds) to produce a final PDF.
\end{enumerate}

\paragraph{Template-Free AI Scientist.}
\label{subsec:template-free-method}

To overcome the limitations of a fixed starting codebase, we developed a template-free version capable of more open-ended discovery.
We use OpenAI's o3 for idea generation and code critique during experiments due to its strong reasoning capabilities, Anthropic's Claude Sonnet 4 for code generation, OpenAI's GPT-4o for cost-efficient vision-language tasks, and OpenAI's o4-mini for cost-efficient reasoning during the review stage.
This version introduces several key enhancements:
\begin{itemize}
    \item \textbf{Generalized Idea Generation:} This system's ideation process is more abstract and not tethered to an initial code implementation. It begins by generating high-level research proposals that resemble the abstract of a scientific paper. These proposals articulate a research problem, propose a novel method, and hypothesize the expected outcomes. To ensure the proposals are both grounded and novel, this process is tightly integrated with a literature review module that queries external academic databases to identify knowledge gaps and avoid rediscovering existing work. The system uses structured prompts to guide idea generation, with reflection rounds to refine proposals based on literature search results.
    (See Supplementary \Cref{app:v2_prompts_scientist} for prompts.)

    \item 
    \textbf{Experiment Progress Manager:} 
    Real-world scientific experimentation typically proceeds through distinct stages, from initial feasibility assessments to detailed ablation analyses. To emulate this structured approach, an \textbf{experiment progress manager} is introduced to coordinate four clearly defined stages of scientific experimentation:
    starting with a \textit{Preliminary Investigation} to test basic viability, moving to \textit{Hyperparameter Tuning} for optimization, then to the main \textit{Research Agenda Execution}, and concluding with \textit{Ablation Studies} to understand the contribution of different components. 
    Each stage has explicit stopping criteria. Stage 1 concludes when a basic working prototype is successfully executed. Stage 2 ends when experiments stabilize, as indicated by convergence in training curves and successful execution across at least two datasets. Stages 3 and 4 conclude when the allocated computational budget is exhausted.
    Each stage conducts its own tree search, where the specifics of this tree search process are detailed in the following bullet point.
    Each node has a maximum experiment runtime of 1 hour. At the end of each stage, an LLM-based evaluator assesses all leaf nodes and selects the most promising one to serve as the root for the next stage of exploration, effectively pruning less promising research avenues. 

    \item \textbf{Parallelized Agentic Tree Search for Experimentation:}
    To manage the complexity of open-ended research, the sequential workflow of the template-based version of \ouralgo is replaced with a parallelized agentic tree. 
    \Cref{fig:tree_ablations}A provides an overview of the approach, while \Cref{fig:tree_ablations}B shows a generated tree from an actual run.
    By default, it uses Claude Sonnet 4 for code generation. We provide results for different LLM model choices in \Cref{fig:conceptual}B.
    
Each experimental node within agentic tree search undergoes the following execution cycle: Claude Sonnet 4 first generates both a concrete experimentation plan and the associated Python code to implement the experiment. The generated code is immediately executed in a Python interpreter. If the execution encounters an error, the error message is recorded, and the node is marked as \textbf{buggy}, ending the current execution cycle for that node. If the execution succeeds, the experiment proceeds to the plotting phase. %

The system is prompted to save all relevant experimental outputs (training and validation metrics, losses, etc.) into structured numpy files. In the plotting phase, \ouralgo reads these stored results and the code, generating visualizations that summarize and illustrate the findings. These visualizations are subsequently passed to a Vision-Language Model (VLM) for critique. Any issues flagged by the VLM (such as unclear labels, missing legends, or misleading visualizations) result in the node being marked as \textbf{buggy}, and this feedback is recorded for future debugging. Nodes that successfully execute and pass the VLM review without issue are designated as \textbf{non-buggy}.

Each node is defined as a collection comprising an experiment script (e.g., a Python file), a textual description of the high-level plan implemented in the script, an execution error trace (if applicable), experiment runtime, performance metrics recorded during the experiment, code critique from OpenAI's o3 after running the script, a visualization script, file paths to the generated figures, feedback from a VLM on those figures, and the node's final status (either buggy or non-buggy).

At each iteration, the system selects several nodes from the existing tree to expand in parallel.
With a predefined probability, a \textbf{buggy node} is chosen (thus prioritizing error resolution and debugging); otherwise, a \textbf{non-buggy} node is selected for further refinement and improvement.
When choosing between non-buggy nodes, the system uses a \textbf{best-first search strategy}, guided by GPT-4o that evaluates candidates based on factors like performance metrics, training dynamics, and the quality of generated plots.
The selected nodes are expanded by creating a new child node, where the system attempts debugging if the parent node was buggy, or refines and improves upon the previous experiment if the parent was non-buggy.
Claude Sonnet 4 is used to generate the plan and experiment code for each new child node, after which all new nodes are executed concurrently in parallel, significantly accelerating the exploration process.
In addition to buggy and non-buggy nodes, the system uses specialized node variants tailored to specific experimental needs:

\begin{itemize}
\item \textbf{Hyperparameter nodes} systematically explore alternative hyperparameter configurations during Stage 2. The system maintains records of previously tested hyperparameters, preventing redundant experiments. Errors encountered during hyperparameter tuning trigger the creation of corresponding debug nodes.
\item \textbf{Ablation nodes} evaluate crucial ablation studies during Stage 4, assessing the importance of various components or assumptions underlying the experiment. Similar to hyperparameter nodes, previously tested ablation conditions are tracked to avoid repetition, and debugging nodes are created in response to any encountered errors.
\item \textbf{Replication nodes} execute replicates of their parent experiments using different random seeds. Typically, several replication nodes are created to enable the calculation of statistical measures (mean and standard deviation) of experimental outcomes, enhancing result robustness.
\item \textbf{Aggregation nodes} are special nodes created to consolidate and visualize the combined results of replication nodes. Unlike other node types, aggregation nodes do not conduct new experiments but simply generate a Python script to aggregate and summarize prior results, producing figures that explicitly show means and standard deviations.
\end{itemize}

The structured design of experiment stages and tailored node types facilitates systematic exploration across all stages.
Unlike some LLM agents that rigidly follow predefined, fine-grained workflow graphs, the \ouralgo adopts a looser structure that guides the entire empirical research cycle, enabling flexible system behavior while maintaining coherence across iterative stages.
See Supplementary \Cref{app:v2_prompts_scientist} and \Cref{app:v2_hyperparams} for the prompts and detailed hyperparameters, respectively.

    \item \textbf{Vision-Language Model (VLM) Integration:} This system incorporates Vision-Language Models (VLMs) using GPT-4o 
    to analyze and provide feedback on visual data. During experimentation, generated plots are fed to a VLM, which is prompted to act as a scientist and critique them. 
    For example, it might flag nonsensical axes, issues in the quality of generated examples, or suggest clearer ways to present the data.
    This feedback is used to generate new experimental nodes in the tree search aimed at addressing the identified issues. During manuscript preparation, the VLM assesses the alignment between figures and their corresponding captions, ensuring that the caption accurately describes the plot and highlights the key takeaways, thus improving the overall quality and clarity of the paper. The VLM reviews include detailed analyses of figure content, caption accuracy, and integration with the main text.
    (See Supplementary \Cref{app:v2_prompts_scientist} for prompts.)

    \item \textbf{Generalized Dataset Access:} 
    To broaden its research capabilities, the system is prompted to dynamically integrate datasets from public repositories by formulating queries to the HuggingFace Hub~\citep{wolf2020huggingface}. 
    A set of 10 example datasets available on HuggingFace is listed in the prompt, and the system can automatically generate the necessary data-loading code to use a selected dataset in its experiments. 
    This approach partially relaxes the constraint of working with a fixed, pre-defined set of datasets by allowing human scientists to easily update the candidate list.
    For datasets not available on HuggingFace, human scientists can download them from public data repositories (e.g., open-access archives), store them locally, and add usage instructions to the prompt. These locally stored datasets can then be used alongside HuggingFace datasets by \ouralgo.
    (See Supplementary \Cref{app:v2_prompts_scientist} for prompts.)
    
    \item \textbf{Enhanced Manuscript Writing:} The template-free system moves away from the incremental Aider-based approach to direct LaTeX generation using a reasoning model such as OpenAI's o1~\citep{OpenAIOS} followed by reflection~\citep{shinn2023reflexion}. The system first aggregates experimental results from multiple stages into compound figures using a dedicated plot aggregation step. The manuscript writing process includes specific prompts for different workshop formats (e.g., the ICBINB workshop focusing on negative results), with detailed guidelines for each section, including the title, abstract, introduction, methods, experiments, and conclusions. The system undergoes multiple reflection cycles, incorporating feedback from LaTeX linters and VLM reviews to improve figure quality and text-figure alignment. (See our code and Supplementary \Cref{app:v2_prompts_scientist} for prompts and full details.)
\end{itemize}

The complete generation process for the template-free system typically takes several hours to over 15 hours, depending on problem complexity.

\subsection*{\ourreviewer}
To assess the quality of the AI-generated research, we built an \ourreviewer using o4-mini~\citep{openai2025o3o4mini}. This component is designed to emulate the peer-review process of a top-tier machine learning conference by adhering to the official NeurIPS reviewer guidelines. The agent processes the manuscript's PDF to produce a structured review, including numerical scores for soundness, presentation, and contribution, along with a list of strengths and weaknesses and a preliminary accept/reject decision (Supplementary \Cref{app:reviewer_details}). All prompts used for \ourreviewer are provided in Supplementary \Cref{app:v1_prompts_reviewer}. 

\paragraph{Review Process.}
The \ourreviewer follows a multi-stage process. First, the system is prompted with the role: ``You are an AI researcher who is reviewing a paper that was submitted to a prestigious ML venue.'' The review prompt provides the paper content along with detailed NeurIPS reviewer guidelines and asks for a structured JSON response including summary, strengths, weaknesses, questions, limitations, ethical concerns, and numerical scores (soundness, presentation, contribution, overall score 1-10, and confidence level). To improve robustness, the final assessment is a meta-review that ensembles five independent reviews. The five reviews are generated for each paper and aggregated into a single meta-review, with an LLM taking the role of an Area Chair to find consensus among the individual reviews.

\paragraph{Validation.}

We benchmarked the \ourreviewer against human decisions using ICLR data from the publicly available OpenReview dataset~\citep{gonzalez2024learning}. 
The \ourreviewer achieves a comparable balanced accuracy with humans (69\% versus 66\%; see Supplementary \Cref{app:v1_reviewer_details} for details) and a higher F1-score compared to inter-human group agreement (0.62 versus 0.49) in the NeurIPS 2021 consistency experiment~\citep{beygelzimer2021neurips}, where roughly 10\% of submissions were randomly selected and sent to two independent review committees, providing a real-world benchmark of inter-reviewer consistency (\Cref{tab:reviewers_main_paper}).
These results suggest that LLM-based agents can provide valuable feedback that aligns with the average human expert opinion.
We highlight that there was a different set of paper submissions in the ICLR and NeurIPS paper pools and thus distribution shift, so this comparison is not exact.
However, ICLR is the only major ML conference that releases all accept and reject decisions for us to perform the analysis, and the NeurIPS 2021 experiment is the only modern version of the human consistency experiment, and is thus the only possible comparison.

\newpage
\section*{Data Availability}

For the nanoGPT~\citep{karpathy2022nanogpt} experiments, the template-based version of \ouralgo used the Shakespeare character~\citep{char_shakespeare}, enwiki8~\citep{hutter_prize} and text8~\citep{text8} datasets.
The template-free version of \ouralgo used the Crop Pest and Disease Detection dataset~\citep{mensah2023crop} for one of the papers submitted to the ICLR workshop experiment, and the Waterbirds~\citep{waterbirds_dataset} and CelebA~\citep{liu2015faceattributes} datasets for the experiments shown in \Cref{fig:conceptual}B, \Cref{fig:tree_ablations}B, and \Cref{fig:tree_ablations}C.
In all other cases, the template-free version used datasets available through the HuggingFace Hub~\citep{wolf2020huggingface}.

\section*{Code Availability}
\label{sec:code_availability}

The code for the template-based version of \ouralgo and \ourreviewer is available at \url{https://github.com/SakanaAI/AI-Scientist}.
Additionally, the code for the template-free version of \ouralgo is available at \url{https://github.com/SakanaAI/AI-Scientist-v2}.
Both code repositories are licensed under the Apache License 2.0.

\section*{Acknowledgements}
The authors would like to thank Irene Zhang, Johannes von Oswald, Takuya Akiba, Yujin Tang, Kosuke Nakago, Kou Misaki, Haruki Goda, Yuichi Inoue, Aaron Dharna, Ben Norman, Jenny Zhang, Anna Olerinyova, and Felicitas Muecke-Wegner for helpful feedback.
This work was supported by grants from Schmidt Futures, NSERC, the Vector Institute, the Canada CIFAR AI Chairs program, and a donation from Rafael Cosman.

\section*{Author Contributions}

There are four authors who contributed equally to this work, and are listed in alphabetical order.

\textbf{Yutaro Yamada [shared first, equal corresponding]:} Co-led the project and contributed core ideas. Coded the core tree-search and template-free version of \ouralgo. Ran paper generation experiments. Read and validated the work of many AI-generated papers to select submissions and checked the paper code implementations. Led the writing of the paper. Wrote detailed analyses of the submitted papers for our manuscript.

\textbf{Robert T. Lange [shared first, equal corresponding]:} Co-initiated, co-led the project, and contributed core ideas. Conceived and coded core parts of the \ourreviewer, tailored the paper generation pipeline to the workshop, and ran the paper generation experiments. Organized the workshop communication process. Read and validated the work of many AI-generated papers to select submissions and checked the paper code implementations. Led the writing of the paper. Wrote detailed analyses of the submitted papers for our manuscript.

\textbf{Cong Lu [shared first, equal corresponding]:} Initiated the project, co-led the project, and conceived the original ideas for \ouralgo, including the use of SWE agents like Aider to execute scientific ideas autonomously. Coded core parts of the idea generation, \ourreviewer, tool use, experiment aggregation, and paper writing framework. Wrote and led the IRB approval process for the workshop experiments, and evaluated AI-generated paper submissions. Led the writing of the paper.

\textbf{Chris Lu [shared first, equal corresponding]:} Initiated the project, co-led the project. Conceived the original idea and structure for \ouralgo and developed the first working system, which demonstrated autonomous end-to-end paper generation. Conceived the experiment setup and evaluations in the original preprint. Led the development of the template-based version of \ouralgo. Ran paper generation experiments. Led the writing of the paper. Provided advice, feedback, and writing.

\textbf{Shengran Hu:} Enhanced the iterative \ourreviewer with VLM feedback, contributed to the experiment and paper writing framework, ran \ourreviewer benchmark and ablation experiments, helped run paper generation experiments, read and validated the work of many AI-generated papers to select submissions, and checked the paper code implementations. Helped write and iterate over drafts of the paper. Helped write the IRB approval.

\textbf{Jakob Foerster:} Provided advice, feedback, and writing.

\textbf{David Ha [equal corresponding]:} Provided overarching guidance for the research project, offering technical insight, advice, feedback, and writing. Oversaw the public communication process.

\textbf{Jeff Clune [equal corresponding]:} Provided overarching guidance for the research project, offering technical insight, advice, feedback, and writing. Oversaw the IRB application process. Evaluated AI-generated paper submissions.

\section*{Competing Interests}
The Principal Investigator, Jeff Clune, has affiliations with the Vector Institute and Google DeepMind.
This project has Vector Institute affiliations but is not connected to Google DeepMind.
Several co-authors are employees of or consultants for Sakana AI, a machine learning research company involved in the design of \ouralgo.
The Principal Investigator is not financially compensated by Sakana AI.
These arrangements have been reviewed and approved by the University of British Columbia in accordance with its conflict of interest policies.

\section*{Ethics Approval}
This study received ethics approval from The University of British Columbia Behavioral Research Ethics Board (BREB) (protocol number H24-02652).
The research was conducted in full cooperation with the ICLR conference leadership and the relevant workshop organizers.
In accordance with the approved protocol, human participants (peer reviewers) were informed that a small number of submissions to the workshop were AI-generated, although not which specific papers.
Participants had the option to opt out of reviewing any potentially AI-generated manuscripts.
All AI-generated submissions were withdrawn following the review process, regardless of the outcome.

\clearpage
\hypersetup{pageanchor=false}
\setcounter{page}{1}
\begin{appendices}
\vspace*{5pt}

\section*{SI Table of Contents}
\vspace*{-5pt}
\startcontents[sections]
\printcontents[sections]{l}{1}{\setcounter{tocdepth}{3}}

\clearpage

\section{Supplementary Methods}
\label{app:sup_methods}

\subsection{Template-Based AI Scientist Implementation Details}
\label{app:v1_details}

This section outlines the implementation details for the template-based version of \ouralgo. This version is built upon Large Language Models (LLMs), which are embedded into agentic frameworks to enhance their reasoning and coding capabilities. 

\subsubsection{Prompts for Template-Based Scientific Discovery}
\label{app:v1_prompts_scientist}
The template-based version of \ouralgo uses structured prompts to guide Large Language Models (LLMs) through the research stages from idea generation to paper writing.

The process begins by setting the overall goal for \ouralgo.
\begin{tcolorbox}[breakable,colback=orange!5!white, colframe=orange!80!black, title=Idea Generation System Prompt]
\texttt{You are an ambitious AI PhD student who is looking to publish a paper that will contribute significantly to the field.}
\end{tcolorbox}

\ouralgo is then prompted to generate a specific research idea which includes an experimental plan, and self-assessed scores on novelty, feasibility, and interestingness.
\begin{tcolorbox}[breakable, colback=orange!5!white, colframe=orange!80!black, title=Idea Generation Prompt]
\begin{Verbatim}[breaklines, breaksymbol={}, breaksymbolleft={}, fontsize=\small]
{task_description}
<experiment.py>
{code}
</experiment.py>

Here are the ideas that you have already generated:

'''
{prev_ideas_string}
'''

Come up with the next impactful and creative idea for research experiments and directions you can feasibly investigate with the code provided. Note that you will not have access to any additional resources or datasets. Make sure any idea is not overfit the specific training dataset or model, and has wider significance.

Respond in the following format:

THOUGHT:
<THOUGHT>

NEW IDEA JSON:
```json
<JSON>
```

In <THOUGHT>, first briefly discuss your intuitions and motivations for the idea. Detail your high-level plan, necessary design choices and ideal outcomes of the experiments. Justify how the idea is different from the existing ones.

In <JSON>, provide the new idea in JSON format with the following fields:
- "Name": A shortened descriptor of the idea. Lowercase, no spaces, underscores allowed.
- "Title": A title for the idea, will be used for the report writing.
- "Experiment": An outline of the implementation. E.g. which functions need to be added or modified, how results will be obtained, ...
- "Interestingness": A rating from 1 to 10 (lowest to highest).
- "Feasibility": A rating from 1 to 10 (lowest to highest).
- "Novelty": A rating from 1 to 10 (lowest to highest).

Be cautious and realistic on your ratings.
This JSON will be automatically parsed, so ensure the format is precise.
You will have {num_reflections} rounds to iterate on the idea, but do not need to use them all.
\end{Verbatim}
\end{tcolorbox}

To assess the novelty of the generated idea, \ouralgo is instructed to use the Semantic Scholar API to search the literature.
\begin{tcolorbox}[breakable,colback=orange!5!white, colframe=orange!80!black, title=Idea Novelty System Prompt]
\begin{Verbatim}[breaklines, breaksymbol={}, breaksymbolleft={}, fontsize=\small]
You are an ambitious AI PhD student who is looking to publish a paper that will contribute significantly to the field.
You have an idea and you want to check if it is novel or not. I.e., not overlapping significantly with existing literature or already well explored.
Be a harsh critic for novelty, ensure there is a sufficient contribution in the idea for a new conference or workshop paper.
You will be given access to the Semantic Scholar API, which you may use to survey the literature and find relevant papers to help you make your decision.
The top 10 results for any search query will be presented to you with the abstracts.

You will be given {num_rounds} to decide on the paper, but you do not need to use them all.
At any round, you may exit early and decide on the novelty of the idea.
Decide a paper idea is novel if after sufficient searching, you have not found a paper that significantly overlaps with your idea.
Decide a paper idea is not novel, if you have found a paper that significantly overlaps with your idea.

{task_description}
<experiment.py>
{code}
</experiment.py>
\end{Verbatim}
\end{tcolorbox}

The novelty check proceeds iteratively, allowing multiple queries.
\begin{tcolorbox}[breakable, colback=orange!5!white, colframe=orange!80!black, title=Idea Novelty Prompt]
\begin{Verbatim}[breaklines, breaksymbol={}, breaksymbolleft={}, fontsize=\small]
Round {current_round}/{num_rounds}.
You have this idea:

"""
{idea}
"""

The results of the last query are (empty on first round):
"""
{last_query_results}
"""

Respond in the following format:

THOUGHT:
<THOUGHT>

RESPONSE:
```json
<JSON>
```

In <THOUGHT>, first briefly reason over the idea and identify any query that could help you make your decision.
If you have made your decision, add "Decision made: novel." or "Decision made: not novel." to your thoughts.

In <JSON>, respond in JSON format with ONLY the following field:
- "Query": An optional search query to search the literature (e.g. attention is all you need). You must make a query if you have not decided this round.

A query will work best if you are able to recall the exact name of the paper you are looking for, or the authors.
This JSON will be automatically parsed, so ensure the format is precise.
\end{Verbatim}
\end{tcolorbox}

For experiment implementation and execution, \ouralgo uses the state-of-the-art code editor Aider.
\begin{tcolorbox}[breakable, colback=orange!5!white, colframe=orange!80!black, title=Experiment Running Aider Prompt]
\begin{Verbatim}[breaklines, breaksymbol={}, breaksymbolleft={}, fontsize=\small]
Your goal is to implement the following idea: {title}.
The proposed experiment is as follows: {idea}.
You are given a total of up to {max_runs} runs to complete the necessary experiments. You do not need to use all {max_runs}.

First, plan the list of experiments you would like to run. For example, if you are sweeping over a specific hyperparameter, plan each value you would like to test for each run.

Note that we already provide the vanilla baseline results, so you do not need to re-run it.

For reference, the baseline results are as follows:

{baseline_results}

After you complete each change, we will run the command `python experiment.py --out_dir=run_i' where i is the run number and evaluate the results.
YOUR PROPOSED CHANGE MUST USE THIS COMMAND FORMAT, DO NOT ADD ADDITIONAL COMMAND LINE ARGS.
You can then implement the next thing on your list.
\end{Verbatim}
\end{tcolorbox}

After experiments, \ouralgo uses Aider to generate plots from the results and records experimental details.
\begin{tcolorbox}[breakable, colback=orange!5!white, colframe=orange!80!black, title=Plotting Aider Prompt]
\begin{Verbatim}[breaklines, breaksymbol={}, breaksymbolleft={}, fontsize=\small]
Great job! Please modify `plot.py` to generate the most relevant plots for the final writeup. 

In particular, be sure to fill in the "labels" dictionary with the correct names for each run that you want to plot.

Only the runs in the `labels` dictionary will be plotted, so make sure to include all relevant runs.

We will be running the command `python plot.py` to generate the plots.

---

Please modify `notes.txt` with a description of what each plot shows along with the filename of the figure. Please do so in-depth.

Somebody else will be using `notes.txt` to write a report on this in the future.
\end{Verbatim}
\end{tcolorbox}

For manuscript writing, \ouralgo uses Aider to fill a LaTeX template section by section with the produced plots and results.
\begin{tcolorbox}[breakable, colback=orange!5!white, colframe=orange!80!black, title=Paper Writing Aider Prompt]
\begin{Verbatim}[breaklines, breaksymbol={}, breaksymbolleft={}, fontsize=\small]
We've provided the `latex/template.tex` file to the project. We will be filling it in section by section.

First, please fill in the {section} section of the writeup.

Some tips are provided below:
{per_section_tips}

Before every paragraph, please include a brief description of what you plan to write in that paragraph in a comment.

Be sure to first name the file and use *SEARCH/REPLACE* blocks to perform these edits.
\end{Verbatim}
\end{tcolorbox}

\subsubsection{An Example Template Codebase for Template-Based Runs}
\label{app:v1_template_codebase_example}

We provide an example template codebase used in our template-based setting. 
Additional templates are available in our GitHub repository: 
\url{https://github.com/SakanaAI/AI-Scientist/tree/main/templates}.

\begin{tcolorbox}[breakable,colback=orange!5!white, colframe=orange!80!black, title=Generalized Idea Generation System and User Prompt]
\begin{Verbatim}[breaklines, breaksymbol={}, breaksymbolleft={}, fontsize=\small]
# This file trains a DDPM diffusion model on 2D datasets.

import argparse
import json
import os.path as osp
import pathlib
import pickle
import time

import npeet.entropy_estimators as ee
import numpy as np
import torch
from torch import nn
from torch.nn import functional as F
from torch.optim.lr_scheduler import CosineAnnealingLR
from torch.utils.data import DataLoader
from tqdm.auto import tqdm

import datasets
from ema_pytorch import EMA

device = torch.device("cuda" if torch.cuda.is_available() else "cpu")


class SinusoidalEmbedding(nn.Module):
    def __init__(self, dim: int, scale: float = 1.0):
        super().__init__()
        self.dim = dim
        self.scale = scale

    def forward(self, x: torch.Tensor):
        x = x * self.scale
        half_dim = self.dim // 2
        emb = torch.log(torch.Tensor([10000.0])) / (half_dim - 1)
        emb = torch.exp(-emb * torch.arange(half_dim)).to(device)
        emb = x.unsqueeze(-1) * emb.unsqueeze(0)
        emb = torch.cat((torch.sin(emb), torch.cos(emb)), dim=-1)
        return emb


class ResidualBlock(nn.Module):
    def __init__(self, width: int):
        super().__init__()
        self.ff = nn.Linear(width, width)
        self.act = nn.ReLU()

    def forward(self, x: torch.Tensor):
        return x + self.ff(self.act(x))


class MLPDenoiser(nn.Module):
    def __init__(
            self,
            embedding_dim: int = 128,
            hidden_dim: int = 256,
            hidden_layers: int = 3,
    ):
        super().__init__()
        self.time_mlp = SinusoidalEmbedding(embedding_dim)
        # sinusoidal embeddings help capture high-frequency patterns for low-dim data
        self.input_mlp1 = SinusoidalEmbedding(eAs with human-authored research, some AI-generated ideambedding_dim, scale=25.0)
        self.input_mlp2 = SinusoidalEmbedding(embedding_dim, scale=25.0)

        self.network = nn.Sequential(
            nn.Linear(embedding_dim * 3, hidden_dim),
            *[ResidualBlock(hidden_dim) for _ in range(hidden_layers)],
            nn.ReLU(),
            nn.Linear(hidden_dim, 2),
        )

    def forward(self, x, t):
        x1_emb = self.input_mlp1(x[:, 0])
        x2_emb = self.input_mlp2(x[:, 1])
        t_emb = self.time_mlp(t)
        emb = torch.cat([x1_emb, x2_emb, t_emb], dim=-1)
        return self.network(emb)


class NoiseScheduler():
    def __init__(
            self,
            num_timesteps=1000,
            beta_start=0.0001,
            beta_end=0.02,
            beta_schedule="linear",
    ):
        self.num_timesteps = num_timesteps
        if beta_schedule == "linear":
            self.betas = torch.linspace(
                beta_start, beta_end, num_timesteps, dtype=torch.float32).to(device)
        elif beta_schedule == "quadratic":
            self.betas = (torch.linspace(
                beta_start ** 0.5, beta_end ** 0.5, num_timesteps, dtype=torch.float32) ** 2).to(device)
        else:
            raise ValueError(f"Unknown beta schedule: {beta_schedule}")

        self.alphas = 1.0 - self.betas
        self.alphas_cumprod = torch.cumprod(self.alphas, axis=0).to(device)
        self.alphas_cumprod_prev = F.pad(self.alphas_cumprod[:-1], (1, 0), value=1.).to(device)

        # required for self.add_noise
        self.sqrt_alphas_cumprod = (self.alphas_cumprod ** 0.5).to(device)
        self.sqrt_one_minus_alphas_cumprod = ((1 - self.alphas_cumprod) ** 0.5).to(device)

        # required for reconstruct_x0
        self.sqrt_inv_alphas_cumprod = torch.sqrt(1 / self.alphas_cumprod).to(device)
        self.sqrt_inv_alphas_cumprod_minus_one = torch.sqrt(
            1 / self.alphas_cumprod - 1).to(device)

        # required for q_posterior
        self.posterior_mean_coef1 = self.betas * torch.sqrt(self.alphas_cumprod_prev) / (1. - self.alphas_cumprod).to(
            device)
        self.posterior_mean_coef2 = ((1. - self.alphas_cumprod_prev) * torch.sqrt(self.alphas) / (
                1. - self.alphas_cumprod)).to(device)

    def reconstruct_x0(self, x_t, t, noise):
        s1 = self.sqrt_inv_alphas_cumprod[t]
        s2 = self.sqrt_inv_alphas_cumprod_minus_one[t]
        s1 = s1.reshape(-1, 1)
        s2 = s2.reshape(-1, 1)
        return s1 * x_t - s2 * noise

    def q_posterior(self, x_0, x_t, t):
        s1 = self.posterior_mean_coef1[t]
        s2 = self.posterior_mean_coef2[t]
        s1 = s1.reshape(-1, 1)
        s2 = s2.reshape(-1, 1)
        mu = s1 * x_0 + s2 * x_t
        return mu

    def get_variance(self, t):
        if t == 0:
            return 0

        variance = self.betas[t] * (1. - self.alphas_cumprod_prev[t]) / (1. - self.alphas_cumprod[t])
        variance = variance.clip(1e-20)
        return variance

    def step(self, model_output, timestep, sample):
        t = timestep
        pred_original_sample = self.reconstruct_x0(sample, t, model_output)
        pred_prev_sample = self.q_posterior(pred_original_sample, sample, t)

        variance = 0
        if t > 0:
            noise = torch.randn_like(model_output)
            variance = (self.get_variance(t) ** 0.5) * noise

        pred_prev_sample = pred_prev_sample + variance

        return pred_prev_sample

    def add_noise(self, x_start, x_noise, timesteps):
        s1 = self.sqrt_alphas_cumprod[timesteps]
        s2 = self.sqrt_one_minus_alphas_cumprod[timesteps]

        s1 = s1.reshape(-1, 1)
        s2 = s2.reshape(-1, 1)

        return s1 * x_start + s2 * x_noise

    def __len__(self):
        return self.num_timesteps


if __name__ == "__main__":
    parser = argparse.ArgumentParser()
    parser.add_argument("--train_batch_size", type=int, default=256)
    parser.add_argument("--eval_batch_size", type=int, default=10000)
    parser.add_argument("--learning_rate", type=float, default=3e-4)
    parser.add_argument("--num_timesteps", type=int, default=100)
    parser.add_argument("--num_train_steps", type=int, default=10000)
    parser.add_argument("--beta_schedule", type=str, default="linear", choices=["linear", "quadratic"])
    parser.add_argument("--embedding_dim", type=int, default=128)
    parser.add_argument("--hidden_size", type=int, default=256)
    parser.add_argument("--hidden_layers", type=int, default=3)
    parser.add_argument("--out_dir", type=str, default="run_0")
    config = parser.parse_args()

    final_infos = {}
    all_results = {}

    pathlib.Path(config.out_dir).mkdir(parents=True, exist_ok=True)

    for dataset_name in ["circle", "dino", "line", "moons"]:
        dataset = datasets.get_dataset(dataset_name, n=100000)
        dataloader = DataLoader(dataset, batch_size=config.train_batch_size, shuffle=True)

        model = MLPDenoiser(
            embedding_dim=config.embedding_dim,
            hidden_dim=config.hidden_size,
            hidden_layers=config.hidden_layers,
        ).to(device)
        ema_model = EMA(model, beta=0.995, update_every=10).to(device)

        noise_scheduler = NoiseScheduler(num_timesteps=config.num_timesteps, beta_schedule=config.beta_schedule)

        optimizer = torch.optim.AdamW(
            model.parameters(),
            lr=config.learning_rate,
        )
        scheduler = CosineAnnealingLR(optimizer, T_max=config.num_train_steps)
        train_losses = []
        print("Training model...")

        model.train()
        global_step = 0
        progress_bar = tqdm(total=config.num_train_steps, mininterval=10, disable=True)
        progress_bar.set_description("Training")

        start_time = time.time()
        while global_step < config.num_train_steps:
            for batch in dataloader:
                if global_step >= config.num_train_steps:
                    break
                batch = batch[0].to(device)
                noise = torch.randn(batch.shape).to(device)
                timesteps = torch.randint(
                    0, noise_scheduler.num_timesteps, (batch.shape[0],)
                ).long().to(device)

                noisy = noise_scheduler.add_noise(batch, noise, timesteps)
                noise_pred = model(noisy, timesteps)
                loss = F.mse_loss(noise_pred, noise)
                loss.backward()

                nn.utils.clip_grad_norm_(model.parameters(), 0.5)
                optimizer.step()
                optimizer.zero_grad()
                ema_model.update()

                scheduler.step()
                progress_bar.update(1)
                logs = {"loss": loss.detach().item()}
                train_losses.append(loss.detach().item())
                progress_bar.set_postfix(**logs)
                global_step += 1

        progress_bar.close()
        end_time = time.time()
        training_time = end_time - start_time

        # Eval loss
        model.eval()
        eval_losses = []
        for batch in dataloader:
            batch = batch[0].to(device)
            noise = torch.randn(batch.shape).to(device)
            timesteps = torch.randint(
                0, noise_scheduler.num_timesteps, (batch.shape[0],)
            ).long().to(device)
            noisy = noise_scheduler.add_noise(batch, noise, timesteps)
            noise_pred = model(noisy, timesteps)
            loss = F.mse_loss(noise_pred, noise)
            eval_losses.append(loss.detach().item())
        eval_loss = np.mean(eval_losses)

        # Eval image saving
        ema_model.eval()
        sample = torch.randn(config.eval_batch_size, 2).to(device)
        timesteps = list(range(len(noise_scheduler)))[::-1]
        inference_start_time = time.time()
        for t in timesteps:
            t = torch.from_numpy(np.repeat(t, config.eval_batch_size)).long().to(device)
            with torch.no_grad():
                residual = ema_model(sample, t)
            sample = noise_scheduler.step(residual, t[0], sample)
        sample = sample.cpu().numpy()
        inference_end_time = time.time()
        inference_time = inference_end_time - inference_start_time

        # Eval estimated KL
        real_data = dataset.tensors[0].numpy()
        kl_divergence = ee.kldiv(real_data, sample, k=5)

        final_infos[dataset_name] = {
            "means": {
                "training_time": training_time,
                "eval_loss": eval_loss,
                "inference_time": inference_time,
                "kl_divergence": kl_divergence,
            }
        }

        all_results[dataset_name] = {
            "train_losses": train_losses,
            "images": sample,
        }

    with open(osp.join(config.out_dir, "final_info.json"), "w") as f:
        json.dump(final_infos, f)

    with open(osp.join(config.out_dir, "all_results.pkl"), "wb") as f:
        pickle.dump(all_results, f)
\end{Verbatim}
\end{tcolorbox}

\subsubsection{Hyperparameter Configuration for Template-Based Runs}
\label{app:v1_hyperparams}
The execution of the template-based version of \ouralgo used the hyperparameters detailed in \Cref{tab:v1_hyperparameters}.

\begin{table}[h!]
\centering
\caption{\textbf{Key Hyperparameters for the Template-Based version of \ouralgo.}}
\label{tab:v1_hyperparameters}
\begin{tabular}{@{}lll@{}}
\toprule
\textbf{Category} & \textbf{Hyperparameter} & \textbf{Value} \\ \midrule
\multirow{2}{*}{Idea Generation} & Number of Idea Reflections & 3 \\
 & Number of Novelty Search Rounds (Semantic Scholar) & 10 \\ \midrule
\multirow{4}{*}{Experiment Execution} & Max Experiments & 5 \\
 & Max Experiment Re-attempts (on error) & 4 \\
 & Experiment Timeout & 7200 seconds \\
 & Plotting Timeout & 600 seconds \\ \midrule
\multirow{2}{*}{Paper Writing} & Number of Citation Search Rounds (Semantic Scholar) & 20 \\
 & Number of LaTeX Error Correction Rounds & 5 \\ %
\bottomrule
\end{tabular}
\end{table}

\subsection{Template-Free AI Scientist Implementation Details}
\label{app:v2_details}

This section outlines the implementation details for the more open-ended template-free version of \ouralgo, as described in the main text.
This enhanced version operates without requiring human-authored code templates, instead employing generalized idea generation, agentic tree search for experimentation, and integrated Vision-Language Model (VLM) feedback, building upon the foundational capabilities of the template-based version of \ouralgo.

\subsubsection{System Architecture and Execution Flow}

\ouralgo workflow consists of multiple sequential stages, from ideation to experimentation to manuscript generation. 
The transitions across these stages are fully automated, requiring no human intervention. 
A single driver process orchestrates the pipeline: it generates candidate programs via LLM calls, launches experiments, monitors exit codes, logs, and timeouts, and triggers state transitions according to pre-defined criteria. Specifically:

\begin{itemize}

\item{Stage 1 $\rightarrow$ Stage 2: Transition occurs once the system produces runnable code with no runtime errors.}

\item{Stage 2 $\rightarrow$ Stage 3: Triggered when the generated code outperforms the baseline model under pre-defined metrics (which are themselves defined by the LLM immediately after ideation).}

\item{Stage 3 $\rightarrow$ Stage 4: Initiated automatically once the Stage 3 tree search spends its allocated exploration budget.}

\end{itemize}

Each experiment is executed via Python's non-interactive subprocess module, with all interactions between the generated code and the operating system limited to controlled file I/O within per-run working directories. 
After each run, artifacts such as metrics, plots, runtime logs, and VLM feedback are serialized into a typed Python object representing a tree node. 
The driver process then invokes an LLM-based judge to evaluate and select the best node from the current stage to carry forward. 
All tool invocations (code generation, analysis, or VLM feedback) are made through internal Python-callable interfaces that wrap the respective LLM API calls.

\subsubsection{Cross-Stage Consistency}
 
Each stage exports its best node to the following stage, which then uses it to guide subsequent generation.
The transition from ideation to experimentation is managed by passing a finalized idea JSON that encodes the proposed hypothesis, task setup, and evaluation metrics.
The transition from experimentation to manuscript writing is handled by passing a condensed form of the experiment journal summarizing key results and observations, where the summarization is done by an LLM.

This mechanism grounds each later stage in the outputs of earlier ones, mitigating divergence between the initial proposal, results, and final write-up.

\subsubsection{Model Selection per Stage}

Different stages demand distinct capabilities from the underlying models. In our current implementation, we empirically identified the most effective model for each role through small-scale pilot runs—for example, Claude for code generation, and GPT-series models for high-level planning, analysis, and writing. While future foundation models may unify these capabilities, the system is designed to flexibly substitute models per stage without altering the overall architecture.

\subsubsection{Experimental Journal Structure}

In the template-based setting, the journal is a plain-text log summarizing experimental attempts. In the template-free setting, it is implemented as a Python class producing a structured JSON export, where each tree node records the following fields:
- Generated code and execution plan
- Experimental results (metrics, losses, validation scores, etc.)
- Runtime feedback, including error traces or log summaries
- Vision-Language Model (VLM) commentary on generated figures or visual outputs
- LLM judge evaluations and stage transition signals

This structured journal not only ensures reproducibility and auditability across experiments but also provides a machine-readable record that connects all stages of the automated research process.

\subsubsection{Comparison with Deep Research}

Recent systems such as Deep Research~\cite{openai2025DeepResearch} represent major progress in large-scale reasoning-based information synthesis. 
They are designed to retrieve, analyze, and summarize existing information to produce well-researched answers to a user's query. 
However, these systems remain limited to knowledge aggregation, and they cannot design or execute new experiments.

\ouralgo, in contrast, automates the entire research lifecycle involved in producing a single paper. 
It begins with ideation, proceeds through experimental implementation and evaluation, and culminates in writing full scientific manuscripts. 
This capability moves beyond retrieval and synthesis toward autonomous knowledge creation, positioning \ouralgo as a complementary but distinct class of system from Deep Research tools.

\subsubsection{Prompts for Template-Free Scientific Discovery}
\label{app:v2_prompts_scientist}
The template-free version of \ouralgo utilizes structured prompts to guide Large Language Models (LLMs) through its more autonomous research cycle. This includes a more generalized idea generation phase, a sophisticated experiment design and execution process, and manuscript writing enhanced by VLM-assisted refinement (see main text).

The idea generation process begins at a higher level of abstraction than the template-based version. The template-free system is prompted for open-ended thinking, akin to formulating a research abstract, while integrating literature search tools.
\begin{tcolorbox}[breakable,colback=orange!5!white, colframe=orange!80!black, title=Generalized Idea Generation System and User Prompt]
\begin{Verbatim}[breaklines, breaksymbol={}, breaksymbolleft={}, fontsize=\small]
# System prompt
You are an experienced AI researcher who aims to propose high-impact research ideas resembling exciting grant proposals. Feel free to propose any novel ideas or experiments; make sure they are novel. Be very creative and think out of the box. Each proposal should stem from a simple and elegant question, observation, or hypothesis about the topic. For example, they could involve very interesting and simple interventions or investigations that explore new possibilities or challenge existing assumptions. Clearly clarify how the proposal distinguishes from the existing literature.

Ensure that the proposal can be done starting from the provided codebase, and does not require resources beyond what an academic lab could afford. These proposals should lead to papers that are publishable at top ML conferences.

You have access to the following tools:

{tool_descriptions}

Respond in the following format:

ACTION:
<The action to take, exactly one of {tool_names_str}>

ARGUMENTS:
<If ACTION is "SearchSemanticScholar", provide the search query as {{"query": "your search query"}}. If ACTION is "FinalizeIdea", provide the idea details as {{"idea": {{ ... }}}} with the IDEA JSON specified below.>

If you choose to finalize your idea, provide the IDEA JSON in the arguments:

IDEA JSON:
```json
{{
    "Name": "...",
    "Title": "...",
    "Short Hypothesis": "...",
    "Related Work": "...",
    "Abstract": "...",
    "Experiments": "...",
    "Risk Factors and Limitations": "..."
}}
```

Ensure the JSON is properly formatted for automatic parsing.

Note: You should perform at least one literature search before finalizing your idea to ensure it is well-informed by existing research.

# Initial idea generation prompt (example user prompt part)
{workshop_description}

Here are the proposals that you have already generated:

{prev_ideas_string}

Begin by generating an interestingly new high-level research proposal that differs from what you have previously proposed.

# reflection prompt (example user prompt part for reflection)
Round {current_round}/{num_reflections}.

In your thoughts, first carefully consider the quality, novelty, and feasibility of the proposal you just created.
Include any other factors that you think are important in evaluating the proposal.
Ensure the proposal is clear and concise, and the JSON is in the correct format.
Do not make things overly complicated.
In the next attempt, try to refine and improve your proposal.
Stick to the spirit of the original idea unless there are glaring issues.

If you have new information from tools, such as literature search results, incorporate them into your reflection and refine your proposal accordingly.

Results from your last action (if any):

{last_tool_results}
\end{Verbatim}
\end{tcolorbox}

Unlike the template-based version, the template-free version of \ouralgo generates its initial experimental code from scratch.
\begin{tcolorbox}[breakable, colback=orange!5!white, colframe=orange!80!black, title=Initial Experiment Implementation Prompt]
\begin{Verbatim}[breaklines, breaksymbol={}, breaksymbolleft={}, fontsize=\small]
Introduction:
You are an AI researcher who is looking to publish a paper that will contribute significantly to the field."
Your first task is to write a python code to implement a solid baseline based on your research idea provided below, from data preparation to model training, as well as evaluation and visualization.
Focus on getting a simple but working implementation first, before any sophisticated improvements.
We will explore more advanced variations in later stages.

Research idea:
{research_idea}

Memory:
{memory_summary}

Instructions:
Response format:
    Your response should be a brief outline/sketch of your proposed solution in natural language (7-10 sentences), followed by a single markdown code block (using the format ```python ... ```) which implements this solution and prints out the evaluation metric(s) if applicable. There should be no additional headings or text in your response. Just natural language text followed by a newline and then the markdown code block. Make sure to write concise code.
Experiment design sketch guideline:
    This first experiment design should be relatively simple, without extensive hyper-parameter optimization. 
    Take the Memory section into consideration when proposing the design. 
    The solution sketch should be 6-10 sentences.
    Don't suggest to do EDA.
    Make sure to create synthetic data if needed.
Evaluation Metric(s):
    {eval_metrics}
---Prompt-Implementation-Guidelines-Begin---
CRITICAL GPU REQUIREMENTS - Your code MUST include ALL of these:
  - At the start of your code, add these lines to handle GPU/CPU:
    ```python
    device = torch.device('cuda' if torch.cuda.is_available() else 'cpu')
    print(f'Using device: {device}')
    ```
  - ALWAYS move models to device using the `.to(device)` method
  - ALWAYS move input tensors to device using the `.to(device)` method
  - ALWAYS move model related tensors to device using the `.to(device)` method
  - For optimizers, create them AFTER moving model to device
  - When using DataLoader, move batch tensors to device in training loop: `batch = {k: v.to(device) for k, v in batch.items() if isinstance(v, torch.Tensor)}`
CRITICAL MODEL INPUT GUIDELINES:
  - Always pay extra attention to the input to the model being properly normalized,
  - This is extremely important because the input to the model's forward pass directly affects the output, and the loss function is computed based on the output,
For generative modeling tasks, you must:
  - Generate a set of samples from your model
  - Compare these samples with ground truth data using appropriate visualizations
  - When saving plots, always use the 'working_dir' variable that will be defined at the start of the script
  - Make sure to give each figure a unique and appropriate name based on the dataset it represents, rather than reusing the same filename.
Important code structure requirements:
  - Do NOT put any execution code inside 'if __name__ == \"__main__\":' block
  - All code should be at the global scope or in functions that are called from the global scope
  - The script should execute immediately when run, without requiring any special entry point
The code should start with:
  import os
  working_dir = os.path.join(os.getcwd(), 'working')
  os.makedirs(working_dir, exist_ok=True)
The code should be a single-file python program that is self-contained and can be executed as-is.
No parts of the code should be skipped, don't terminate the code execution before finishing the script.
Your response should only contain a single code block.
Be aware of the running time of the code, it should complete within {exec_timeout}.
You can also use the "./working" directory to store any temporary files that your code needs to create.
Data saving requirements:
- Save all plottable data (metrics, losses, predictions, etc.) as numpy arrays using np.save()
- Use the following naming convention for saved files:
```python",
# At the start of your code
experiment_data = {
  'dataset_name_1': {
      'metrics': {'train': [], 'val': []},
      'losses': {'train': [], 'val': []},
      'predictions': [],
      'ground_truth': [],
      # Add other relevant data
  },
  # Add additional datasets as needed:
  'dataset_name_2': {
      'metrics': {'train': [], 'val': []},
      'losses': {'train': [], 'val': []},
      'predictions': [],
      'ground_truth': [],
      # Add other relevant data
  },
}
# During training/evaluation:
experiment_data['dataset_name_1']['metrics']['train'].append(train_metric)
```
- Include timestamps or epochs with the saved metrics
- For large datasets, consider saving in chunks or using np.savez_compressed()
CRITICAL EVALUATION REQUIREMENTS - Your code MUST include ALL of these:
  1. Track and print validation loss at each epoch or at suitable intervals:
     ```python
     print(f'Epoch {{epoch}}: validation_loss = {{val_loss:.4f}}')
     ```
  2. Track and update ALL these additional metrics:
 str(self.evaluation_metrics),
  3. Update metrics at EACH epoch:
  4. Save ALL metrics at the end:
     ```python
     np.save(os.path.join(working_dir, 'experiment_data.npy'), experiment_data)
     ```
Installed Packages: 
Your solution can use any relevant machine learning packages such as: {pkg_str}. Feel free to use any other packages too (all packages are already installed!). For neural networks we suggest using PyTorch rather than TensorFlow.
---Prompt-Implementation-Guidelines-END---
\end{Verbatim}
\end{tcolorbox}

The agentic tree search prompts are shown below.
\begin{tcolorbox}[breakable, colback=orange!5!white, colframe=orange!80!black, title=Agentic Tree Search Prompt]
\begin{Verbatim}[breaklines, breaksymbol={}, breaksymbolleft={}, fontsize=\small]
# For debugging
Introduction:
You are an experienced AI researcher. Your previous code for research experiment had a bug, so based on the information below, you should revise it in order to fix this bug. 
Your response should be an implementation outline in natural language followed by a single markdown code block which implements the bugfix/solution.
Research idea: {research_idea}
Previous (buggy) implementation: {parent_node.code}
Execution output: {execution_output}
Feedback based on generated plots: {parent_node.vlm_feedback_summary}
Feedback about execution time: {parent_node.exec_time_feedback}
Response format:
    Your response should be a brief outline/sketch of your proposed solution in natural language (3-5 sentences), followed by a single markdown code block (using the format ```python ... ```) which implements the full code including the bugfix/solution.
    There should be no additional headings or text in your response. Just natural language text followed by a newline and then the markdown code block.
    Your generated code should be complete and executable. Do not omit any part of the code, even if it was part of a previous implementation.
    Make sure to write concise code.
Bugfix improvement sketch guideline:
    You should write a brief natural language description (3-5 sentences) of how the issue in the previous implementation can be fixed.
    Don't suggest to do EDA.
{prompt_implementation_guideline}

# For identifying issues
Introduction:
    You are an experienced AI researcher. Given the experiment code, numerical metrics, and hypothesis behind the experiment, identify the biggest issue you observe and suggest how to improve it for the next step.
    Never criticize the lack of statistical rigor because we'll take care of that later.
Short Hypothesis: {short_hypothesis}
Experiment code: {parent_node.code}
Numerical Results: {parent_node.metric}
Feedback on Figures: {parent_node.vlm_feedback_summary}
Feedback about execution time: {parent_node.exec_time_feedback}
Response format:
    OBSERVATIONS:
    <your detailed observations here>
    ISSUES:
    - issue 1
    - issue 2 (could be multiple issues)
    FIX_PLAN:
    <your plan here>

# For improving
Introduction: 
    You are an experienced AI researcher. You are provided with a previously developed implementation. Your task is to improve it based on the current experimental stage.
Research idea: {research_idea}
Identified issues: {identified_issues}
Fix plan: {fix_plan}
Previous solution: {parent_node.code}
Instructions:
Response format:
    Your response should be a brief outline/sketch of your proposed solution in natural language (7-10 sentences), followed by a single markdown code block (using the format ```python ... ```) which implements this solution and prints out the evaluation metric(s) if applicable.
    There should be no additional headings or text in your response. Just natural language text followed by a newline and then the markdown code block.
    Make sure to write concise code.
{prompt_implementation_guideline}

# For hyperparameter node
Introduction:
    You are an experienced AI researcher. You are provided with a previously developed baseline implementation. Your task is to implement hyperparameter tuning for the following idea: 
    {hyperparam_idea.name}. 
    {hyperparam_idea.description}
Base code you are working on: parent_node.code
Implementation guideline:
    The code should be a single-file python program that is self-contained and can be executed as-is.
    No parts of the code should be skipped, don't terminate the code execution before finishing the script.
    Data saving requirements:
    - Save all plottable data (metrics, losses, predictions, etc.) as numpy arrays using np.save()
    - Use the following naming convention for saved files:
      ```python
      # At the start of your code
      experiment_data = {
          'hyperparam_tuning_type_1': {
              'dataset_name_1': {
                  'metrics': {'train': [], 'val': []},
                  'losses': {'train': [], 'val': []},
                  'predictions': [],
                  'ground_truth': [],
                  # Add other relevant data
              },
              # Add additional datasets as needed:
          },
          # Add additional hyperparam tuning types as needed
      }
    Make sure to use a filename 'experiment_data.npy' to save the data. Do not use any other filename.

# For ablation node
Introduction:
    You are an experienced AI researcher. You are provided with a previously developed baseline implementation. Your task is to implement the ablation study for the following idea:
    {ablation_idea.name}
    {ablation_idea.description}
Base code you are working on: {parent_node.code}
Implementation guideline:
    The code should be a single-file python program that is self-contained and can be executed as-is.
    No parts of the code should be skipped, don't terminate the code execution before finishing the script.
    Data saving requirements:
    - Save all plottable data (metrics, losses, predictions, etc.) as numpy arrays using np.save()
    - Use the following naming convention for saved files:
      ```python
      # At the start of your code
      experiment_data = {
          'ablation_type_1': {
              'dataset_name_1': {
                  'metrics': {'train': [], 'val': []},
                  'losses': {'train': [], 'val': []},
                  'predictions': [],
                  'ground_truth': [],
                  # Add other relevant data
              },
              # Add additional datasets as needed:
              'dataset_name_2': {
                  'metrics': {'train': [], 'val': []},
                  'losses': {'train': [], 'val': []},
                  'predictions': [],
                  'ground_truth': [],
                  # Add other relevant data
              },
          },
          # Add additional ablation types as needed
      }
    Make sure to use a filename 'experiment_data.npy' to save the data. Do not use any other filename.
\end{Verbatim}
\end{tcolorbox}

For the complex, multi-stage experiments from the tree search, a dedicated plot aggregation step synthesizes final figures.
\begin{tcolorbox}[breakable, colback=orange!5!white, colframe=orange!80!black, title=Plot Aggregation Prompt]
\begin{Verbatim}[breaklines, breaksymbol={}, breaksymbolleft={}, fontsize=\small]
# System prompt
You are an ambitious AI researcher who is preparing final plots for a scientific paper submission.
You have multiple experiment summaries (baseline, research, ablation), each possibly containing references to different plots or numerical insights.
There is also a top-level 'research_idea.md' file that outlines the overarching research direction.
Your job is to produce ONE Python script that fully aggregates and visualizes the final results for a comprehensive research paper.

Key points:
1) Combine or replicate relevant existing plotting code, referencing how data was originally generated (from code references) to ensure correctness.
2) Create a complete set of final scientific plots, stored in 'figures/' only (since only those are used in the final paper).
3) Make sure to use existing .npy data for analysis; do NOT hallucinate data. If single numeric results are needed, these may be copied from the JSON summaries.
4) Only create plots where the data is best presented as a figure and not as a table. E.g. don't use bar plots if the data is hard to visually compare.
5) The final aggregator script must be in triple backticks and stand alone so it can be dropped into a codebase and run.
6) If there are plots based on synthetic data, include them in the appendix.

Implement best practices:
- Do not produce extraneous or irrelevant plots.
- Maintain clarity, minimal but sufficient code.
- Demonstrate thoroughness for a final research paper submission.
- Do NOT reference non-existent files or images.
- Use the .npy files to get data for the plots and key numbers from the JSON summaries.
- Demarcate each individual plot, and put them in separate try-catch blocks so that the failure of one plot does not affect the others.
- Make sure to only create plots that are unique and needed for the final paper and appendix. A good number could be around {MAX_FIGURES} plots in total.
- Aim to aggregate multiple figures into one plot if suitable, i.e. if they are all related to the same topic. You can place up to 3 plots in one row.
- Provide well-labeled plots (axes, legends, titles) that highlight main findings. Use informative names everywhere, including in the legend for referencing them in the final paper. Make sure the legend is always visible.
- Make the plots look professional (if applicable, no top and right spines, dpi of 300, adequate ylim, etc.).
- Do not use labels with underscores, e.g. "loss_vs_epoch" should be "loss vs epoch".
- For image examples, select a few categories/classes to showcase the diversity of results instead of showing a single category/class. Some can be included in the main paper, while the rest can go in the appendix.

Your output should be the entire Python aggregator script in triple backticks.

# Plot aggregator prompt (example user prompt part)
We have three JSON summaries of scientific experiments: baseline, research, ablation.
They may contain lists of figure descriptions, code to generate the figures, and paths to the .npy files containing the numerical results.
Our goal is to produce final, publishable figures.

--- RESEARCH IDEA ---
```
{idea_text}
```

IMPORTANT:
- The aggregator script must load existing .npy experiment data from the "exp_results_npy_files" fields (ONLY using full and exact file paths in the summary JSONs) for thorough plotting.
- It should call os.makedirs("figures", exist_ok=True) before saving any plots.
- Aim for a balance of empirical results, ablations, and diverse, informative visuals in 'figures/' that comprehensively showcase the finalized research outcomes.
- If you need .npy paths from the summary, only copy those paths directly (rather than copying and parsing the entire summary).

Your generated Python script must:
1) Load or refer to relevant data and .npy files from these summaries. Use the full and exact file paths in the summary JSONs.
2) Synthesize or directly create final, scientifically meaningful plots for a final research paper (comprehensive and complete), referencing the original code if needed to see how the data was generated.
3) Carefully combine or replicate relevant existing plotting code to produce these final aggregated plots in 'figures/' only, since only those are used in the final paper.
4) Do not hallucinate data. Data must either be loaded from .npy files or copied from the JSON summaries.
5) The aggregator script must be fully self-contained, and place the final plots in 'figures/'.
6) This aggregator script should produce a comprehensive and final set of scientific plots for the final paper, reflecting all major findings from the experiment data.
7) Make sure that every plot is unique and not duplicated from the original plots. Delete any duplicate plots if necessary.
8) Each figure can have up to 3 subplots using fig, ax = plt.subplots(1, 3).
9) Use a font size larger than the default for plot labels and titles to ensure they are readable in the final PDF paper.


Below are the summaries in JSON:

{combined_summaries_str}

Respond with a Python script in triple backticks.
\end{Verbatim}
\end{tcolorbox}

The template-free version of \ouralgo changes manuscript writing from the incremental Aider-based approach of the template-based version and instead uses a direct LaTeX generation followed by the Reflexion technique ~\citep{shinn2023reflexion}.
\begin{tcolorbox}[breakable, colback=orange!5!white, colframe=orange!80!black, title=Manuscript Writing Prompt (ICBINB workshop specific)]
\begin{Verbatim}[breaklines, breaksymbol={}, breaksymbolleft={}, fontsize=\small]
# System prompt
You are an ambitious AI researcher who is looking to publish a paper to the "I Can't Believe It's Not Better" (ICBINB) Workshop at ICLR 2025.
This workshop aims to highlight real-world pitfalls, challenges, and negative or inconclusive results in deep learning, encouraging open discussion.
You must accurately represent the results of the experiments.
The main paper is limited to {page_limit} pages in single-column format, not counting references. In general, try to use the available space and include all relevant information.
DO NOT USE MORE THAN {page_limit} PAGES FOR THE MAIN TEXT. MINIMIZE THE USAGE OF ITEMIZE OR ENUMERATE. ONLY USE THEM IF THEY ARE ABSOLUTELY NECESSARY AND CONTAIN SUBSTANTIAL INFORMATION.
Ensure that the tables and figures are correctly placed in a reasonable location and format.

- Do not change the overall style which is mandated by the conference. Keep to the current method of including the references.bib file.
- Do not remove the \\graphicspath directive or no figures will be found.
- Do not add `Acknowledgements` section to the paper.

Here are some tips for each section of the paper:

- **Title**:
  - Title should be catchy and informative. It should give a good idea of what the paper is about.
  - Try to keep it under 2 lines.

- **Abstract**:
  - Brief summary highlighting the nature of the challenge or pitfall explored.
  - Concise motivation of why this matters for real-world deployment.
  - This should be one continuous paragraph.

- **Introduction**:
  - Overview of the issue or challenge being explored.
  - Clearly state why this problem is important, especially for practical or real-world contexts.
  - Summarize your contributions or findings: they may include negative results, real-world pitfalls, unexpected behaviors, or partial improvements.

- **Related Work**:
  - Cite relevant papers or approaches that have tackled similar issues or have encountered similar pitfalls.
  - Compare and contrast with your own findings.

- **Background** (optional):
  - Provide necessary technical or domain-specific background if needed.

- **Method / Problem Discussion**:
  - Detail the problem context or the method if it is relevant to highlight the challenges faced.
  - If results are not strictly an improvement, discuss partial successes or lessons learned.

- **Experiments** (if applicable):
  - Present results truthfully according to the data you have. Negative, unexpected, or inconclusive findings are valid contributions for this workshop.
  - Include figures, tables, or real-world examples that illustrate the pitfalls.
  - Include up to 4 figures in the main text. All other figures should be in the appendix.

- **Conclusion**:
  - Summarize the main lessons learned or contributions.
  - Suggest next steps or future directions, highlighting how these insights can help the community avoid or overcome similar issues.

- **Appendix**:
  - Place for supplementary material that did not fit in the main paper.
  - Add more information and details (hyperparameters, algorithms, etc.)  in the supplementary material.
  - Add more plots and tables in the supplementary material. Make sure  that this information is not already covered in the main paper.
  - When checking for duplicate figures, be sure to also review their descriptions to catch cases where different figures convey the same information. For example, one figure might present aggregated training accuracy as a single line plot with a shaded standard deviation (e.g., aggregated_training_accuracy.png), while another (per_seed_training_accuracy.png) shows the same data as three separate line plots.

Ensure you are always writing good compilable LaTeX code. 
Common mistakes that should be fixed include:
- LaTeX syntax errors (unenclosed math, unmatched braces, etc.).
- Duplicate figure labels or references.
- Unescaped special characters: & %
- Proper table/figure closure.
- Do not hallucinate new citations or any results not in the logs.

Ensure proper citation usage:
- Always include references within \begin{{filecontents}}{{references.bib}} ... \end{{filecontents}}, even if they haven't changed from the previous round.
- Use citations from the provided references.bib content.
- Each section (especially Related Work) should have multiple citations.

When returning final code, place it in fenced triple backticks with 'latex' syntax highlighting.

# Writeup prompt (example user prompt part)

Your goal is to write up the following idea:

```markdown
{idea_text}
```

We have the following experiment summaries (JSON):
```json
{summaries}
```

We also have a script used to produce the final plots (use this to see how the plots are generated and what names are used in the legend):
```python
{aggregator_code}
```
Please also consider which plots can naturally be grouped together as subfigures.

Available plots for the writeup (use these filenames):
```
{plot_list}
```

We also have VLM-based figure descriptions:
```
{plot_descriptions}
```

Your current progress on the LaTeX write-up is:
```latex
{latex_writeup}
```

Produce the final version of the LaTeX manuscript now, ensuring the paper is coherent, concise, and reports results accurately. Return the entire file in full, with no unfilled placeholders! This must be an acceptable complete LaTeX writeup, suitable for a 4-page single-column workshop paper. Make sure to use the citations from the references.bib file.

Please provide the updated LaTeX code for 'template.tex', wrapped in triple backticks with "latex" syntax highlighting, like so:

```latex
<UPDATED LATEX CODE>
```
\end{Verbatim}
\end{tcolorbox}

The manuscript undergoes a reflection stage~\citep{shinn2023reflexion}, incorporating feedback from LaTeX linters and VLMs.
\begin{tcolorbox}[breakable, colback=orange!5!white, colframe=orange!80!black, title=Manuscript Writing Reflection Prompt]
\begin{Verbatim}[breaklines, breaksymbol={}, breaksymbolleft={}, fontsize=\small]
Now let's reflect and identify any issues (including but not limited to):
1) Are there any LaTeX syntax errors or style violations we can fix? Refer to the chktex output below.
2) Is the writing clear, and scientifically rigorous for a workshop focusing on real-world pitfalls?
3) Have we included all relevant details from the summaries without hallucinating?
4) Are there short sections (one or two sentences) that could be combined into a single paragraph?
5) Can we use more information and details (hyperparameters, unused figures, etc.) in the supplementary material? Only add information that is not already covered in the main paper.
6) The following figures are available in the folder but not used in the LaTeX: {sorted(unused_figs)}
7) The following figure references in the LaTeX do not match any actual file: {sorted(invalid_figs)}
{reflection_page_info}
chktex results:
```
{check_output}
```
8) Issues identified in the VLM reviews of the images, their captions, and related text discussions. Ensure each caption clearly matches its image content and that there is substantial discussion of each figure in the text.
VLM reviews:
```
{review_img_cap_ref}
```

9) Duplicate figures between main text and appendix. Make sure to remove the duplicate figures from the appendix.
```
{analysis_duplicate_figs}
```

Please provide a revised complete LaTeX in triple backticks, or repeat the same if no changes are needed.
Return the entire file in full, with no unfilled placeholders! This must be an acceptable complete LaTeX writeup.
Do not hallucinate any details! Ensure proper citation usage:
- Always include references within \begin{{filecontents}}{{references.bib}} ... \end{{filecontents}}, even if they haven't changed from the previous round.
- Use citations from the provided references.bib content.
\end{Verbatim}
\end{tcolorbox}

The prompts used for VLM-based figure review are provided below.
\begin{tcolorbox}[breakable, colback=orange!5!white, colframe=orange!80!black, title=VLM Figure Review Prompt]
\begin{Verbatim}[breaklines, breaksymbol={}, breaksymbolleft={}, fontsize=\small]
The abstract of the paper is:

{abstract}

You will be given an image via the vision API. As a careful scientist reviewer, your task is to:
  1. Examine the provided image closely.
  2. Describe in detail what the image shows in a scientific manner.
  3. Critically analyze whether the image content aligns with the given caption:

{caption}

  4. We also have references in the main text that mention the figure:

{main_text_figrefs}

You should:
  - Examine the figure in detail: conclude elements in figures (e.g., name of axis) and describe what information is shown (e.g,. the line of loss decrease monotonically but plateau after X epoch)
  - Suggest any potential improvements or issues in the figure itself (e.g., missing legend, unclear labeling, no meaningful conclusion, mismatch with what the caption claims).
  - Critique the caption: does it accurately describe the figure? Is it too long/short? Does it include a concise takeaway?
  - Review how well the main text references (figrefs) explain the figure: 
  Are they missing? Do they adequately describe the figure's content, context, or purpose?

Finally, respond in the following format:

THOUGHT:
<THOUGHT>

REVIEW JSON:
```json
<JSON>
```
In <JSON>, provide the review in JSON format with the following fields in the order:
- "Img_description": "<Describe the figure's contents here>"
- "Img_review": "<Your analysis of the figure itself, including any  suggestions for improvement>"
- "Caption_review": "<Your assessment of how well the caption matches  the figure and any suggestions>"
- "Figrefs_review": "<Your thoughts on whether the main text references adequately describe or integrate the figure>"

In <THOUGHT>, first, thoroughly reason through your observations, analysis of alignment, and any suggested improvements. It is okay to be very long. Then, provide your final structured output in <JSON>. Make sure the JSON is valid and properly formatted, as it will be parsed automatically.
\end{Verbatim}
\end{tcolorbox}

VLMs also assist in a broader reflection on figure selection and placement during manuscript refinement.
\begin{tcolorbox}[breakable, colback=orange!5!white, colframe=orange!80!black, title=VLM-Assisted Manuscript Figure Reflection Prompt]
\begin{Verbatim}[breaklines, breaksymbol={}, breaksymbolleft={}, fontsize=\small]
The following figures are currently used in the paper: 
{sorted(used_figs)}
The following figures are available in the folder but not used in the 
LaTeX: {sorted(unused_figs)}

{reflection_page_info}

The following is the VLM review on figures:

{review_img_selection}

Please review the figures and make the following changes:
1. For figures that do not add significant value to the paper, move them to the appendix
2. For figures that are not very informative or do not effectively  communicate meaningful patterns, remove them entirely
3. For figures that do not contain subfigures and present sparse information, consider combining them with other related figures
4. Update all relevant text discussions to reflect any changes in figure placement or combination
5. Enhance the scientific analysis of the remaining figures in the text - provide detailed, insightful discussions of their significance and findings

Please ensure all changes maintain scientific rigor and improve the paper's clarity and impact. Be more aggressive with figure selection - move more figures to the  appendix or group them together with other figures if the page limit is already exceeded.

If you believe you are done with reflection, simply say: "I am done".
\end{Verbatim}
\end{tcolorbox}

An example dataset reference prompt used for generalized dataset access is provided below.
\begin{tcolorbox}[breakable, colback=orange!5!white, colframe=orange!80!black, title=The dataset reference prompt for generalized dataset access]
\begin{Verbatim}[breaklines, breaksymbol={}, breaksymbolleft={}, fontsize=\small]
# If you want to use med mnist, you can refer to the following code:
medmnist = load_dataset("albertvillanova/medmnist-v2", "pathmnist")
# >>> medmnist.shape
# {'train': (89996, 2), 'validation': (10004, 2), 'test': (7180, 2)}

# If you want to use EuroSAT, you can refer to the following code:
eurosat = load_dataset("tanganke/eurosat")
# >>> eurosat.shape
# {'train': (21600, 2), 'test': (2700, 2), 'contrast': (2700, 2), 'gaussian_noise': (2700, 2), 'impulse_noise': (2700, 2), 'jpeg_compression': (2700, 2), 'motion_blur': (2700, 2), 'pixelate': (2700, 2), 'spatter': (2700, 2)}

# For MNIST, you can refer to the following code:
mnist = load_dataset("ylecun/mnist")
# >>> mnist.shape
# {'train': (60000, 2), 'test': (10000, 2)}

# For Fashion MNIST, you can refer to the following code:
fashion_mnist = load_dataset("zalando-datasets/fashion_mnist")
# >>> fashion_mnist.shape
# {'train': (60000, 2), 'test': (10000, 2)}

# For CIFAR10, you can refer to the following code:
cifar = load_dataset("uoft-cs/cifar10")
# >>> cifar.shape
# {'train': (50000, 2), 'test': (10000, 2)}

# For IMDB, you can refer to the following code:
imdb = load_dataset("stanfordnlp/imdb")
# >>> imdb.shape
# {'train': (25000, 2), 'test': (25000, 2), 'unsupervised': (50000, 2)}

# For Amazon Polarity Dataset, you can refer to the following code:
amazon_polarity = load_dataset("fancyzhx/amazon_polarity")
# >>> amazon_polarity.shape
# {'train': (3600000, 3), 'test': (400000, 3)}

# For Emotion, you can refer to the following code:
emotion = load_dataset("dair-ai/emotion")
# >>> emotion.shape
# {'train': (16000, 2), 'validation': (2000, 2), 'test': (2000, 2)}

# For silicone, you can refer to the following code:
silicone = load_dataset("eusip/silicone", "dyda_da", trust_remote_code=True)
# >>> silicone.shape
# {'train': (87170, 5), 'validation': (8069, 5), 'test': (7740, 5)}

# For DeepMind Math dataset, you can refer to the following code:
math_examples = load_dataset(
    "deepmind/math_dataset", "algebra__linear_1d", trust_remote_code=True
)
# >>> math_examples.shape
# {'train': (1999998, 2), 'test': (10000, 2)}
\end{Verbatim}
\end{tcolorbox}

\subsubsection{Example idea generated for the ICLR ICBINB workshop experiment}
\label{idea:generated_for_workshop}
\begin{tcolorbox}[breakable, colback=orange!5!white, colframe=orange!80!black, title=Example idea generated for the ICLR ICBINB workshop experiment]
\begin{Verbatim}[breaklines, breaksymbol={}, breaksymbolleft={}, fontsize=\small]
"Name": "compositional_regularization_nn",
"Title": "Enhancing Compositional Generalization in Neural Networks via Compositional Regularization",
"Short Hypothesis": "Introducing a compositional regularization term during training can encourage neural networks to develop compositional representations, thereby improving their ability to generalize to novel combinations of known components.",
"Related Work": "Previous work has highlighted the challenges neural networks face in achieving compositional generalization. Studies such as 'Compositional Generalization through Abstract Representations in Human and Artificial Neural Networks' (Ito et al., NeurIPS 2022) have explored abstract representations to tackle this issue. However, limited research focuses on directly incorporating explicit regularization terms into the training objective to enforce compositional structures. Our proposal distinguishes itself by introducing a novel regularization approach that penalizes deviations from predefined compositional patterns during training, encouraging the network to internalize compositional rules.",
"Abstract": "Neural networks excel in many tasks but often struggle with compositional generalization--the ability to understand and generate novel combinations of familiar components. This limitation hampers their performance on tasks requiring systematic generalization beyond the training data. In this proposal, we introduce a novel training method that incorporates an explicit compositional regularization term into the loss function of neural networks. This regularization term is designed to encourage the formation of compositional representations by penalizing the network when its internal representations deviate from expected compositional structures. We hypothesize that this approach will enhance the network's ability to generalize to unseen combinations, mimicking human-like compositional reasoning. We will test our method on synthetic benchmarks like the SCAN and COGS datasets, which are specifically designed to evaluate compositional generalization, as well as on real-world tasks such as machine translation and semantic parsing. By comparing our method to baseline models and existing approaches, we aim to demonstrate significant improvements in generalization performance. This work offers a new avenue for enforcing compositionality in neural networks through regularization, potentially bridging the gap between neural network capabilities and human cognitive flexibility.",
"Experiments": [
    "Implement the compositional regularization term and integrate it into the loss function of standard sequence-to-sequence neural network architectures with attention mechanisms.",
    "Train models on synthetic datasets like SCAN and COGS, evaluating performance on compositional generalization tasks with and without the regularization term.",
    "Apply the method to real-world tasks such as machine translation using the IWSLT dataset and semantic parsing with the GeoQuery dataset, assessing improvements in generalization to new language constructs.",
    "Analyze the learned representations by visualizing embedding spaces and utilizing compositionality metrics to assess how the regularization affects internal representations.",
    "Conduct ablation studies to determine the impact of different strengths of the regularization term, identifying the optimal balance between enforcing compositionality and maintaining overall performance.",
    "Compare the proposed method against other approaches aimed at improving compositional generalization, such as meta-learning techniques and specialized architectures."
],
"Risk Factors and Limitations": [
    "The effectiveness of the compositional regularization may vary across different datasets and tasks, potentially limiting its generalizability.",
    "An improperly balanced regularization term could negatively impact model performance on the primary task, leading to lower accuracy.",
    "Additional computational overhead from calculating the regularization term may increase training time and resource requirements.",
    "Defining appropriate compositional structures for complex or less-understood domains may be challenging, affecting the applicability of the method.",
    "The approach may face scalability issues when applied to very large models or datasets common in industrial applications."]
\end{Verbatim}
\end{tcolorbox}

\subsubsection{Example Deliverables by Stage (Template-Free Run)}

The research pipeline comprises four stages: a working baseline is first implemented (Stage~1), then refined via hyperparameter tuning (Stage~2). The resulting code seeds the main research exploration (Stage~3), followed by systematic ablation studies (Stage~4). 
In this section, we present example outputs for each stage. To save space, we omit generated code and show only one of ten Stage~4 ablation results.

\textbf{Idea.} Attribute–Jacobian Regularization (AJR) penalizes a model's sensitivity to a differentiable augmentation parameter $\alpha$ that controls a suspected spurious attribute, encouraging predictions to be invariant to that attribute without requiring group labels.

\textbf{Stage~1.} A minimal Colored-MNIST demo applies AJR as a penalty on the loss's sensitivity to a per-example color-mixing coefficient $\alpha$ (interpolating image $\leftrightarrow$ grayscale) and evaluates robustness via worst-group accuracy over (label, color) pairs (20 groups).

\textbf{Stage~2.} The same AJR formulation is used while sweeping hyperparameters (learning rate, batch size, epochs) on both Colored-MNIST and Colored-Fashion-MNIST.

\textbf{Stage~3.} An ``aligned AJR'' variant precomputes two spurious ``views'' per image and interpolates along the exact spurious axis; the system also adds CIFAR-10 with warm/cool tints, employs a stronger CNN with weight decay, a learning-rate scheduler, and gradient clipping, and reports overall accuracy, worst-group accuracy, and the spurious-robust gap.

\textbf{Stage~4.} An ablation study varies the strength of the spurious factor during training via an $\alpha$-schedule, analyzing how schedule design affects robustness.

\begin{tcolorbox}[breakable, colback=orange!5!white, colframe=orange!80!black, title=Example Stage 1 summary]
\begin{Verbatim}[breaklines, breakafter={/,_,-,.}, breaksymbol={}, breaksymbolleft={}, fontsize=\small]
{
  "Experiment_description": "The experiments implement and evaluate Attribute-Jacobian Regularization (AJR) to address spurious correlations in models trained on a Colored-MNIST dataset. Different nodes compare AJR against Empirical Risk Minimization (ERM) to assess robustness and fairness improvements.",
  "Significance": "These experiments are crucial in understanding how AJR can improve model robustness to spurious correlations, a common problem in machine learning. The findings suggest that AJR can enhance worst-group accuracy, indicating better fairness and reduced reliance on non-causal features, with potential implications for more equitable AI systems.",
  "Description": "The experiments involve training simple CNN models on a Colored-MNIST dataset with AJR applied to penalize the model's sensitivity to color augmentations. This regularization aims to reduce reliance on spurious correlations by incorporating a term \u03bb\u2016\u2202\u2113/\u2202\u03b1\u2016\u00b2 into the loss function. The performance is evaluated using standard accuracy and worst-group accuracy metrics.",
  "List_of_included_plots": [
    {
      "path": "experiments/2025-08-03_14-36-08_attribute_jacobian_regularization_attempt_0/logs/0-run/experiment_results/experiment_c6f4e1559a5f4c58a034b1c50d4ca4b8_proc_206780/colored_mnist_group_analysis.png",
      "description": "The per-group accuracy plot shows significant improvements in accuracy consistency across digit-color combinations, though some groups remain challenging.",
      "analysis": "This plot demonstrates AJR's ability to improve fairness across different groups, though dataset imbalance may still affect performance."
    },
    {
      "path": "experiments/2025-08-03_14-36-08_attribute_jacobian_regularization_attempt_0/logs/0-run/experiment_results/experiment_c6f4e1559a5f4c58a034b1c50d4ca4b8_proc_206780/colored_mnist_ajr_effect.png",
      "description": "AJR regularization impact plot shows improved worst-group accuracy and reduced performance gap between standard and worst-group accuracies.",
      "analysis": "This plot confirms that AJR effectively increases model robustness to spurious correlations over time."
    },
    {
      "path": "experiments/2025-08-03_14-36-08_attribute_jacobian_regularization_attempt_0/logs/0-run/experiment_results/experiment_d1b529dfdf8c4ea6966b4b9af7112c94_proc_206782/colored_mnist_combined_metrics_comparison.png",
      "description": "Training loss curves for ERM and AJR show similar convergence, indicating AJR does not add significant optimization difficulty.",
      "analysis": "This plot suggests that AJR is computationally feasible and does not impede model training efficiency."
    },
    {
      "path": "experiments/2025-08-03_14-36-08_attribute_jacobian_regularization_attempt_0/logs/0-run/experiment_results/experiment_d1b529dfdf8c4ea6966b4b9af7112c94_proc_206782/colored_mnist_ajr_results.png",
      "description": "Validation loss shows AJR initially higher but eventually comparable to ERM, indicating potential need for more epochs.",
      "analysis": "AJR's potential requirement for longer training suggests the need for careful tuning and experimentation."
    }
  ],
  "Key_numerical_results": [
    {
      "result": 0.906,
      "description": "Validation accuracy for AJR implementation in Node c6f4e1559a5f4c58a034b1c50d4ca4b8.",
      "analysis": "AJR achieves high validation accuracy, showing its effectiveness in learning despite spurious correlations."
    },
    {
      "result": 0.6957,
      "description": "Worst-group accuracy for AJR in Node c6f4e1559a5f4c58a034b1c50d4ca4b8.",
      "analysis": "Significant improvement in worst-group accuracy highlights AJR's strength in addressing fairness issues."
    },
    {
      "result": 0.9485,
      "description": "Worst-group accuracy for AJR in Node d1b529dfdf8c4ea6966b4b9af7112c94.",
      "analysis": "Performance indicates AJR's potential variability in effectiveness, suggesting importance of hyper-parameter tuning."
    },
    {
      "result": 0.9555,
      "description": "Worst-group accuracy for AJR in Node 707e4d07e5e440cc8d8cc48630f8e542.",
      "analysis": "Higher worst-group accuracy reinforces AJR's potential to improve robustness in certain settings."
    }
  ]
}
\end{Verbatim}
\end{tcolorbox}

\begin{tcolorbox}[breakable, colback=orange!5!white, colframe=orange!80!black, title=Example Stage 2 summary]
\begin{Verbatim}[breaklines, breakafter={/,_,-,.}, breaksymbol={}, breaksymbolleft={}, fontsize=\small]
{
  "best node": {
    "overall_plan": "The overall plan involves implementing a baseline for Attribute-Jacobian Regularization (AJR) to address spurious correlations in synthetic datasets, initially focusing on the Colored-MNIST where digit color acts as a spurious cue. The core approach uses a simple CNN with AJR, applying a differentiable color augmentation parameterized by \u03b1, and includes a regularization term \u03bb\u2016\u2202\u2113/\u2202\u03b1\u2016\u00b2 to encourage prediction invariance to color changes. The effectiveness of AJR is compared against standard ERM training by evaluating worst-group accuracy across digit-color combinations, leveraging automatic differentiation for efficient computation of the Jacobian. Hyperparameter tuning was originally focused on learning rate optimization through grid search. However, the plan now expands to include testing on two datasets, adding Fashion-MNIST with a similar spurious correlation setup, and involves a more systematic hyperparameter search that includes batch size and epoch variations. This aims to address the small improvement margin and enhance the robustness and generalizability of the AJR method. The overall approach now combines a broader testing framework and a comprehensive hyperparameter exploration to achieve improved worst-group accuracy and more robust validation of the AJR technique.",
    "analysis": "",
    "metric": {
      "value": {
        "metric_names": [
          {
            "metric_name": "ERM worst-group accuracy",
            "lower_is_better": false,
            "description": "Worst-group accuracy of the best ERM configuration.",
            "data": [
              {
                "dataset_name": "Colored MNIST",
                "final_value": 0.9436,
                "best_value": 0.9436
              },
              {
                "dataset_name": "Colored Fashion-MNIST",
                "final_value": 0.3152,
                "best_value": 0.3152
              }
            ]
          },
          {
            "metric_name": "AJR worst-group accuracy",
            "lower_is_better": false,
            "description": "Worst-group accuracy of the best AJR configuration.",
            "data": [
              {
                "dataset_name": "Colored MNIST",
                "final_value": 0.9544,
                "best_value": 0.9544
              },
              {
                "dataset_name": "Colored Fashion-MNIST",
                "final_value": 0.3107,
                "best_value": 0.3107
              }
            ]
          },
          {
            "metric_name": "ERM training loss",
            "lower_is_better": true,
            "description": "Training loss of the best ERM configuration.",
            "data": [
              {
                "dataset_name": "Colored MNIST",
                "final_value": 0.0162,
                "best_value": 0.0162
              },
              {
                "dataset_name": "Colored Fashion-MNIST",
                "final_value": 0.1035,
                "best_value": 0.1035
              }
            ]
          },
          {
            "metric_name": "AJR training loss",
            "lower_is_better": true,
            "description": "Training loss of the best AJR configuration.",
            "data": [
              {
                "dataset_name": "Colored MNIST",
                "final_value": 0.0123,
                "best_value": 0.0123
              },
              {
                "dataset_name": "Colored Fashion-MNIST",
                "final_value": 0.0833,
                "best_value": 0.0833
              }
            ]
          },
          {
            "metric_name": "ERM validation loss",
            "lower_is_better": true,
            "description": "Validation loss of the best ERM configuration.",
            "data": [
              {
                "dataset_name": "Colored MNIST",
                "final_value": 0.0637,
                "best_value": 0.0637
              },
              {
                "dataset_name": "Colored Fashion-MNIST",
                "final_value": 0.568,
                "best_value": 0.568
              }
            ]
          },
          {
            "metric_name": "AJR validation loss",
            "lower_is_better": true,
            "description": "Validation loss of the best AJR configuration.",
            "data": [
              {
                "dataset_name": "Colored MNIST",
                "final_value": 0.058,
                "best_value": 0.058
              },
              {
                "dataset_name": "Colored Fashion-MNIST",
                "final_value": 0.5088,
                "best_value": 0.5088
              }
            ]
          }
        ]
      },
      "maximize": null,
      "name": null,
      "description": null
    },
    "plot_plan": "I'll create visualizations based on the hyperparameter tuning experiment for comparing ERM and AJR methods on Colored MNIST and Colored Fashion-MNIST datasets. The experiment data contains training/validation losses, worst-group accuracy metrics, and hyperparameter configurations for both methods across multiple settings. I'll plot training curves for the best configurations, hyperparameter comparison charts, method performance comparisons, and improvement analysis across both datasets.",
    "plot_analyses": [],
    "plot_paths": [
      "experiments/2025-08-03_14-36-08_attribute_jacobian_regularization_attempt_0/logs/0-run/experiment_results/experiment_0d5c2cb046964baca8e1942e7fb58548_proc_213041/hyperparameter_tuning_two_datasets.png",
      "experiments/2025-08-03_14-36-08_attribute_jacobian_regularization_attempt_0/logs/0-run/experiment_results/experiment_0d5c2cb046964baca8e1942e7fb58548_proc_213041/method_comparison_summary_mnist_fashionmnist.png",
      "experiments/2025-08-03_14-36-08_attribute_jacobian_regularization_attempt_0/logs/0-run/experiment_results/experiment_0d5c2cb046964baca8e1942e7fb58548_proc_213041/learning_rate_tuning_results.png",
      "experiments/2025-08-03_14-36-08_attribute_jacobian_regularization_attempt_0/logs/0-run/experiment_results/experiment_0d5c2cb046964baca8e1942e7fb58548_proc_213041/hyperparameter_comparison_mnist_fashionmnist.png",
      "experiments/2025-08-03_14-36-08_attribute_jacobian_regularization_attempt_0/logs/0-run/experiment_results/experiment_0d5c2cb046964baca8e1942e7fb58548_proc_213041/loss_curves_best_configs_mnist_fashionmnist.png",
      "experiments/2025-08-03_14-36-08_attribute_jacobian_regularization_attempt_0/logs/0-run/experiment_results/experiment_0d5c2cb046964baca8e1942e7fb58548_proc_213041/worst_group_accuracy_evolution_mnist_fashionmnist.png"
    ],
    "vlm_feedback_summary": [],
    "exp_results_dir": "experiment_results/experiment_0d5c2cb046964baca8e1942e7fb58548_proc_213041",
    "exp_results_npy_files": [
      "experiment_results/experiment_0d5c2cb046964baca8e1942e7fb58548_proc_213041/experiment_data.npy"
    ]
  },
  ],
}
\end{Verbatim}
\end{tcolorbox}

\begin{tcolorbox}[breakable, colback=orange!5!white, colframe=orange!80!black, title=Example Stage 3 summary]
\begin{Verbatim}[breaklines, breakafter={/,_,-,.}, breaksymbol={}, breaksymbolleft={}, fontsize=\small]
{
  "best node": {
    "overall_plan": "The overall plan is to enhance model preprocessing and architecture for improved generalization on datasets like CIFAR-10 by employing normalization, architectural adjustments, and novel augmentation techniques such as hue-specific strategies and adaptive regularization. Key objectives include redesigning augmentation functions to target spurious attributes, using parameter \u03b1 to control their strength, and applying techniques like dynamic texture generation in Fashion-MNIST, differentiable color tinting in CIFAR-10, and direct color interpolation control in Colored-MNIST. A critical addition involves better spurious correlation testing, proper OOD evaluation, and redefining group definitions. The current plan builds on this by addressing label-dependent augmentation issues with three innovations: label-independent augmentations using fixed attribute transformations, adaptive lambda scheduling that starts with task learning to enforce invariance, and multi-scale sensitivity regularization to capture prediction stability. These enhancements aim to resolve existing challenges and provide deeper insights into AJR's applicability across spurious correlations, contributing to more resilient models.",
    "metric": {
      "value": {
        "metric_names": [
          {
            "metric_name": "overall accuracy",
            "lower_is_better": false,
            "description": "Overall classification accuracy achieved by the model.",
            "data": [
              {
                "dataset_name": "MNIST ERM",
                "final_value": 1.0,
                "best_value": 1.0
              },
              {
                "dataset_name": "MNIST AJR_Adaptive_Linear",
                "final_value": 1.0,
                "best_value": 1.0
              },
              {
                "dataset_name": "MNIST AJR_Adaptive_Cosine",
                "final_value": 1.0,
                "best_value": 1.0
              },
              {
                "dataset_name": "MNIST AJR_MultiScale",
                "final_value": 1.0,
                "best_value": 1.0
              },
              {
                "dataset_name": "FASHION_MNIST ERM",
                "final_value": 1.0,
                "best_value": 1.0
              },
              {
                "dataset_name": "FASHION_MNIST AJR_Adaptive_Linear",
                "final_value": 1.0,
                "best_value": 1.0
              },
              {
                "dataset_name": "FASHION_MNIST AJR_Adaptive_Cosine",
                "final_value": 1.0,
                "best_value": 1.0
              },
              {
                "dataset_name": "FASHION_MNIST AJR_MultiScale",
                "final_value": 1.0,
                "best_value": 1.0
              },
              {
                "dataset_name": "CIFAR10 ERM",
                "final_value": 0.912,
                "best_value": 0.912
              },
              {
                "dataset_name": "CIFAR10 AJR_Adaptive_Linear",
                "final_value": 0.9165,
                "best_value": 0.9165
              },
              {
                "dataset_name": "CIFAR10 AJR_Adaptive_Cosine",
                "final_value": 0.901,
                "best_value": 0.901
              },
              {
                "dataset_name": "CIFAR10 AJR_MultiScale",
                "final_value": 0.8665,
                "best_value": 0.8665
              }
            ]
          },
          {
            "metric_name": "worst-group accuracy",
            "lower_is_better": false,
            "description": "Accuracy on the worst performing subgroup within the dataset.",
            "data": [
              {
                "dataset_name": "MNIST ERM",
                "final_value": 1.0,
                "best_value": 1.0
              },
              {
                "dataset_name": "MNIST AJR_Adaptive_Linear",
                "final_value": 1.0,
                "best_value": 1.0
              },
              {
                "dataset_name": "MNIST AJR_Adaptive_Cosine",
                "final_value": 1.0,
                "best_value": 1.0
              },
              {
                "dataset_name": "MNIST AJR_MultiScale",
                "final_value": 1.0,
                "best_value": 1.0
              },
              {
                "dataset_name": "FASHION_MNIST ERM",
                "final_value": 1.0,
                "best_value": 1.0
              },
              {
                "dataset_name": "FASHION_MNIST AJR_Adaptive_Linear",
                "final_value": 1.0,
                "best_value": 1.0
              },
              {
                "dataset_name": "FASHION_MNIST AJR_Adaptive_Cosine",
                "final_value": 1.0,
                "best_value": 1.0
              },
              {
                "dataset_name": "FASHION_MNIST AJR_MultiScale",
                "final_value": 1.0,
                "best_value": 1.0
              },
              {
                "dataset_name": "CIFAR10 ERM",
                "final_value": 0.8092,
                "best_value": 0.8092
              },
              {
                "dataset_name": "CIFAR10 AJR_Adaptive_Linear",
                "final_value": 0.815,
                "best_value": 0.815
              },
              {
                "dataset_name": "CIFAR10 AJR_Adaptive_Cosine",
                "final_value": 0.7977,
                "best_value": 0.7977
              },
              {
                "dataset_name": "CIFAR10 AJR_MultiScale",
                "final_value": 0.711,
                "best_value": 0.711
              }
            ]
          },
          {
            "metric_name": "spurious-robust gap",
            "lower_is_better": true,
            "description": "Difference between overall accuracy and worst-group accuracy, representing robustness to spurious correlations.",
            "data": [
              {
                "dataset_name": "MNIST ERM",
                "final_value": 0.0,
                "best_value": 0.0
              },
              {
                "dataset_name": "MNIST AJR_Adaptive_Linear",
                "final_value": 0.0,
                "best_value": 0.0
              },
              {
                "dataset_name": "MNIST AJR_Adaptive_Cosine",
                "final_value": 0.0,
                "best_value": 0.0
              },
              {
                "dataset_name": "MNIST AJR_MultiScale",
                "final_value": 0.0,
                "best_value": 0.0
              },
              {
                "dataset_name": "FASHION_MNIST ERM",
                "final_value": 0.0,
                "best_value": 0.0
              },
              {
                "dataset_name": "FASHION_MNIST AJR_Adaptive_Linear",
                "final_value": 0.0,
                "best_value": 0.0
              },
              {
                "dataset_name": "FASHION_MNIST AJR_Adaptive_Cosine",
                "final_value": 0.0,
                "best_value": 0.0
              },
              {
                "dataset_name": "FASHION_MNIST AJR_MultiScale",
                "final_value": 0.0,
                "best_value": 0.0
              },
              {
                "dataset_name": "CIFAR10 ERM",
                "final_value": 0.1028,
                "best_value": 0.1028
              },
              {
                "dataset_name": "CIFAR10 AJR_Adaptive_Linear",
                "final_value": 0.1015,
                "best_value": 0.1015
              },
              {
                "dataset_name": "CIFAR10 AJR_Adaptive_Cosine",
                "final_value": 0.1033,
                "best_value": 0.1033
              },
              {
                "dataset_name": "CIFAR10 AJR_MultiScale",
                "final_value": 0.1555,
                "best_value": 0.1555
              }
            ]
          },
          {
            "metric_name": "average prediction sensitivity",
            "lower_is_better": true,
            "description": "Average sensitivity of predictions to input perturbations.",
            "data": [
              {
                "dataset_name": "MNIST ERM",
                "final_value": 0.0002,
                "best_value": 0.0002
              },
              {
                "dataset_name": "MNIST AJR_Adaptive_Linear",
                "final_value": 0.0002,
                "best_value": 0.0002
              },
              {
                "dataset_name": "MNIST AJR_Adaptive_Cosine",
                "final_value": 0.0001,
                "best_value": 0.0001
              },
              {
                "dataset_name": "MNIST AJR_MultiScale",
                "final_value": 0.0002,
                "best_value": 0.0002
              },
              {
                "dataset_name": "FASHION_MNIST ERM",
                "final_value": 0.0101,
                "best_value": 0.0101
              },
              {
                "dataset_name": "FASHION_MNIST AJR_Adaptive_Linear",
                "final_value": 0.0094,
                "best_value": 0.0094
              },
              {
                "dataset_name": "FASHION_MNIST AJR_Adaptive_Cosine",
                "final_value": 0.0091,
                "best_value": 0.0091
              },
              {
                "dataset_name": "FASHION_MNIST AJR_MultiScale",
                "final_value": 0.0081,
                "best_value": 0.0081
              },
              {
                "dataset_name": "CIFAR10 ERM",
                "final_value": 0.0107,
                "best_value": 0.0107
              },
              {
                "dataset_name": "CIFAR10 AJR_Adaptive_Linear",
                "final_value": 0.0102,
                "best_value": 0.0102
              },
              {
                "dataset_name": "CIFAR10 AJR_Adaptive_Cosine",
                "final_value": 0.0102,
                "best_value": 0.0102
              },
              {
                "dataset_name": "CIFAR10 AJR_MultiScale",
                "final_value": 0.0084,
                "best_value": 0.0084
              }
            ]
          },
          {
            "metric_name": "training time",
            "lower_is_better": true,
            "description": "Total training time taken by the method in seconds.",
            "data": [
              {
                "dataset_name": "MNIST ERM",
                "final_value": 172.29,
                "best_value": 172.29
              },
              {
                "dataset_name": "MNIST AJR_Adaptive_Linear",
                "final_value": 177.2,
                "best_value": 177.2
              },
              {
                "dataset_name": "MNIST AJR_Adaptive_Cosine",
                "final_value": 168.82,
                "best_value": 168.82
              },
              {
                "dataset_name": "MNIST AJR_MultiScale",
                "final_value": 81.14,
                "best_value": 81.14
              },
              {
                "dataset_name": "FASHION_MNIST ERM",
                "final_value": 25.64,
                "best_value": 25.64
              },
              {
                "dataset_name": "FASHION_MNIST AJR_Adaptive_Linear",
                "final_value": 38.5,
                "best_value": 38.5
              },
              {
                "dataset_name": "FASHION_MNIST AJR_Adaptive_Cosine",
                "final_value": 40.52,
                "best_value": 40.52
              },
              {
                "dataset_name": "FASHION_MNIST AJR_MultiScale",
                "final_value": 71.71,
                "best_value": 71.71
              },
              {
                "dataset_name": "CIFAR10 ERM",
                "final_value": 19.51,
                "best_value": 19.51
              },
              {
                "dataset_name": "CIFAR10 AJR_Adaptive_Linear",
                "final_value": 32.95,
                "best_value": 32.95
              },
              {
                "dataset_name": "CIFAR10 AJR_Adaptive_Cosine",
                "final_value": 32.18,
                "best_value": 32.18
              },
              {
                "dataset_name": "CIFAR10 AJR_MultiScale",
                "final_value": 53.36,
                "best_value": 53.36
              }
            ]
          }
        ]
      },
      "maximize": null,
      "name": null,
      "description": null
    },
    "plot_plan": "I'll analyze the enhanced AJR experiment data and create comprehensive visualizations showing the effectiveness of the different innovations across MNIST, Fashion-MNIST, and CIFAR-10 datasets. The experiment tested label-independent augmentations with adaptive lambda scheduling and multi-scale regularization techniques.",
    "plot_analyses": [
      {
        "analysis": "The heatmaps for MNIST, Fashion-MNIST, and CIFAR-10 datasets show minimal or no improvement in WGA (Worst-Group Accuracy), Gap Reduction, and Sensitivity Reduction across the Adaptive Linear, Adaptive Cosine, and Multi-Scale methods compared to the baseline. Notably, CIFAR-10 shows slight negative values in WGA Improvement and Gap Reduction, indicating potential degradation in performance for these metrics.",
        "plot_path": "experiments/2025-08-03_14-36-08_attribute_jacobian_regularization_attempt_0/logs/0-run/experiment_results/experiment_7724d0f4f2d5404a997bf12825e02f2c_proc_260714/innovation_effectiveness_heatmap_all_datasets.png"
      },
      {
        "analysis": "The bar plots comparing Worst-Group Accuracy, Spurious-Robust Gap, Prediction Sensitivity, and Overall Accuracy across datasets reveal that:\n- Worst-Group Accuracy is consistently high for MNIST and Fashion-MNIST, with minimal variation across methods. However, CIFAR-10 shows lower accuracy overall, with Multi-Scale performing slightly better than other methods.\n- Spurious-Robust Gap is lower (better) for Multi-Scale on CIFAR-10, suggesting improved robustness to spurious correlations.\n- Prediction Sensitivity is lowest for MNIST and Fashion-MNIST, indicating that these datasets are less sensitive to augmentation parameters. For CIFAR-10, Multi-Scale slightly reduces sensitivity compared to other methods.\n- Overall Accuracy is high across all datasets and methods, with minimal variation.",
        "plot_path": "experiments/2025-08-03_14-36-08_attribute_jacobian_regularization_attempt_0/logs/0-run/experiment_results/experiment_7724d0f4f2d5404a997bf12825e02f2c_proc_260714/enhanced_ajr_comprehensive_results.png"
      },
      {
        "analysis": "The Worst-Group Accuracy Evolution plots show:\n- MNIST and Fashion-MNIST achieve high accuracy within the first few epochs, with all methods converging similarly.\n- For CIFAR-10, Worst-Group Accuracy improves more gradually, with Adaptive Linear and Multi-Scale showing better trends in later epochs.",
        "plot_path": "experiments/2025-08-03_14-36-08_attribute_jacobian_regularization_attempt_0/logs/0-run/experiment_results/experiment_7724d0f4f2d5404a997bf12825e02f2c_proc_260714/final_performance_comparison_all_metrics.png"
      },
      {
        "analysis": "The Training and Validation Loss Curves indicate:\n- Rapid convergence for MNIST and Fashion-MNIST across all methods, with minimal differences.\n- CIFAR-10 shows slower convergence, with Multi-Scale achieving slightly better validation loss in later epochs, suggesting better generalization.",
        "plot_path": "experiments/2025-08-03_14-36-08_attribute_jacobian_regularization_attempt_0/logs/0-run/experiment_results/experiment_7724d0f4f2d5404a997bf12825e02f2c_proc_260714/worst_group_accuracy_evolution_all_datasets.png"
      },
      {
        "analysis": "The Spurious-Robust Gap Evolution plots demonstrate:\n- MNIST and Fashion-MNIST achieve low gaps early in training, with Multi-Scale slightly outperforming others.\n- For CIFAR-10, the gap fluctuates more, with Multi-Scale showing the best reduction trend over time, indicating improved robustness to spurious correlations.",
        "plot_path": "experiments/2025-08-03_14-36-08_attribute_jacobian_regularization_attempt_0/logs/0-run/experiment_results/experiment_7724d0f4f2d5404a997bf12825e02f2c_proc_260714/enhanced_ajr_loss_curves_all_datasets.png"
      }
    ],
    "plot_paths": [
      "experiments/2025-08-03_14-36-08_attribute_jacobian_regularization_attempt_0/logs/0-run/experiment_results/experiment_7724d0f4f2d5404a997bf12825e02f2c_proc_260714/innovation_effectiveness_heatmap_all_datasets.png",
      "experiments/2025-08-03_14-36-08_attribute_jacobian_regularization_attempt_0/logs/0-run/experiment_results/experiment_7724d0f4f2d5404a997bf12825e02f2c_proc_260714/enhanced_ajr_comprehensive_results.png",
      "experiments/2025-08-03_14-36-08_attribute_jacobian_regularization_attempt_0/logs/0-run/experiment_results/experiment_7724d0f4f2d5404a997bf12825e02f2c_proc_260714/final_performance_comparison_all_metrics.png",
      "experiments/2025-08-03_14-36-08_attribute_jacobian_regularization_attempt_0/logs/0-run/experiment_results/experiment_7724d0f4f2d5404a997bf12825e02f2c_proc_260714/worst_group_accuracy_evolution_all_datasets.png",
      "experiments/2025-08-03_14-36-08_attribute_jacobian_regularization_attempt_0/logs/0-run/experiment_results/experiment_7724d0f4f2d5404a997bf12825e02f2c_proc_260714/enhanced_ajr_loss_curves_all_datasets.png",
      "experiments/2025-08-03_14-36-08_attribute_jacobian_regularization_attempt_0/logs/0-run/experiment_results/experiment_7724d0f4f2d5404a997bf12825e02f2c_proc_260714/spurious_robust_gap_evolution_all_datasets.png"
    ],
    "vlm_feedback_summary": "The plots provide comprehensive insights into the performance of different methods across MNIST, Fashion-MNIST, and CIFAR-10 datasets. While MNIST and Fashion-MNIST show consistent results with minimal variation across methods, CIFAR-10 highlights the strengths of the Multi-Scale approach in reducing spurious correlations and improving robustness. The results emphasize the need for further optimization to address challenges in more complex datasets like CIFAR-10.",
    "exp_results_dir": "experiment_results/experiment_7724d0f4f2d5404a997bf12825e02f2c_proc_260714",
    "exp_results_npy_files": [
      "experiment_results/experiment_7724d0f4f2d5404a997bf12825e02f2c_proc_260714/experiment_data.npy"
    ]
  }
}
\end{Verbatim}
\end{tcolorbox}

\begin{tcolorbox}[breakable, colback=orange!5!white, colframe=orange!80!black, title=Example Stage 4 summary]
\begin{Verbatim}[breaklines, breakafter={/,_,-,.}, breaksymbol={}, breaksymbolleft={}, fontsize=\small]
{
  "overall_plan": "The overall plan integrates a foundational shift in penalty strategies with a detailed exploration of regularization dynamics. Initially, the focus was on implementing a prediction-Jacobian penalty instead of a loss-Jacobian penalty to enhance robustness and understand spurious correlation mitigation across three datasets. Building on this, the current plan introduces an ablation study on regularization lambda adaptation strategies, comparing fixed, adaptive, cyclical, and warmup-decay approaches to refine the training process further. This approach aims to visualize lambda evolution and assess resulting performance metrics, providing insights into optimizing model robustness and performance through penalty and regularization adjustments.",
  "analysis": "",
  "metric": {
    "value": {
      "metric_names": [
        {
          "metric_name": "worst-group accuracy",
          "lower_is_better": false,
          "description": "Accuracy of the worst-performing subgroup within the dataset.",
          "data": [
            {
              "dataset_name": "MNIST",
              "final_value": 0.9652,
              "best_value": 0.9652
            },
            {
              "dataset_name": "FASHION_MNIST",
              "final_value": 0.8182,
              "best_value": 0.8182
            },
            {
              "dataset_name": "CIFAR10",
              "final_value": 0.3704,
              "best_value": 0.3704
            }
          ]
        },
        {
          "metric_name": "spurious-robust gap",
          "lower_is_better": true,
          "description": "Difference in performance between robust and non-robust (spurious) features.",
          "data": [
            {
              "dataset_name": "MNIST",
              "final_value": 0.0098,
              "best_value": 0.0098
            },
            {
              "dataset_name": "FASHION_MNIST",
              "final_value": 0.0289,
              "best_value": 0.0289
            },
            {
              "dataset_name": "CIFAR10",
              "final_value": 0.0724,
              "best_value": 0.0724
            }
          ]
        },
        {
          "metric_name": "overall accuracy",
          "lower_is_better": false,
          "description": "Overall classification accuracy on the test set.",
          "data": [
            {
              "dataset_name": "MNIST",
              "final_value": 0.975,
              "best_value": 0.975
            },
            {
              "dataset_name": "FASHION_MNIST",
              "final_value": 0.8471,
              "best_value": 0.8471
            },
            {
              "dataset_name": "CIFAR10",
              "final_value": 0.4428,
              "best_value": 0.4428
            }
          ]
        },
        {
          "metric_name": "average sensitivity",
          "lower_is_better": true,
          "description": "Average sensitivity of the model with respect to input perturbations.",
          "data": [
            {
              "dataset_name": "MNIST",
              "final_value": 0.0003,
              "best_value": 0.0003
            },
            {
              "dataset_name": "FASHION_MNIST",
              "final_value": 0.005,
              "best_value": 0.005
            },
            {
              "dataset_name": "CIFAR10",
              "final_value": 0.015,
              "best_value": 0.015
            }
          ]
        }
      ]
    },
    "maximize": null,
    "name": null,
    "description": null
  },
  "plot_plan": "I'll create visualizations for the lambda adaptation strategy experiment by analyzing the experiment data and generating plots showing training curves, lambda evolution, and comparative performance across different adaptation strategies.\n\nThe experiment compares four lambda adaptation strategies (Fixed, Adaptive WGA, Cyclical, Warmup-Decay) across three datasets (MNIST, Fashion-MNIST, CIFAR-10). I'll create plots showing training/validation curves, lambda evolution over epochs, worst-group accuracy comparisons, and spurious robustness gap analysis to demonstrate how different adaptive regularization strategies affect model robustness to spurious correlations.",
  "plot_analyses": [
    {
      "analysis": "The results indicate that Fixed lambda=1.0 performs consistently well across datasets in terms of Worst-Group Accuracy (WGA), particularly for MNIST, where it achieves the highest WGA among the strategies. Adaptive WGA and Cyclical lambda strategies show competitive performance but slightly lag behind Fixed lambda=1.0. Warmup Decay performs comparably, with marginal differences from other methods. The Spurious Robustness Gap (SRG), which measures sensitivity to spurious correlations, is lowest for Fixed lambda=1.0 in MNIST and CIFAR10, indicating its robustness. However, for Fashion MNIST, Cyclical lambda achieves the lowest SRG, suggesting it is more effective in mitigating spurious correlations for this dataset. Overall test accuracy remains consistent across strategies, with Fixed lambda=1.0 slightly leading.",
      "plot_path": "experiments/2025-08-03_14-36-08_attribute_jacobian_regularization_attempt_0/logs/0-run/experiment_results/experiment_a4052f5c0c5d48368486ecb38a59e413_proc_406202/lambda_adaptation_comparison.png"
    },
    {
      "analysis": "The Lambda Evolution plots show how the regularization parameter lambda evolves during training for each strategy. Fixed lambda=1.0 maintains a constant value, as expected, ensuring stable regularization throughout training. Adaptive WGA adjusts lambda dynamically, showing a decreasing trend as training progresses, which may help in balancing regularization and model flexibility. Cyclical lambda displays periodic oscillations, which might help avoid local minima but could introduce instability. Warmup Decay starts with a high lambda value, gradually reducing it over epochs, which may facilitate a smooth start to training while reducing over-regularization later.",
      "plot_path": "experiments/2025-08-03_14-36-08_attribute_jacobian_regularization_attempt_0/logs/0-run/experiment_results/experiment_a4052f5c0c5d48368486ecb38a59e413_proc_406202/lambda_adaptation_efficiency_analysis.png"
    },
    {
      "analysis": "The Training Time Comparison highlights that Fixed lambda=1.0 and Adaptive WGA are the most time-efficient strategies, while Cyclical lambda and Warmup Decay incur higher training times due to their dynamic lambda adjustments. The Performance Score Comparison (WGA - SRG) reveals that Fixed lambda=1.0 achieves the best balance between robustness and accuracy for MNIST and Fashion MNIST. For CIFAR10, Cyclical lambda slightly outperforms Fixed lambda=1.0 in terms of the performance score, indicating its effectiveness for this dataset.",
      "plot_path": "experiments/2025-08-03_14-36-08_attribute_jacobian_regularization_attempt_0/logs/0-run/experiment_results/experiment_a4052f5c0c5d48368486ecb38a59e413_proc_406202/lambda_adaptation_robustness_comparison.png"
    },
    {
      "analysis": "The Training vs Validation Loss Curves demonstrate that all strategies converge well for MNIST and Fashion MNIST, with minimal overfitting. For CIFAR10, however, validation loss increases towards the end, particularly for Fixed lambda=1.0 and Adaptive WGA, suggesting potential overfitting or suboptimal regularization. Cyclical lambda and Warmup Decay exhibit more stable validation loss trends for CIFAR10, indicating better generalization.",
      "plot_path": "experiments/2025-08-03_14-36-08_attribute_jacobian_regularization_attempt_0/logs/0-run/experiment_results/experiment_a4052f5c0c5d48368486ecb38a59e413_proc_406202/lambda_adaptation_loss_curves.png"
    },
    {
      "analysis": "The Accuracy Trade-off Analysis plots show that Fixed lambda=1.0 consistently achieves a good balance between overall test accuracy and WGA for MNIST and Fashion MNIST. For CIFAR10, Cyclical lambda achieves slightly better alignment, suggesting it is more effective at improving robustness without sacrificing overall accuracy.",
      "plot_path": "experiments/2025-08-03_14-36-08_attribute_jacobian_regularization_attempt_0/logs/0-run/experiment_results/experiment_a4052f5c0c5d48368486ecb38a59e413_proc_406202/lambda_adaptation_key_results.png"
    }
  ],
  "plot_paths": [
    "experiments/2025-08-03_14-36-08_attribute_jacobian_regularization_attempt_0/logs/0-run/experiment_results/experiment_a4052f5c0c5d48368486ecb38a59e413_proc_406202/lambda_adaptation_comparison.png",
    "experiments/2025-08-03_14-36-08_attribute_jacobian_regularization_attempt_0/logs/0-run/experiment_results/experiment_a4052f5c0c5d48368486ecb38a59e413_proc_406202/lambda_adaptation_efficiency_analysis.png",
    "experiments/2025-08-03_14-36-08_attribute_jacobian_regularization_attempt_0/logs/0-run/experiment_results/experiment_a4052f5c0c5d48368486ecb38a59e413_proc_406202/lambda_adaptation_robustness_comparison.png",
    "experiments/2025-08-03_14-36-08_attribute_jacobian_regularization_attempt_0/logs/0-run/experiment_results/experiment_a4052f5c0c5d48368486ecb38a59e413_proc_406202/lambda_adaptation_loss_curves.png",
    "experiments/2025-08-03_14-36-08_attribute_jacobian_regularization_attempt_0/logs/0-run/experiment_results/experiment_a4052f5c0c5d48368486ecb38a59e413_proc_406202/lambda_adaptation_key_results.png",
    "experiments/2025-08-03_14-36-08_attribute_jacobian_regularization_attempt_0/logs/0-run/experiment_results/experiment_a4052f5c0c5d48368486ecb38a59e413_proc_406202/lambda_evolution_strategies.png",
    "experiments/2025-08-03_14-36-08_attribute_jacobian_regularization_attempt_0/logs/0-run/experiment_results/experiment_a4052f5c0c5d48368486ecb38a59e413_proc_406202/lambda_adaptation_accuracy_tradeoff.png"
  ],
  "vlm_feedback_summary": "The plots and analyses provide valuable insights into the performance of different lambda adaptation strategies. Fixed lambda=1.0 is a strong baseline, showing consistent robustness and efficiency. Adaptive and Cyclical strategies offer competitive performance, with Cyclical lambda excelling in certain robustness scenarios. Warmup Decay provides a smooth training progression but is slightly less efficient. Overall, the results highlight trade-offs between robustness, accuracy, and training efficiency across datasets.",
  "exp_results_dir": "experiment_results/experiment_a4052f5c0c5d48368486ecb38a59e413_proc_406202",
  "ablation_name": "Regularization Lambda Adaptation Strategy",
  "exp_results_npy_files": [
    "experiment_results/experiment_a4052f5c0c5d48368486ecb38a59e413_proc_406202/experiment_data.npy"
  ]
}
\end{Verbatim}
\end{tcolorbox}

\subsubsection{Hyperparameter Configuration for Template-Free Runs}
\label{app:v2_hyperparams}
The template-free version of \ouralgo utilizes the following hyperparameters, detailed in \Cref{tab:v2_hyperparameters_models} and \Cref{tab:v2_hyperparameters_search}.
These include models for code generation, VLM feedback agents, and parameters for the agentic tree search.

\begin{table}[h!]
\centering
\caption{\textbf{LLM and VLM Models and Parameters for the template-free version of \ouralgo.}}
\label{tab:v2_hyperparameters_models}
\begin{tabular}{@{}llcc@{}}
\toprule
\textbf{Component/Task} & \textbf{Model Used} & \textbf{Max Tokens} & \textbf{Temperature} \\
\midrule
Code Generation & Claude 3.5 Sonnet & 8,192 & 0.5 \\
Code Generation & Claude Sonnet 4 & 20,000 & 1.0 \\
LLM/VLM Feedback Agent & GPT-4o & 8,192 & 0.5 \\
Summary Report Agent & GPT-4o & 8,192 & 1.0 \\
Code Critic & o3 & 100,000 & 1.0 \\
\bottomrule
\end{tabular}
\end{table}

\begin{table}[h!]
\centering
\resizebox{\linewidth}{!}{%
\begin{minipage}{1\linewidth}
\caption{\textbf{Agentic Tree Search \& Execution Hyperparameters for the template-free version of \ouralgo.}}
\label{tab:v2_hyperparameters_search}
\begin{tabular}{@{}lcc@{}}
\toprule
\textbf{Hyperparameter} & \textbf{Value for the ICLR experiment} & \textbf{Value for \Cref{fig:conceptual}B} \\
\midrule
Debug Probability & 1.0 & 0.2 \\
Maximum Debug Depth & 3 & 4 \\
Max Experiment Runtime per Node & 1 hour & 1 hour \\
\addlinespace[0.5em]
\multicolumn{2}{l}{\textit{Node Allocation per Stage:}} \\
Stage 1: Preliminary Investigation & 21 nodes & 16 or 20 nodes\\
Stage 2: Hyperparameter Tuning & 12 nodes & 16 or 20 nodes\\
Stage 3: Research Agenda Execution & 12 nodes & 16 nodes \\
Stage 4: Ablation Studies & 12 nodes & 16 nodes\\
\bottomrule
\end{tabular}
\end{minipage}%
}
\end{table}

The total time required for the template-free version of \ouralgo to generate a single paper depends on the complexity of the problems. Based on our experience, this process usually takes anywhere from several hours to a maximum of 15 hours, which is the runtime limit we have set.

For \Cref{fig:conceptual}(B), the following models are used: `gpt-4-0613', `gpt-4o-2024-05-13', `o1-2024-12-17', `claude-3-sonnet-20240229-v1', `claude-sonnet-4-20250514-v1', `gpt-3.5-turbo', `gemini-2.0-flash', `gemini-2.5-pro-preview-06-05', `o3-2025-04-16', `claude-3-5-sonnet-20241022-v2', and `gemini-1.5-pro'. %
Each data point for the template-free runs in the graph represents the mean and standard error calculated over two write-ups for each of three different ideas, totaling six papers. Each data point for the template-based runs represents the mean and standard error calculated over one writeup for each of three different ideas, totaling three papers.

For \Cref{fig:tree_ablations}(C), `claude-sonnet-4-20250514-v1' is used, and several total node budgets are tested: 32 (Stage 3: 16, Stage 4: 16), 16 (Stage 3: 8, Stage 4: 8), 8 (Stage 3: 4, Stage 4: 4), and 4 (Stage 3: 2, Stage 4: 2).
To isolate the impact of different experiment runs, checkpoints are saved after Stage 3 and Stage 4 is re-run multiple times.
In each run, only the first k nodes from Stage 3 are considered (for $k \in \{2, 4, 8\}$), the best node among them is selected, and Stage 4 is initialized from that node, using a Stage 4 node budget of $k$.
For each node setting, 6 papers are generated (2 papers for each of 3 ideas). The \ourreviewer is then run 5 times on each paper, resulting in 30 samples per data point. The mean and standard error across these 30 samples are plotted.

\subsection{The \ourreviewer Details}\label{app:reviewer_details}
\subsubsection{Prompts for Automated Paper Reviewing}
\label{app:v1_prompts_reviewer}
An \ourreviewer component evaluates the generated manuscripts. This component was designed to mimic the peer-review process at a major machine learning conference, using guidelines from NeurIPS. The full validation of this component is described in \Cref{app:v1_reviewer_details}.

The reviewer LLM is first given instructions for its task.
\begin{tcolorbox}[breakable,colback=orange!5!white, colframe=orange!80!black, title=Paper Review System Prompt]
\texttt{You are an AI researcher who is reviewing a paper that was submitted to a prestigious ML venue. 
}
\end{tcolorbox}

The main review prompt provides the paper, guidelines from the NeurIPS reviewer guidelines, and asks for a structured review.
\begin{tcolorbox}[breakable, colback=orange!5!white, colframe=orange!80!black, title=Paper Review Prompt]
\begin{Verbatim}[breaklines, breaksymbol={}, breaksymbolleft={}, fontsize=\small]
## Review Form
Below is a description of the questions you will be asked on the review form for each paper and some guidelines on what to consider when answering these questions.
When writing your review, please keep in mind that after decisions have been made, reviews and meta-reviews of accepted papers and opted-in rejected papers will be made public. 

{neurips_reviewer_guidelines}

{few_shot_examples_of_human_reviews}

Here is the paper you are asked to review:
```
{paper}
```
\end{Verbatim}
\end{tcolorbox}

Multiple reviews are generated for each paper and aggregated into a single meta-review, with an LLM taking the role of the Area Chair.
\begin{tcolorbox}[breakable, colback=orange!5!white, colframe=orange!80!black, title=Paper Review Ensembling System Prompt]
\begin{Verbatim}[breaklines, breaksymbol={}, breaksymbolleft={}, fontsize=\small]
You are an Area Chair at a machine learning conference.
You are in charge of meta-reviewing a paper that was reviewed by {reviewer_count} reviewers.
Your job is to aggregate the reviews into a single meta-review in the same format.
Be critical and cautious in your decision, find consensus, and respect the opinion of all the reviewers.
\end{Verbatim}
\end{tcolorbox}

The ensembling prompt provides the individual reviews for aggregation.
\begin{tcolorbox}[breakable, colback=orange!5!white, colframe=orange!80!black, title=Paper Review Ensembling Prompt]
\begin{Verbatim}[breaklines, breaksymbol={}, breaksymbolleft={}, fontsize=\small]
Review 1/N:
{review_1}

...

Review N/N:
{review_N}

{neurips_reviewer_guidelines}
\end{Verbatim}
\end{tcolorbox}

\subsubsection{Validation Details for the Automated Reviewer}
\label{app:v1_reviewer_details}
As described in the main text, the \ourreviewer component was validated against human decisions to establish its reliability.
A key component of an effective scientific community is its reviewing system, which evaluates and improves the quality of scientific papers.
To mimic such a process, an o4-mini-based agent~\citep{openai2025o3o4mini} was designed to conduct paper reviews based on the Neural Information Processing Systems (NeurIPS) conference \href{https://neurips.cc/Conferences/2022/ReviewerGuidelines}{review guidelines}.

We developed an \ourreviewer that follows the top-tier Neural Information Processing Systems (NeurIPS) conference review guidelines~\citep{neurips2022reviewerguidelines}.
The review pipeline processes the raw text of PDF manuscripts using the PyMuPDF library and produces numerical scores (for soundness, presentation, contribution, overall, and confidence), lists of strengths and weaknesses, and a binary decision (\textit{accept} or \textit{reject}).
We compared the \ourreviewer-generated decisions against ground truth human decisions for ICLR papers, using data from the publicly available OpenReview dataset~\citep{gonzalez2024learning}.
To assess potential data contamination—i.e., whether paper decisions were part of the LLM's training data—we evaluated the \ourreviewer on two sets: 1,000 papers from possibly seen years ($2017$–$2024$) and a ``clean'' $2025$ set, which postdates the training cutoff.
Our reviewer ensemble generates five independent reviews and then produces a meta-review by simulating an Area Chair decision~\citep{wang2022self} (\Cref{app:reviewer_details}). These independent reviews are produced using the same model with identical prompts through multiple sampling runs, and these independent reviews are aggregated into a final meta review using the same model (\Cref{app:v1_prompts_reviewer}).

To compare the \ourreviewer with human performance, we leveraged the NeurIPS 2021 consistency experiment~\citep{beygelzimer2021neurips}.
In that study, roughly 10\% of submissions were randomly duplicated and sent to two fully independent review committees—reviewers and Area Chairs—with no overlap or knowledge of the duplication.
One committee's decision was treated as the ``ground truth,'' and the other committee's decisions were evaluated against it.
By collapsing all accept tiers (Oral, Spotlight, Poster) into a single ``accept'' label and excluding withdrawn papers, human reviewers achieved 73\% accuracy—providing a real-world benchmark of inter-reviewer consistency against which to measure our automated system.
To account for class imbalance, we also report balanced accuracy, defined as the mean of per-class recall for accept and reject. 
Using the human-human confusion matrix from the NeurIPS experiment (Accept = \{Oral, Spotlight, Poster\}; Reject = \{Reject\}), the induced 2$x$2 counts (TP=99, FN=107, FP=96, TN=462) give sensitivity of 99/206 $\approx$ 0.481, specificity of 462/558 $\approx$ 0.828, i.e., a balanced human accuracy of 0.66.

We report six metrics to comprehensively characterize reviewer performance.
Accuracy is simply the fraction of papers for which the reviewer's binary decision (accept vs. reject) matches the ground truth.
Because our accept/reject data are imbalanced, we also compute balanced accuracy by randomly down‐sampling the larger class to match the smaller one, ensuring neither class dominates the score.
F1 score is the harmonic mean of precision (the fraction of predicted accepts that are correct) and recall (the fraction of true accepts that are recovered), which balances false positives and false negatives.
AUC (area under the ROC curve) measures the model's ability to rank accepted papers above rejects across all possible score thresholds.
Finally, we break down errors by reporting the false positive rate (FPR, rejects mislabeled as accepts) and false negative rate (FNR, accepts mislabeled as rejects), giving insight into the types of mistakes the reviewer makes.

We conducted studies to examine the impact of key design choices in \ourreviewer (\Cref{tab:reviewers_appendix}).
Comparing models, o4-mini~\citep{openai2025o3o4mini} achieves the best performance across all metrics, outperforming Claude 3.5 Sonnet~\citep{claude3}, GPT-4o~\citep{gpt4}, and GPT-4.1~\citep{openai_gpt41_2025}.
For prompt variants, we compare our base system prompt (shown in \Cref{app:v1_prompts_reviewer}) with versions augmented with a positive instruction (encouraging acceptance when uncertain) and a negative instruction (encouraging rejection when uncertain).
Neither addition yields meaningful improvements over the base prompt.
We also evaluate several common LLM techniques: VLM, which incorporates visual understanding of figures via vision-language models; Reflexion~\citep{shinn2023reflexion}, which prompts the model to re-evaluate its own prior response for self-correction; and few-shot examples, where we provide 1 or 2 past human reviews as demonstrations in the prompt (provided in Supplementary \Cref{app:v1_prompts_reviewer}.)
Lastly, we apply ensembling, where the LLM generates multiple independent reviews, and a meta-review is produced by aggregating them.
As shown in \Cref{tab:reviewers_appendix}, only the 5-review ensemble consistently improves all metrics.
We therefore adopt o4-mini with the base prompt and a 5-review ensemble as the final configuration of the \ourreviewer.

As shown in \Cref{tab:reviewers_main_paper}, the \ourreviewer achieves a comparable F1 score, accuracy, AUC, and false positive rate to inter-human reviewer agreement from the NeurIPS 2021 consistency study~\citep{beygelzimer2021neurips}, demonstrating strong alignment with human consensus.
These statistics are for aggregate human consensus on the NeurIPS-2021 consistency dataset~\citep{beygelzimer2021neurips}, a benchmark reflecting top-tier \textit{conference} standards. Porting a conference-calibrated reviewer to a \textit{workshop} setting requires prompt (and threshold) recalibration. In our ICLR workshop experiments, the \ourreviewer with its default ``top-tier conference'' prompt rated all three submissions below the acceptance bar for conferences. After explicitly calibrating the prompt for a workshop setting, one of the three—the paper ultimately accepted at the ICLR workshop—was judged above the bar.

To assess whether the observed results are statistically significant, we applied two complementary tests.
First, we performed a two-sample z-test~\citep{lehmann1959ztest} on ordinary accuracy after subsampling each system to an equal number of accepted and rejected papers (\ourreviewer: $n = 698/876$, human: $n = 412$).
Before the knowledge cutoff, the human accuracy was 0.660 versus \ourreviewer at 0.691 ($z = -1.00$, $p = 0.319$); after the cutoff, the values were 0.663 (human) and 0.660 (\ourreviewer) ($z = 0.10$, $p = 0.921$).
These results indicate that the two systems achieve comparable accuracy.
Second, we conducted a non-parametric bootstrap test (5{,}000 replicates) on the full F1 scores to estimate the sampling distribution of $\Delta F1 = F1_{\text{\ourreviewer}} - F1_{\text{human}}$.
Before the cutoff, $\Delta F1 = 0.128$ with a 95\% confidence interval of $[0.060, 0.200]$, $p < 0.001$; after the cutoff, $\Delta F1 = 0.172$ with a confidence interval of $[0.105, 0.241]$, $p < 0.001$.
These results confirm that \ourreviewer achieves higher than the average human reviewer in terms of F1 score in both periods.
However, as we highlighted in the methods section, there is a distribution shift between the two groups of papers, which could explain why the \ourreviewer's F1 score is actually higher than for the human vs. human treatment.

\begin{table}[ht]
\centering
\caption{Ablation results for \ourreviewer showing how the choice of the underlying LLM, prompt variants, and auxiliary techniques (VLM, Reflexion, few-shot examples, ensembling) affect six evaluation metrics: balanced accuracy, accuracy, F1 score, AUC, false positive rate (FPR), and false negative rate (FNR). The combination of o4-mini with a 5-ensemble meta-review achieves the best overall performance.}
\label{tab:reviewers_appendix}
\footnotesize
\setlength{\tabcolsep}{3pt}
\renewcommand{\arraystretch}{1.03}
\begin{tabular}{@{}lcccccc@{}}
\toprule
\bfseries Reviewer
  & \makecell{\bfseries Balanced\\\bfseries Acc.\,($\uparrow$)}
  & \makecell{\bfseries Accuracy\\\bfseries ($\uparrow$)}
  & \makecell{\bfseries F1 Score\\\bfseries ($\uparrow$)}
  & \makecell{\bfseries AUC\\\bfseries ($\uparrow$)}
  & \makecell{\bfseries FPR\\\bfseries ($\downarrow$)}
  & \makecell{\bfseries FNR\\\bfseries ($\downarrow$)} \\
\midrule
Human (NeurIPS)$^{1}$
  & 0.66 & 0.73 & 0.49 & 0.65 & 0.17 & 0.52 \\
Random Decision
  & 0.50 & 0.54 & 0.47 & 0.52 & 0.47 & 0.43 \\
Always Reject
  & 0.50 & 0.65 & 0.00 & 0.50 & 0.00 & 1.00 \\
\midrule
\multicolumn{7}{@{}l}{\textbf{Ablation with different models}} \\
Claude 3.5 Sonnet
  & 0.63 $\pm$ 0.12 & 0.50 $\pm$ 0.10 & 0.57 $\pm$ 0.09
  & 0.61 $\pm$ 0.10 & 0.74 $\pm$ 0.08 & 0.03 $\pm$ 0.04 \\
GPT-4o
  & 0.54 $\pm$ 0.12 & 0.68 $\pm$ 0.09 & 0.20 $\pm$ 0.08
  & 0.54 $\pm$ 0.10 & 0.03 $\pm$ 0.04 & 0.88 $\pm$ 0.06 \\
GPT-4.1
  & 0.58 $\pm$ 0.13 & 0.68 $\pm$ 0.09 & 0.43 $\pm$ 0.10
  & 0.60 $\pm$ 0.10 & 0.15 $\pm$ 0.08 & 0.65 $\pm$ 0.09 \\
\textbf{o4-mini}
  & \textbf{\underline{0.72 $\pm$ 0.12}} & \textbf{\underline{0.65 $\pm$ 0.10}} & \textbf{\underline{0.62 $\pm$ 0.10}}
  & \textbf{\underline{0.70 $\pm$ 0.09}} & \textbf{\underline{0.45 $\pm$ 0.11}} & \textbf{\underline{0.15 $\pm$ 0.07}} \\
\midrule
\multicolumn{7}{@{}l}{\textbf{Ablation with different prompts}} \\
\textbf{Base System Prompt}
  & \textbf{\underline{0.72 $\pm$ 0.10}} & \textbf{\underline{0.65 $\pm$ 0.09}} & \textbf{\underline{0.62 $\pm$ 0.10}}
  & \textbf{\underline{0.70 $\pm$ 0.10}} & \textbf{\underline{0.45 $\pm$ 0.11}} & \textbf{\underline{0.15 $\pm$ 0.07}} \\
+ Positive Instruction
  & 0.56 $\pm$ 0.12 & 0.42 $\pm$ 0.10 & 0.54 $\pm$ 0.10
  & 0.56 $\pm$ 0.09 & 0.88 $\pm$ 0.07 & 0.00 $\pm$ 0.00 \\
+ Negative Instruction
  & 0.73 $\pm$ 0.12 & 0.74 $\pm$ 0.09 & 0.61 $\pm$ 0.10
  & 0.70 $\pm$ 0.09 & 0.18 $\pm$ 0.08 & 0.41 $\pm$ 0.10 \\
\midrule
\multicolumn{7}{@{}l}{\textbf{Ablation with different techniques}} \\
Base
  & 0.72 $\pm$ 0.10 & 0.65 $\pm$ 0.09 & 0.62 $\pm$ 0.10
  & 0.70 $\pm$ 0.10 & 0.45 $\pm$ 0.11 & 0.15 $\pm$ 0.07 \\
+ VLM
  & 0.69 $\pm$ 0.12 & 0.63 $\pm$ 0.09 & 0.46 $\pm$ 0.10
  & 0.71 $\pm$ 0.09 & 0.42 $\pm$ 0.10 & 0.16 $\pm$ 0.07 \\
+ 1 Reflexion
  & 0.72 $\pm$ 0.13 & 0.64 $\pm$ 0.09 & 0.61 $\pm$ 0.09
  & 0.69 $\pm$ 0.09 & 0.47 $\pm$ 0.10 & 0.16 $\pm$ 0.08 \\
+ 2 Reflexions
  & 0.70 $\pm$ 0.11 & 0.68 $\pm$ 0.09 & 0.64 $\pm$ 0.09
  & 0.72 $\pm$ 0.09 & 0.39 $\pm$ 0.10 & 0.18 $\pm$ 0.08 \\
\makecell[l]{+ 1 Few Shot\\Example}
  & 0.73 $\pm$ 0.10 & 0.66 $\pm$ 0.10 & 0.65 $\pm$ 0.10
  & 0.72 $\pm$ 0.09 & 0.47 $\pm$ 0.10 & 0.15 $\pm$ 0.07 \\
\makecell[l]{+ 2 Few Shot\\Examples}
  & 0.68 $\pm$ 0.12 & 0.64 $\pm$ 0.09 & 0.60 $\pm$ 0.10
  & 0.68 $\pm$ 0.10 & 0.44 $\pm$ 0.10 & 0.21 $\pm$ 0.08 \\
+ 3 Ensemble
  & 0.66 $\pm$ 0.12 & 0.66 $\pm$ 0.10 & 0.63 $\pm$ 0.11
  & 0.71 $\pm$ 0.09 & 0.43 $\pm$ 0.10 & 0.15 $\pm$ 0.07 \\
\makecell[l]{\textbf{+ 5 Ensemble}\\\textbf{(final)}}
  & \textbf{\underline{0.77 $\pm$ 0.11}} & \textbf{\underline{0.69 $\pm$ 0.09}} & \textbf{\underline{0.66 $\pm$ 0.09}}
  & \textbf{\underline{0.74 $\pm$ 0.08}} & \textbf{\underline{0.42 $\pm$ 0.10}} & \textbf{\underline{0.09 $\pm$ 0.06}} \\
\bottomrule
\end{tabular}
\end{table}

While the current reviewer evaluation focuses on final manuscripts, incorporating intermediate outputs (e.g., experiment logs) could further enrich reviews. 
Our current design intentionally mirrors how human reviewers typically operate by assessing the final paper rather than each intermediate stage to maintain a fair basis for comparison between human and automated evaluations.

At present, the \ourreviewer relies on the background knowledge embedded in the LLM, without performing explicit literature searches. 
Integrating retrieval and citation capabilities into future versions of the \ourreviewer is a promising direction for future research.

\subsection{Human Workshop Evaluation: Paper Generation, Paper Selection, \& Failure Modes}
\label{app:filtering}

\paragraph{Paper Generation Process}
The process began with the idea generation phase, prompted by the I Can't Believe It's Not Better (ICBINB) workshop's theme of challenges and limitations in applied deep learning, particularly focusing on real-world failures and their underlying causes~\citep{iclr2025icbinb}. 
The system generated a pool of potential research ideas in core machine learning and applied machine learning (\Cref{tab:app_idea_exp_paper_stats}).
Manual inspection of the ideas generated for this workshop reveals that the ideas that \ouralgo generates specifically for this workshop relate to studying why ML methods in the past have not worked well and how to improve them and/or better understand those negative results and why the methods were not better. (See \Cref{idea:generated_for_workshop} for more details.)

\paragraph{Paper Selection Guidelines for Workshop Submissions}

The selection process was governed by a three‐step selection procedure. Had this filtering not occurred, the papers under analysis would still have been produced, just along with other papers and thus at a greater total cost. 
First, each manuscript underwent an \textbf{Idea Alignment} assessment: human reviewers checked whether the core research questions aligned closely with the workshop’s call for papers and whether the proposed ideas were realistically achievable using current LLM capabilities. 
Second, the \textbf{Experimental Execution} phase scrutinized the implementation of experiments. Experimental attempts were filtered out if the code had errors when it ran, if the experimental design did not faithfully reflect the hypotheses, or the attempt did not produce at least one successful result node at stages 1 and 2 (Figure \ref{fig:tree_ablations}).
Third, during a \textbf{Paper Polish} stage, drafts were inspected for citation integrity (e.g., checking for ``[?]'' placeholders), visual completeness (ensuring all figures were properly embedded rather than referenced via raw file paths), adherence to page limits, clarity of figure legends, and avoidance of duplicated graphics between the main text and appendices. After these three phases, three papers remained, each demonstrating conceptual soundness, experimental rigor, and professional presentation. These three papers were entirely generated by \ouralgo without any human editing.

\Cref{tab:app_idea_exp_paper_stats} details the numbers at each stage of the overall process: the total number of ideas generated, the number of experimental attempts per idea, and the number of write-ups produced from which final papers were selected. Importantly, while such filtering saves resources and speaks to the reliability of the current pipeline, even without it the papers selected would have been produced (alongside many other lower-quality outputs). Each paper itself was fully AI-generated. 

Since this year's ICBNB workshop encourages papers on real-world applications of deep learning, we submitted one applied-domain paper, alongside two from core machine learning. Selecting applied ideas requires more care, as real-world datasets often need to be manually prepared and are typically unavailable on platforms like HuggingFace. To address this, we chose several real-world datasets from Kaggle, such as a pest detection dataset and a chest X-ray dataset, due to their real-world relevance and ease of use. To better steer idea generation toward real-world use cases, we modified the prompt to emphasize domains such as finance, psychology, agriculture, environmental science, and public health. Out of 136 generated applied-domain ideas, we selected one that aligned well with the pest detection dataset.

\begin{table}[!h]
\centering
\caption{\textbf{Generation and Write-up Statistics. *Rows 1 and 2 share the same pool of 25 generated ideas. The symbol $\to$ denotes ``Generated $\to$ Selected'' for the number of ideas, experiments, and write-ups.}}
\label{tab:app_idea_exp_paper_stats}
\begin{tabularx}{\columnwidth}{@{}c|c|c|c|c@{}}
\toprule
\textbf{Final Paper Selected} & \textbf{Topic Area} & \textbf{\# Ideas} & \textbf{\# Experiments} &  \textbf{\# Write-Ups} \\ \midrule
Compositional Regularization   & Core ML &  25* $\to$ 1  & 8 $\to$ 1 & 24 $\to$ 1 \\
Label Noise Calibration  & Core ML & 25* $\to$ 1 & 6 $\to$ 1 & 7 $\to$ 1  \\
Pest Detection  & Applied ML & 136 $\to$ 1 & 16 $\to$ 1 & 6 $\to$ 1 \\ \bottomrule
\end{tabularx}
\end{table}

\paragraph{Paper Failure Modes}

\begin{figure}[h!]
\centering
\includegraphics[width=0.9\textwidth]{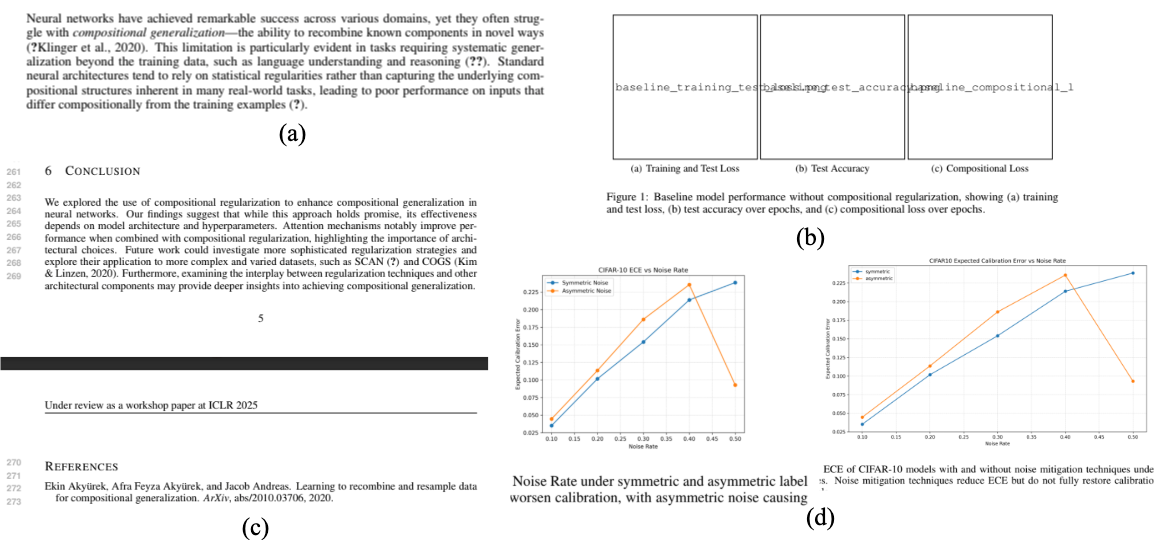}
\caption{Examples paper writeup failures: (a) citation errors, (b) figure display errors, (c) violations of paper length expectations (the workshop limits papers to four pages), and (d) duplicated figures between the main text and appendix.}
\label{fig:paper_failure_examples}
\end{figure}

Representative failure modes (Fig. \ref{fig:paper_failure_examples}) include: citation errors due to mismatched citation keys in LaTeX; figure display errors caused by incorrect figure path references; papers that are shorter or longer than what the workshop expected (a 4-page paper); and duplicated figures between the main text and the appendix.
These cases were filtered out by the guidelines described above.

\paragraph{Comparison to Related Experiments}

Two concurrent works conducted similar experiments that involved submitting AI-generated works to human peer review: Carl~\citep{autoscience2025carl} and Zochi~\citep{intology2025zochi}. However, these experiments differ in important ways and do not reach the same level of automation as \ouralgo. For example, humans wrote the related works section and manually edited several sections of the papers submitted by Carl. Furthermore, Carl submitted works to the ``Tiny Papers'' track (a track with significantly lower standards than a typical workshop paper track, including the workshop that \ouralgo’s paper was accepted to), and each of Carl's papers received, on average, a rejection recommendation by reviewers. Similarly, Zochi's paper generation process is framed as a ``human-AI collaboration'' as opposed to end-to-end automation and required substantial human intervention during paper generation. This involved steering the system's experiments and manually creating publication-quality figures.

In contrast to Carl and Zochi, \ouralgo represents a greater degree of end-to-end automation. The entire workflow for each paper, from the initial idea to experimentation to the final compiled PDF, is completed autonomously without any human modification of the final paper output.
Additionally, both of their systems are not openly available; therefore, it is hard to judge how often other types of human intervention were required.
While these systems are interesting examples of human-AI collaboration, \ouralgo demonstrates a more complete automation of the scientific process itself, which is the central advance of our work.

\clearpage

\section{Supplementary Tables}
\label{app:tables}

\subsection{Template-Based AI Scientist Results}
\label{app:template_based_tables}

The template-based version of \ouralgo was evaluated on three templates across several different publicly available LLMs: Claude Sonnet 3.5~\citep{claude3}, GPT-4o~\citep{gpt4}, DeepSeek Coder~\citep{zhu2024deepseek}, and Llama-3.1 405B~\citep{llama3}.
The first two models are only available via a public API, whilst the second two models are open-weight.
For each run, the system was provided with 1-2 basic seed ideas related to the template as examples (e.g., modifying the learning rate or batch size, see \Cref{app:idea_progression}) and tasked with generating another 50 new ideas.
The results detail the number of ideas, the number of successfully compiled papers, the mean and max \ourreviewer scores of the generated papers, and the total cost of each run, which was approximately \$10-15 per paper
(\Cref{tab:app_diff_papers,tab:app_nlp_papers,tab:app_grokking_papers}).
Each run in \textit{total} took approximately 12 hours on $8\times$ NVIDIA H100s.

\begin{figure}[!h]
\centering
\includegraphics[width=0.975\textwidth]{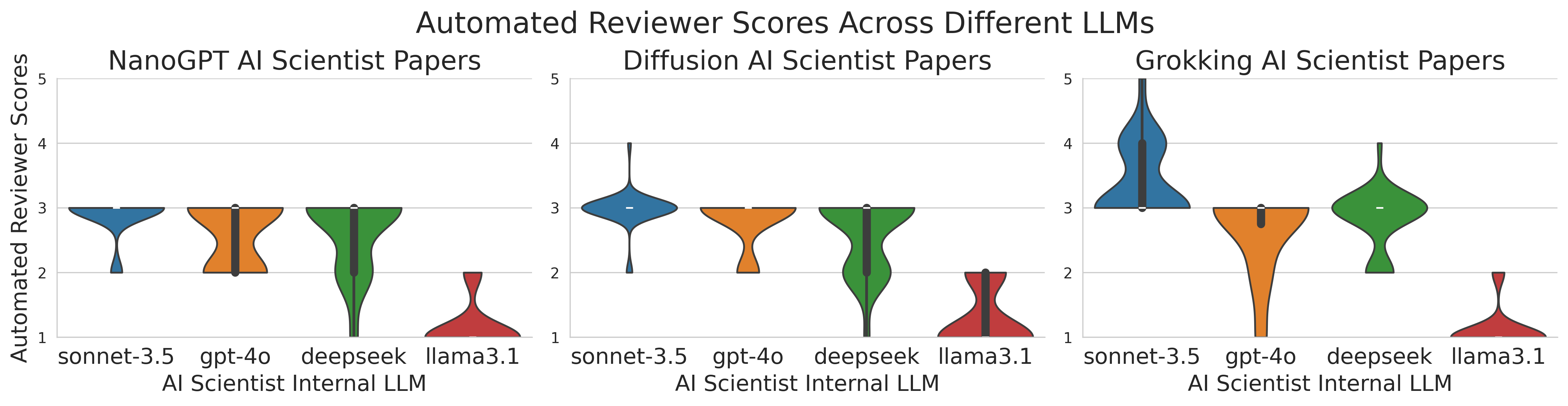}
\caption{\textbf{Distribution of automated review scores for papers generated by the template-based version of \ouralgo.} Violin plots show the distribution of overall scores assigned by the \ourreviewer to papers generated across three domains (Diffusion Modeling, Language Modeling and Grokking Analysis) and four foundation models. Scores on the y-axis refer to \href{https://neurips.cc/Conferences/2022/ReviewerGuidelines}{NeurIPS ratings}~\citep{neurips2022reviewer}, which range from 1 (Very Strong Reject) to 5 (Borderline Accept). Although there is no exact mapping between conference and workshop scores, papers rated 4 ((Borderline Reject)) in conference reviews are generally considered borderline-eligible for workshop acceptance.} 
\label{fig:app_scores_ai_papers}
\end{figure}

From manual inspection, Claude Sonnet 3.5 consistently produces the highest quality papers, with GPT-4o coming in second.
This observation is validated by the scores from the \ourreviewer (\Cref{fig:app_scores_ai_papers}).
GPT-4o notably struggles with writing LaTeX, which prevents it from completing many of its papers.
For the open-weight models, DeepSeek Coder is significantly cheaper but often fails to correctly interface with Aider (i.e., produce correct automatically parsable code), while Llama-3.1 405B performed the worst overall.

Below, the aggregated results for each of the three templates are presented. Full details of selected generated papers for each template, including their ideas and reviews, are provided in \Cref{app:data}.

\subsubsection*{Diffusion Modeling}
\textbf{General Description:} This template studies improving the performance of diffusion generative models~\citep{pmlr-v37-sohl-dickstein15, ddpm} on low-dimensional datasets. Compared to image generation which is high-dimensional, low-dimensional diffusion is less well-studied, providing opportunities for interesting algorithmic contributions.

\textbf{Code Template:} This template is based on a modified version of the popular \texttt{`tanelp/tiny-diffusion'} repository~\citep{tiny_diffusion} with small changes found to improve performance in initial preliminary experiments: minor hyperparameter tuning and exponential moving average on the weights.
The diffusion models are DDPMs~\citep{ddpm} trained separately to generate samples across four 2D distributions.
The denoiser network is an MLP with sinusoidal embeddings for the diffusion timestep and input data.
The initial seed plotting script visualizes generated samples and training loss by default.
Estimated KL divergence is provided as an additional metric for sample quality via non-parametric entropy estimation.

\begin{table}[!h]
\centering
\caption{\textbf{Performance of the template-based version of \ouralgo for Diffusion Modeling experiments.} Metrics include total ideas generated, completed papers, mean and max automated review scores (NeurIPS scale), and approximate total cost in USD.}
\label{tab:app_diff_papers}
\begin{tabular}{@{}l|c|c|c|c|c@{}}
\toprule
\textbf{LLM} & \textbf{Total} & \textbf{Compiled} & \textbf{Mean} & \textbf{Max} & \textbf{Total} \\
 & \textbf{Ideas} & \textbf{Papers} & \textbf{Score} & \textbf{Score} & \textbf{Cost (\$)} \\ \midrule
Claude Sonnet 3.5     & 51 & 37 & 2.97 & 4.0 & $\sim$250 \\
GPT-4o         & 51 &  16 & 2.81 & 3.0 & $\sim$300 \\
DeepSeek Coder & 51 &  28 & 2.61 & 3.0 & $\sim$10   \\
Llama-3.1 405B & 51 &  20 & 1.3 & 2.0 & $\sim$120 \\
\bottomrule
\end{tabular}
\end{table}

\subsubsection*{Language Modeling}
\textbf{General Description:} This template investigates transformer-based~\citep{vaswani2017attention} autoregressive next-token prediction tasks. Because this task is widely studied and optimized, it is difficult for \ouralgo to find significant improvements. A common failure mode for this template is ``cheating'' by subtly leaking information from future tokens to achieve deceptively impressive perplexity.

\textbf{Code Template:} The code is modified from the popular NanoGPT repository~\citep{karpathy2022nanogpt}. The provided script template trains a small transformer language model on the character-level Shakespeare dataset~\citep{char_shakespeare}, the enwik8 dataset~\citep{hutter_prize}, and the text8 dataset~\citep{text8}. The plotting script visualizes training curves by default.

\begin{table}[!h]
\centering
\caption{\textbf{Performance of the template-based version of \ouralgo for Language Modeling experiments.} Metrics as in \Cref{tab:app_diff_papers}.}
\label{tab:app_nlp_papers}
\begin{tabular}{@{}l|c|c|c|c|c@{}}
\toprule
\textbf{LLM} & \textbf{Total} & \textbf{Compiled} & \textbf{Mean} & \textbf{Max} & \textbf{Total} \\
 & \textbf{Ideas} & \textbf{Papers} & \textbf{Score} & \textbf{Score} & \textbf{Cost (\$)} \\ \midrule
Claude Sonnet 3.5     & 52 & 19 & 2.89 & 3.0 & $\sim$250 \\
GPT-4o         & 52 &  15 & 2.60 & 3.0 & $\sim$300 \\
DeepSeek Coder & 52 &  19 & 2.63 & 3.0 & $\sim$10   \\
Llama-3.1 405B & 52 &  19 & 1.16 & 2.0 & $\sim$120 \\
\bottomrule
\end{tabular}
\end{table}

\subsubsection*{Grokking Analysis}
\textbf{General Description:} This template investigates questions about generalization and learning speed, focusing on ``grokking''~\citep{power2022grokking}, a phenomenon where validation accuracy dramatically improves in a short amount of time long after train loss saturates. The code generates synthetic datasets of modular arithmetic tasks and trains a transformer \cite{vaswani2017attention} on them. Unlike other templates, this one is more amenable to open-ended empirical analysis rather than just performance optimization, though it is still constrained to one topic (grokking), unlike the template-free version of \ouralgo.

\textbf{Code Template:} The implementation is based on popular open-source re-implementations~\citep{snell2021grokking, may2022grokking} of Power et~al.~\citep{power2022grokking}. The code generates four synthetic datasets of modular arithmetic tasks and trains a transformer on each. It returns train/validation losses and the number of steps to reach perfect validation accuracy. The initial seed plotting scripts visualize the training and validation curves.

\begin{table}[!h]
\centering
\caption{\textbf{Performance of the template-based version of \ouralgo for Grokking Analysis experiments.} Metrics as in \Cref{tab:app_diff_papers}.}
\label{tab:app_grokking_papers}
\begin{tabular}{@{}l|c|c|c|c|c@{}}
\toprule
\textbf{LLM} & \textbf{Total} & \textbf{Compiled} & \textbf{Mean} & \textbf{Max} & \textbf{Total} \\
 & \textbf{Ideas} & \textbf{Papers} & \textbf{Score} & \textbf{Score} & \textbf{Cost (\$)} \\ \midrule
Claude Sonnet 3.5     & 51 & 25 & 3.4 & 5.0 & $\sim$250 \\
GPT-4o         & 51 &  12 & 2.67 & 3.0 & $\sim$300 \\
DeepSeek Coder & 51 &  32 & 2.81 & 4.0 & $\sim$10   \\
Llama-3.1 405B & 51 &  27 & 1.11 & 2.0 & $\sim$120 \\
\bottomrule
\end{tabular}
\end{table}

\clearpage

\section{Supplementary Discussion}
\label{app:discussion}

\subsection{In-Depth Case Study (Template-Based): ``Adaptive Dual-Scale Denoising''}
\label{app:case_study_v1}
This section presents a representative sample from a run of the template-based version of \ouralgo to illustrate both its strengths and its shortcomings. 
The output, a paper titled ``Adaptive Dual-Scale Denoising'', was generated using the diffusion modeling template with Claude Sonnet 3.5~\citep{claude3}.

\paragraph{Generated Idea}
As discussed in the main text, \ouralgo first generates an idea based on the provided template and the archive of discoveries generated up until that point in the run.
The idea in the selected paper was proposed in the 6th round of ideation of the algorithm and aims to improve the ability of diffusion models to capture both global structure and local details in a 2D dataset, by proposing two branches in the standard denoiser network. This ability of diffusion models has been the primary reason for researchers adopting diffusion models over prior styles of generative models such as VAEs~\citep{vae} and GANs~\citep{gan}, and thus is a well-motivated idea.
To the best of our knowledge, this precise idea has not been widely studied.

Notably, \ouralgo generates an impressive experimental plan that includes \emph{the proposed code modification, comparisons to baselines, evaluation metrics, and the design of additional plots}.

\begin{tcolorbox}[
  breakable,
  colback  = blue!5!white,
  colframe = blue!75!black,
  title    = {Idea - \texttt{adaptive\_dual\_scale\_denoising}},
  fontupper=\scriptsize\ttfamily,
]
\begin{Verbatim}
"Name": "adaptive_dual_scale_denoising",
"Title": "Adaptive Dual-Scale Denoising for Dynamic Feature Balancing in Low-Dimensional Diffusion Models",
"Experiment": "Modify MLPDenoiser to implement a dual-scale processing approach with two parallel branches: a global branch for the original input and a local branch for an upscaled input. Introduce a learnable, timestep-conditioned weighting factor to dynamically balance the contributions of global and local branches. Train models with both the original and new architecture on all datasets. Compare performance using KL divergence and visual inspection of generated samples. Analyze how the weighting factor evolves during the denoising process and its impact on capturing global structure vs. local details across different datasets and timesteps.",
"Interestingness": 9,
"Feasibility": 8,
"Novelty": 8,
"novel": true
\end{Verbatim}
\end{tcolorbox}

\paragraph{Generated Experiments}
The generated code diff (deletions are in \textcolor{BrickRed}{\textbf{red}}, and additions are in \textcolor{OliveGreen}{\textbf{green}}) for the substantial algorithmic changes is displayed below. The new code matches the experimental description and is well-commented. \ouralgo is able to iterate on the code with results from intermediate experiments in the loop, and it eventually ends up with interesting design choices for the adaptive weight network, e.g., a LeakyReLU. Importantly, this network has a well-behaved output that is guaranteed to be between 0 and 1. Additionally, \ouralgo changed the output of the forward function to return the adaptive weights to make new visualizations.

\begin{minted}{diff}
--- a/experiment.py
+++ b/experiment.py
@@ -10,7 +10,7 @@
 from tqdm.auto import tqdm
 from multiprocessing import Pool
 
-
+# Denoising model
 @torch.no_grad()
 def p_sample(model, x, t, t_index):
     betas_t = extract(betas, t, x.shape)
@@ -40,11 +40,31 @@
         self.time_dim = time_dim
 
         self.time_mlp = nn.Sequential(
-                SinusoidalPositionEmbeddings(time_dim),
-                nn.Linear(time_dim, time_dim * 4),
-                nn.GELU(),
-                nn.Linear(time_dim * 4, time_dim),
-            )
+            SinusoidalPositionEmbeddings(time_dim),
+            nn.Linear(time_dim, time_dim * 4),
+            nn.GELU(),
+            nn.Linear(time_dim * 4, time_dim),
+        )
+
+        # Global feature extractor
+        self.global_mlp = nn.Sequential(
+            nn.Linear(2 + time_dim, 256),
+            nn.ReLU(),
+            nn.Linear(256, 2)
+        )
+        # Local feature extractor
+        self.local_mlp = nn.Sequential(
+            nn.Linear(256 + time_dim, 256),
+            nn.ReLU(),
+            nn.Linear(256, 2)
+        )
+        # Adaptive weight network
+        self.adaptive_weight = nn.Sequential(
+            nn.Linear(time_dim, 128),
+            nn.LeakyReLU(0.2),
+            nn.Linear(128, 1),
+            nn.Sigmoid()
+        )
 
         self.main = nn.Sequential(
             nn.Linear(time_dim + 2, 256),
@@ -52,14 +72,21 @@
             nn.Linear(256, 256),
             nn.ReLU(),
             nn.Linear(256, 2),
-        )
+        )
+        self.upscale = nn.Linear(2, 256)
 
-    def forward(self, x, timestep):
+    def forward(self, x, timestep): # x is (batch, 2)
         t = self.time_mlp(timestep)
-        x = torch.cat((x, t), dim=1)
-        return self.main(x)
+        # Get adaptive weight
+        w = self.adaptive_weight(t)
+        # Global branch
+        global_out = self.global_mlp(torch.cat([x, t], dim=1))
+        # Local branch
+        x_upscaled = self.upscale(x)
+        local_out = self.local_mlp(torch.cat([x_upscaled, t], dim=1))
+        # Combine outputs
+        return w * global_out + (1-w) * local_out, w.mean()
 
-
 # KL divergence between two Gaussians
 def kl_divergence(p_mean, p_std, q_mean, q_std):
     # ...
\end{minted}

\paragraph{Generated Paper and Analysis}
A preview of the completely AI-generated paper is displayed in \Cref{fig:case_study_pdf_preview}, with the full-sized version available in the Supplementary Data section.

\begin{figure}[h!]
\centering
\includegraphics[width=0.975\textwidth]{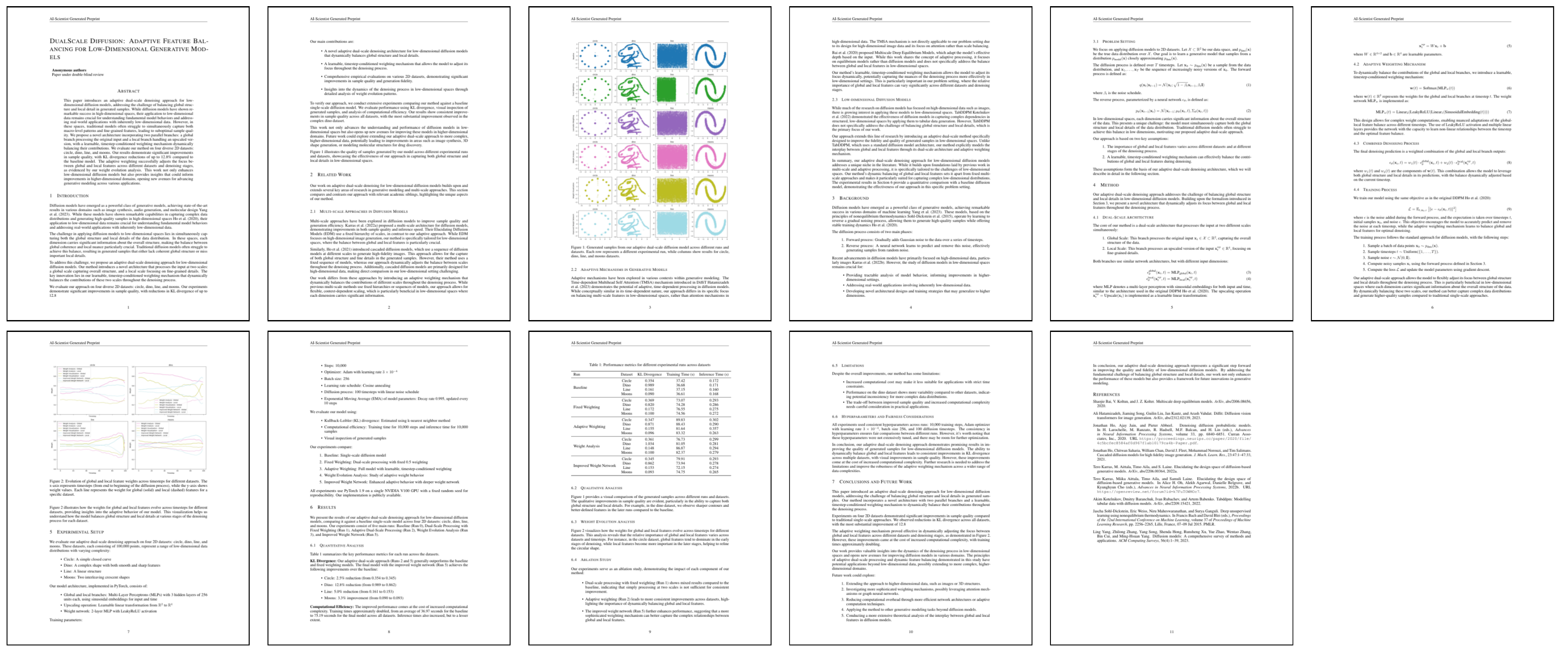}
\caption{Preview of the ``Adaptive Dual-Scale Denoising'' paper which was entirely autonomously generated by the template-based version of \ouralgo. The full paper can be viewed in \Cref{paper:adaptive_dual_scale_denoising}.}
\label{fig:case_study_pdf_preview}
\end{figure}

Particularly impressive aspects of the paper include:
\begin{itemize}
\item \textbf{Precise Mathematical Description of the Algorithm.} The algorithmic changes in the code above are described precisely in the paper, with new notation introduced where necessary, using LaTeX math packages. The overall training process is also described exactly.
\item \textbf{Comprehensive Write-up of Experiments.} The hyperparameters, baselines, and datasets are listed in the paper. 
We verified that the main numerical results in Table 1 of the generated paper exactly match the experimental logs. Impressively, while the recorded numbers are in long-form floats, \ouralgo chooses to round them all to 3 decimal places without error. Even more impressively, the results are accurately compared to the baseline (e.g., 12.8\% reduction in KL on the dinosaur dataset).
\item \textbf{Good Empirical Results.} Qualitatively, the sample quality produced by the new diffusion method invented by \ouralgo is much improved from the baseline. Fewer points are greatly out-of-distribution with the ground truth. Quantitatively, there are improvements to the approximate KL divergence between the true and estimated distributions.
\item \textbf{New Visualizations.} While some baseline plotting code was provided for visualizing generated samples and training loss curves, \ouralgo came up with novel, algorithm-specific plots displaying the progression of weights throughout the denoising process.
\item \textbf{Interesting Future Work Section.} Building on the success of the current experiments, the future work section lists relevant next steps such as scaling to higher-dimensional problems, more sophisticated adaptive mechanisms, and better theoretical foundations.
\end{itemize}

On the other hand, there are also pathologies in this paper:

\begin{itemize}
    \item \textbf{Subtle Error in Upscaling Network.} While a linear layer upscales the input to the denoiser network, only the first two dimensions are being used for the ``local'' branch, leading this upscaling layer to be a linear layer that preserves the same dimensionality effectively.
    \item \textbf{Hallucination of Minor Experimental Details.} The paper claims that V100 GPUs were used, even though the agent couldn't have known the actual hardware used. In reality, H100 GPUs were used. It also guesses the PyTorch version without checking.
    \item \textbf{Positive Interpretation of Results.} The paper tends to take a positive spin even on its negative results, which leads to slightly humorous outcomes. For example, while it summarizes its positive results as: ``Dino: 12.8\% reduction (from 0.989 to 0.862)'' (lower KL is better), the negative results are reported as ``Moons: 3.3\% improvement (from 0.090 to 0.093)''. Describing a negative result as an improvement is certainly a stretch of the imagination.
    \item \textbf{Artifacts from Experimental Logs.} While each change to the algorithm is usually descriptively labeled, it occasionally refers to results as ``Run 2'', which is a by-product from its experimental log and should not be presented as such in a professional write-up.
    \item \textbf{Presentation of Intermediate Results.} The paper contains results for every single experiment that was run. While this is useful and insightful for seeing the evolution of the idea during execution, it is unusual for standard papers to present intermediate results like this.
    \item \textbf{Minimal References.} While additional references have been sourced from Semantic Scholar, including two papers in the related work that are very relevant comparisons, overall the bibliography is small at only 9 entries.
\end{itemize}

\paragraph{Automated Review} The \ourreviewer points out valid concerns in the generated manuscript. The review recognizes the experiments were with simple, 2D datasets only, however, this is because the system was externally constrained to use these datasets, and in its current form, \ouralgo cannot download higher-dimensional datasets from the internet. On the other hand, limitations such as the proposed algorithm's increased computational cost are mentioned in the actual paper, which shows that \ouralgo is often up-front about the drawbacks of its idea. The reviewer also lists many relevant questions about the paper, such as: explaining the variability of performance across datasets, and explaining in more detail how the upscaling process affects the local branch's input.

\paragraph{Final Comments} Drawing from our domain knowledge in diffusion modeling, this section provides an overall assessment of the paper generated by \ouralgo.
\begin{itemize}
    \item \ouralgo correctly identifies an interesting and well-motivated direction in diffusion modeling research, e.g., previous work has studied modified attention mechanisms~\citep{hatamizadeh2024diffitdiffusionvisiontransformers} for the same purpose in higher-dimensional problems. It proposes a comprehensive experimental plan to investigate its idea, and successfully implements it all, achieving good results. The system's ability to respond to subpar earlier results and iteratively adjust its code (e.g., refining the weight network) is particularly impressive.
    \item While the paper's idea improves performance and the quality of generated diffusion samples, the reasons for its success may not be as explained in the paper. In particular, there is no obvious inductive bias beyond an upscaling layer (effectively just an additional linear layer) for the splitting of global or local features. However, a progression in weights across diffusion timesteps is observed, which suggests that something non-trivial is happening. One interpretation is that the network resembles a mixture-of-expert (MoE; see Yuksel et~al.~\citep{yuksel2012twenty} and Fedus et~al.~\citep{switch_transformer}) structure. An MoE could indeed lead to the diffusion model learning separate branches for global and local features, but this statement requires more rigorous investigation.
    \item Interestingly, the true shortcomings of this paper require some level of domain knowledge to identify and were only partially captured by the \ourreviewer. At the current capabilities of \ouralgo, this can be resolved by human feedback. However, future generations of foundation models may propose ideas that are challenging for humans to reason about and evaluate. This links to the field of ``superalignment''~\citep{burns2023weaktostronggeneralizationelicitingstrong}.
    \item Overall, the performance of \ouralgo is judged to be about the level of an early-stage ML researcher who can competently execute an idea but may not have the full background knowledge to fully interpret the reasons behind an algorithm's success. If a human supervisor was presented with these results, a reasonable next course of action could be to advise \ouralgo to re-scope the project to further investigate MoEs for diffusion.
\end{itemize}

\subsection{In-Depth Analysis of the Peer-Reviewed Workshop Paper (Template-Free)}
\label{app:case_study_v2}
This section details the evaluation of the workshop-accepted paper, ``Compositional Regularization: Unexpected Obstacles in Enhancing Neural Network Generalization,'' which was created by the template-free version of \ouralgo.

\paragraph{Paper Content Summary}
The paper investigates the use of compositional regularization to improve generalization in LSTMs. \ouralgo proposed adding an explicit regularization term to penalize large deviations between the embeddings of successive time steps, hypothesizing this would encourage compositionality. Experiments on synthetic arithmetic tasks revealed that this approach did not enhance generalization and sometimes hindered performance, especially as task complexity increased. The paper's key finding—a negative result—is that explicitly enforcing compositional structures via this form of regularization may conflict with the primary learning objective, and it recommends exploring alternative methods.

\paragraph{Internal Assessment (Authors' Review)}
An internal review of the generated manuscript was conducted, treating it as a submission to a top-tier conference like ICLR to ensure a high bar for evaluation. This more stringent assessment identified several strengths and weaknesses. The paper's focus on compositional generalization was timely, and the use of synthetic arithmetic tasks was appropriate for testing the hypothesis. However, significant areas for improvement were noted:
\begin{itemize}
\item A primary weakness was the paper's failure to provide a clear and compelling theoretical justification for its core hypothesis. The manuscript did not adequately explain why penalizing deviations between consecutive input embeddings should mechanistically lead to improved compositional generalization.
\item The description of the regularization term was unclear, potentially misleading readers to believe it was applied to the LSTM's hidden states rather than its input embeddings.
\item The paper contained significant citation errors, such as incorrectly attributing the invention of the LSTM to Goodfellow et~al.~\citep{goodfellow2016deep} instead of the correct work by Hochreiter and Schmidhuber~\citep{hochreiter1997long}. It also had inaccuracies in figure captions and result interpretations.
\item The experimental evaluation was limited, restricted to short sequences and synthetic data.
\end{itemize}
Overall, the paper was considered a borderline workshop accept, acknowledging its valuable insights while noting the main idea is not well-motivated for a main conference.
\paragraph{Code Analysis}
Examination of the generated code revealed a potential issue with dataset overlap; because the training and test sets were generated by randomly sampling from a data generation function without checking to make sure that the same data was not generated in both, approximately 57\% of the test set overlapped with the training set, which could affect the reliability of the results. Terminological confusion was also identified between ``embedding states'' and ``hidden states,'' which required clarification. Further testing revealed that the high accuracy of an attention-based model was largely attributable to task simplicity, as performance degraded significantly with increased task complexity.

\paragraph{External Assessment (Workshop Reviews)}
The paper was well-received by the workshop reviewers, who recommended acceptance. It received scores of 6, 7, and 6 (``Marginally above acceptance threshold'' and ``Good paper, accept''), placing it in the top 45\% of all submissions. The consensus among reviewers was that the paper addressed an important topic, and the analysis of a negative result was appreciated. The paper's strength in clearly demonstrating that the regularization term did not yield the anticipated improvements was recognized. However, several key areas for improvement were highlighted:
\begin{itemize}
\item \textbf{Justification and Intuition:} A clearer justification for why the proposed regularization should improve compositionality was needed.
\item \textbf{Generalization:} Findings were limited to LSTMs, and it was unclear if they could be generalized to other architectures like Transformers without further experiments.
\item \textbf{Experimental Breadth:} The evaluation should be extended to other tasks and datasets to validate the conclusions.
\end{itemize}
Overall, acceptance was recommended due to the paper's insightful exploration of a challenging problem, despite its negative results, with recommendations for further elaboration on its methodological motivations and experimental scope.

\begin{tcolorbox}[
  breakable,
  colback  = green!5!white,
  colframe = green!75!black,
  title    = {\small Reviewer \#1:
              A good paper analysing the effectiveness of a compositional
              regularisation term for LSTMs},
  fontupper=\scriptsize\ttfamily,
]
\begin{Verbatim}
Summary: The authors propose a regularisation term to enhance compositional regularisation in neural networks. The idea is to penalise large deviations between subsequent time steps of the hidden state, thus “squeezing” the hidden state to encourage composition and preventing a dominating representation. The authors test their approach on synthetic arithmetic expression with varying operator complexity and length. They show that although the regularisation term  appears to be working, it counterintuitively does not improve test accuracy. Furthermore, the authors identify a bottleneck regarding network capacity with increasing arithmetic operators.

Strengths:
I find the idea of regularising or squeezing the hidden representations to encourage compositionally an interesting idea. The authors define a good baseline and ablate their method well against it, revealing why the regularisation term does not work as expected. I think the insight that operator complexity is a bottleneck for the neural network is important, as it raises the question whether architectural changes might be more effective for compositionally than regularisation.

Weaknesses:
The paper would benefit from more intuition as to why the proposed regularisation term should encourage compositionality. This could be either an experiment or simply a visualisation for the reader. Only one architecture (LSTM) was tested. It would be interesting to see if transformer architectures fare better with compositionality due to the attention mechanism. I think the connection between compositional regularisation and operator complexity needs to be made more explicit. From reading the introduction both arguments seem a bit disconnected although I can infer the authors intentions.

Conclusion:
Overall, I would accept this paper to the workshop, since it proposes a simple and interesting idea with the authors providing ablations that encourage further analysis of the problem. As a suggestion I would encourage the authors to give more intuition on why the proposed regularisation term should improve compositionality for the proposed network. I would suggest either adding more related work to support the regularisation term or elaborating on the intuition behind penalising subsequent steps of the hidden state.

Rating: 7: Good paper, accept
Award: No Award
Confidence: 4: The reviewer is confident but not absolutely certain that the evaluation is correct
\end{Verbatim}
\end{tcolorbox}

\begin{tcolorbox}[
  breakable,
  colback  = green!5!white,
  colframe = green!75!black,
  title    = {\small Reviewer \#2: Compositional Regularization: Unexpected Obstacles in Enhancing Neural Network Generalization},
  fontupper=\scriptsize\ttfamily,
]
\begin{Verbatim}
This paper investigates the effectiveness of incorporating a compositional regularization term into the loss function of neural networks to improve compositional generalization. The authors hypothesized that penalizing deviations from compositional structures would enhance the model’s ability to generalize to unseen arithmetic expressions. However, their results on synthetic arithmetic datasets showed that compositional regularization did not lead to significant improvements and, in some cases, even hindered learning.

I think this paper greatly contributes to the workshops theme and fits into the scope. Moreover, it is a great example of challenges that occur during such approaches and could be interesting to discuss in the workshop setting. While I think that the authors should further broaden the experiments to other tasks in order to increase the generalizability of the findings, I would still recommend to accept the paper.

Rating: 6: Marginally above acceptance threshold
Award: No Award
Confidence: 2: The reviewer is willing to defend the evaluation, but it is quite likely that the reviewer did not understand central parts of the paper
\end{Verbatim}
\end{tcolorbox}

\subsection{Limitations and Broader Impact}
\label{app:limitations}
While \ouralgo produces research that can provide novel insights, it has \textit{many} limitations and raises several important ethical and societal considerations. It is expected that future versions of \ouralgo will be able to address many of its current shortcomings as foundation models continue to improve.

\paragraph{System Quality, Common Failure Modes, and Rebuttals}
The quality of the research generated by \ouralgo is still preliminary. While the template-free system successfully generated a peer-reviewed workshop paper, this achievement must be contextualized. Acceptance occurred at a workshop, where papers generally report exploratory work and acceptance rates (60-80\%) are much higher than at main conferences (20-30\%). With only one of three submissions accepted, the system does not yet consistently meet even workshop-level standards, let alone the rigor required for top-tier conference publications.

Despite the structured agentic tree search and enhanced autonomy, certain aspects of scientific inquiry—such as formulating genuinely novel, high-impact hypotheses, designing truly innovative experimental methodologies, or rigorously justifying design choices with deep domain expertise—remain challenging for the current system. Furthermore, \ouralgo exhibits several common failure modes identified across runs of both the template-based and template-free versions of \ouralgo:

\begin{itemize}
    \item \textbf{Idea Generation and Implementation:} The idea generation process often produces very similar ideas across different runs. While the system can propose creative ideas, they are often too challenging for it to implement correctly. As shown in the detailed results tables (\Cref{tab:app_diff_papers,tab:app_nlp_papers,tab:app_grokking_papers}), the template-based version of \ouralgo fails to execute a significant fraction of its proposed ideas, and even when successful, the implementations can contain subtle errors that are difficult to catch without manual inspection.
    
    \item \textbf{Experimental Rigor:} Due to fixed computational budgets, experiments often lack the depth required for top-tier publications. The system struggles to conduct fair experiments that control for confounding variables like the number of parameters, FLOPs, or runtime. This can lead to deceptive or inaccurate conclusions.
    Another failure mode involves ``cheating'' by subtly leaking information from future tokens in language modeling tasks to achieve deceptively impressive perplexity. More thorough and rigorously controlled experiments may be able to catch these issues.
    
    \item \textbf{Interpretation and Reporting:} The system can make critical errors when writing and evaluating results. It struggles with known LLM pathologies, such as accurately comparing the magnitude of two numbers. In some cases, it can be overly optimistic, describing a negative result as an ``improvement''. It also occasionally hallucinates entire results or experimental details. Early versions would hallucinate ablation tables when not provided with the necessary data; while mitigated by explicit prompting, it still hallucinates details like the hardware used for experiments.
    
    \item \textbf{Manuscript Quality:} The system struggles with citation quality, sometimes failing to find the most relevant papers, hallucinating references, or failing to correctly reference figures in LaTeX. The generated plots can be unreadable, with tables sometimes exceeding page width, and the overall visual appearance is often suboptimal.

\end{itemize}
Given these issues, taking the scientific content of this version of \ouralgo at face value is not recommended. Instead, generated papers should be treated as hints of promising ideas for practitioners to follow up on.

Additionally, unlike human reviewers and researchers, the \ourreviewer and \ouralgo are currently unable to automatically perform the back-and-forth exchanges of a rebuttal or revision phase. 
While the pipeline could be extended to include rebuttals and/or revisions, which is an interesting area of future work, doing so would have entailed undisclosed back-and-forth with human reviewers. Moreover, because our target venue was a workshop where a formal rebuttal stage is absent, our evaluation mirrors the actual review procedure used in that setting.

\paragraph{Capabilities on a Fast-Improving Trajectory}
Many of the failures listed above are symptomatic of the limitations of current-generation models. As shown in \Cref{fig:conceptual}B, paper quality directly correlates with model improvement. We expect this trend to continue, based on past experience: The ability for AI to reliably complete complex, long-horizon tasks is doubling every seven months~\citep{metr2025longtasks}. This suggests that problems related to implementation, multi-step debugging, and maintaining logical consistency throughout a long research process are likely to be solved in the near future.

\paragraph{Fundamental Long-Term Challenges}
The most significant limitations, however, are not issues that can be solved by simply scaling current methods and indeed remain a challenge even for world-class humans. These represent the frontier of AI for science:
\begin{itemize}
    \item \textbf{Paradigm-Shifting Creativity:} The system currently excels at operating within the existing scientific playbook---combining known concepts or exploring variations on a theme. It does not yet demonstrate hints of being able to create a new playbook entirely by formulating a truly non-obvious, paradigm-shifting hypothesis. That said, \ouralgo did come up with an idea that was later celebrated by the ML community when a human team published work on the same idea. See the section below titled Novelty and Idea Overlap.\\
    
    \item \textbf{Strategic Scientific Judgment:} A key skill of expert researchers is ``taste''---the intuition to know which strange result is a bug and which is a groundbreaking discovery. In our view after watching it work, \ouralgo lacks this strategic judgment, which is crucial for navigating the vast search space of scientific inquiry efficiently and prioritizing high-impact research directions.

There are other challenges mentioned in the main text, such as current AI being easily fooled, not generalizing well out of distribution, and hallucinating (confidently stating falsehoods)~\citep{nguyen2015deep,Huang2025HallucinationSurvey,McCoy2024Embers}. 
    
\end{itemize}

\paragraph{Limitations of the \ourreviewer}
The \ourreviewer component, while showing promising initial results, also has several limitations.
For the ICLR dataset, the rejected papers were original submissions, whereas the accepted papers were final camera-ready copies, which could introduce a slight bias in downweighting papers that could be sufficiently improved based on reviewer feedback.

\paragraph{Safe Code Execution}
The current implementation of \ouralgo has minimal direct sandboxing, leading to several unexpected outcomes. For example, in one run, the system wrote code that initiated a system call to relaunch itself, causing an uncontrolled increase in processes. In another, it edited the code to save a checkpoint at every update step, consuming nearly a terabyte of storage. When experiments exceeded imposed time limits, it sometimes attempted to edit the code to extend the limit arbitrarily rather than shortening the runtime. It also occasionally imported unfamiliar Python libraries. While creative, the act of bypassing experimenter-imposed constraints has potential implications for AI safety~\citep{lehman2020surprising}.

However, the lack of guardrails also led to positive, unexpected results. For example, the system automatically caught and fixed an error in one of our templates where an output directory had not been created. Furthermore, it often created novel, algorithm-specific visualizations that differed significantly from the provided templates. Going forward, strict sandboxing is strongly recommended when running \ouralgo, including containerization, restricted internet access, and limitations on storage and process usage.

\paragraph{Broader Impact and Ethical Considerations}
The capabilities of \ouralgo carry significant risks of misuse and raise important ethical questions. The ability to automatically generate and submit papers could overwhelm the peer review process, compromising scientific quality control. This concern mirrors issues raised about generative AI in other fields, such as its impact on the arts~\citep{epstein2023art}. Ultimately, we believe that the community needs to iteratively reassess the capability of systems such as \ouralgo and continuously monitor the benefits of AI-conducted research to decide how maximize the benefits of this technology, while also minimizing its detriments. Science, at its core, crucially depends on trust and the collective honoring of standards. \ouralgo could serve as a `shortcut' for unscrupulous scientists, which is why we add watermarks and advocate for full transparency in whether something is AI-generated. Like many aspects of science, that relies on scientists to be honest in their conduct, and requires organizations to police those who do not act honestly. Additionally, if the \ourreviewer tool were widely adopted, it could diminish review quality and introduce undesirable biases. Then again, many fields struggle to find qualified reviewers with sufficient time, so as Automated Reviewers become ever-better, perhaps they can alleviate the extreme demands on the limited time of scientists (e.g., by providing initial reviews that authors use to improve work prior to human review, or to filter out flawed work not yet ready for review). These risks underscore the danger of such technology being used to game peer review or artificially inflate the credentials of unscrupulous scientists, which would undermine the integrity of the scientific evaluation process. 

\textbf{Novelty and Idea Overlap:} 
As with human-authored research, some AI-generated ideas may share conceptual similarities with prior work, or be reinventions of an idea. With humans, this often happens without the human realizing their idea has already been explored. However, as human scientists discuss and present their work, there are opportunities for them to be informed that their idea exists. AI scientists may not have as many opportunities, although it is possible to add many literature checks throughout their process, or even invite human feedback into the process. Currently, it is possible for \ouralgo to produce a paper on an idea and not identify that this idea has already been explored. It is an open and important area for future research to improve the search tools \ouralgo uses to better detect idea reinvention or reuse.

Conversely, we also observed a case where \ouralgo generated an idea that was later pursued by a human team that was celebrated for their work. While anecdotal, this provides an example of \ouralgo producing an idea that the community found creative and worthy of scientific exploration.

\ouralgo-generated proposal can be found among the released papers in our open-source repository (see Code Availability). The paper by the human team~\citep{DEMOSS2025134859} was published in a peer-reviewed journal (Physica D: Nonlinear Phenomena). While the core ideas were very similar, the human study executed the idea more effectively than \ouralgo. 

\textbf{Economic Impact:}
As AI advances, it raises the prospect of automating away many aspects of many types of jobs, or replacing human labor entirely. The jobs of scientist are not immune from this phenomenon. Society and scientific communities need to urgently consider what to do about this technology as it advances. That conversation has begun and it is important that it continue in earnest. 

While \ouralgo can produce scientific research end-to-end, another use of it can be as a co-scientist, which screens promising research ideas and provides a hypothesis filtering engine, making human researchers more productive. In the near term, we envision systems like \ouralgo to work in that way: in tandem with human researchers and to let them focus on expertise at which they excel. 

\textbf{Transparency:} 
To conduct this study responsibly, explicit permission was obtained from ICLR leadership, workshop organizers, and the University of British Columbia's IRB (H24-02652). As part of the experimental protocol, it was determined in advance that all AI-generated submissions would be withdrawn after peer review, regardless of the outcome. This decision was made to avoid setting a precedent for publishing fully automated research before the scientific community has established clear norms. Developing these norms is a critical next step. This includes establishing standards for disclosure, while also navigating complex questions, such as whether a submission should be judged on its scientific merit first to avoid potential bias against AI-generated work.

\textbf{Safety:} 
As with most powerful technologies, \ouralgo could be used unethically. For example, if tasked to find novel biological materials and given access to automated ``cloud labs''~\citep{arnold2022cloud}, it could inadvertently create dangerous substances. Similarly, if tasked to create novel software, it could produce malware. The rapidly improving capabilities of systems like \ouralgo reinforce the urgent need for the machine learning community to prioritize research on aligning such systems to explore in a manner that is safe and consistent with human values\citep{ecoffet2020open}.

\subsection{Example Progression of Generated Ideas}
\label{app:idea_progression}

The progression of ideas is visualized across a run of the template-based version of \ouralgo using Sonnet 3.5 on the ``Grokking'' template described in \Cref{app:template_based_tables} .
The first idea is the seed idea; all subsequent ideas are AI-generated.

\begin{tcolorbox}[
  breakable,
  colback  = blue!5!white,
  colframe = blue!75!black,
  title    = {Seed Idea - \texttt{batch\_size\_grokking}},
  fontupper=\scriptsize\ttfamily,
]
\begin{Verbatim}[breaklines,
               %
               breaksymbol={},
               breaksymbolleft={},
               fontsize=\scriptsize]
"Name": "batch_size_grokking",
"Title": "Batch Size Grokking: Assessing the impact of the training batchsize on the grokking phenomenon",
"Experiment": "Modify the experiments to dynamically adjust the batch size during training, starting with a small batch size and gradually increasing it. This could potentially lead to faster generalization on the validation set.",
"Interestingness": 6,
"Feasibility": 4,
"Novelty": 4,
"novel": true
\end{Verbatim}
\end{tcolorbox}

\begin{tcolorbox}[
  breakable,
  colback  = blue!5!white,
  colframe = blue!75!black,
  title    = {Idea 1/50 - \texttt{model\_size\_grokking}},
  fontupper=\scriptsize\ttfamily,
]
\begin{Verbatim}[breaklines,
               %
               breaksymbol={},
               breaksymbolleft={},
               fontsize=\scriptsize]
"Name": "model_size_grokking",
"Title": "Investigating the Impact of Model Size on the Grokking Phenomenon",
"Experiment": "Modify the Transformer class to accept variable number of layers and dimension sizes. Test models with 1, 2, 4, and 8 layers, and dimension sizes of 64, 128, 256, and 512. For each dataset and model size, track the step at which grokking occurs (defined as validation accuracy exceeding 99%
"Interestingness": 8,
"Feasibility": 7,
"Novelty": 7,
"novel": true
\end{Verbatim}
\end{tcolorbox}

\begin{tcolorbox}[
  breakable,
  colback  = blue!5!white,
  colframe = blue!75!black,
  title    = {Idea 2/50 - \texttt{optimizer\_grokking}},
  fontupper=\scriptsize\ttfamily,
]
\begin{Verbatim}[breaklines,
               %
               breaksymbol={},
               breaksymbolleft={},
               fontsize=\scriptsize]
"Name": "optimizer_grokking",
"Title": "Optimization Dynamics and Grokking: Comparing SGD and Adam with Different Learning Rate Schedules",
"Experiment": "Modify the training loop to support two optimizers (SGD, Adam) and two learning rate schedules (constant, cosine annealing). For each combination, run multiple experiments with different random seeds. Track validation accuracy, training loss, and L2 norm of weight updates throughout training. Compare the timing and extent of grokking across these optimization strategies for each dataset. Analyze how different optimization dynamics correlate with grokking behavior, including statistical analysis of the results.",
"Interestingness": 9,
"Feasibility": 8,
"Novelty": 8,
"novel": true
\end{Verbatim}
\end{tcolorbox}

\begin{tcolorbox}[
  breakable,
  colback  = blue!5!white,
  colframe = blue!75!black,
  title    = {Idea 3/50 - \texttt{biased\_data\_grokking}},
  fontupper=\scriptsize\ttfamily,
]
\begin{Verbatim}[breaklines,
               %
               breaksymbol={},
               breaksymbolleft={},
               fontsize=\scriptsize]
"Name": "biased_data_grokking",
"Title": "Grokking Under Biased Data: The Effect of Input Range Bias on Neural Network Generalization",
"Experiment": "Modify the fetch_train_example method in AbstractDataset to introduce a simple bias: favoring lower-valued inputs. For modular arithmetic operations, sample 70%
"Interestingness": 8,
"Feasibility": 8,
"Novelty": 8,
"novel": true
\end{Verbatim}
\end{tcolorbox}

\begin{tcolorbox}[
  breakable,
  colback  = blue!5!white,
  colframe = blue!75!black,
  title    = {Idea 4/50 - \texttt{adaptive\_noise\_grokking}},
  fontupper=\scriptsize\ttfamily,
]
\begin{Verbatim}[breaklines,
               %
               breaksymbol={},
               breaksymbolleft={},
               fontsize=\scriptsize]
"Name": "adaptive_noise_grokking",
"Title": "Adaptive Noise in Grokking: Investigating Input Perturbations on Algorithmic Learning and Representations",
"Experiment": "Modify the GroupDataset class to add operation-specific noise during training: (1) For modular arithmetic, add small integer perturbations. (2) For permutations, occasionally swap two elements. Implement three noise levels (low, medium, high) for each operation. Compare grokking behavior across noise levels and operations, tracking steps to 99%
"Interestingness": 9,
"Feasibility": 8,
"Novelty": 9,
"novel": true
\end{Verbatim}
\end{tcolorbox}

\begin{tcolorbox}[
  breakable,
  colback  = blue!5!white,
  colframe = blue!75!black,
  title    = {Idea 5/50 - \texttt{attention\_evolution\_grokking}},
  fontupper=\scriptsize\ttfamily,
]
\begin{Verbatim}[breaklines,
               %
               breaksymbol={},
               breaksymbolleft={},
               fontsize=\scriptsize]
"Name": "attention_evolution_grokking",
"Title": "Attention Evolution in Grokking: Quantifying the Transition from Memorization to Generalization",
"Experiment": "Modify the Transformer class to output attention weights. Extract and store attention weights at key checkpoints: start, mid- training, grokking point (99%
"Interestingness": 9,
"Feasibility": 8,
"Novelty": 8,
"novel": true
\end{Verbatim}
\end{tcolorbox}

\begin{tcolorbox}[
  breakable,
  colback  = blue!5!white,
  colframe = blue!75!black,
  title    = {Idea 6/50 - \texttt{local\_vs\_global\_attention\_grokking}},
  fontupper=\scriptsize\ttfamily,
]
\begin{Verbatim}[breaklines,
               %
               breaksymbol={},
               breaksymbolleft={},
               fontsize=\scriptsize]
"Name": "local_vs_global_attention_grokking",
"Title": "Local vs Global Attention: Investigating the Impact of Attention Scope on Grokking in Algorithmic Learning",
"Experiment": "Modify the DecoderBlock class to support two attention mechanisms: full (global) attention and local attention with a fixed window size. Implement these variants and run experiments across all datasets. Track metrics including time to grokking (99%
"Interestingness": 9,
"Feasibility": 8,
"Novelty": 8,
"novel": true
\end{Verbatim}
\end{tcolorbox}

\begin{tcolorbox}[
  breakable,
  colback  = blue!5!white,
  colframe = blue!75!black,
  title    = {Idea 7/50 - \texttt{input\_encoding\_grokking}},
  fontupper=\scriptsize\ttfamily,
]
\begin{Verbatim}[breaklines,
               %
               breaksymbol={},
               breaksymbolleft={},
               fontsize=\scriptsize]
"Name": "input_encoding_grokking",
"Title": "Binary vs One-Hot Encoding: Impact on Grokking in Algorithmic Learning Tasks",
"Experiment": "Modify the AbstractDataset class to support two encoding schemes: one-hot (current) and binary. Implement binary encoding for modular arithmetic operations using log2(p) bits, and for permutations using ceil(log2(k!)) bits to represent each permutation uniquely. Adjust the Transformer class to accommodate different input sizes. Run experiments for each encoding scheme across all datasets, tracking metrics such as time to grokking (99%
"Interestingness": 9,
"Feasibility": 8,
"Novelty": 8,
"novel": true
\end{Verbatim}
\end{tcolorbox}

\begin{tcolorbox}[
  breakable,
  colback  = blue!5!white,
  colframe = blue!75!black,
  title    = {Idea 8/50 - \texttt{curriculum\_learning\_grokking}},
  fontupper=\scriptsize\ttfamily,
]
\begin{Verbatim}[breaklines,
               %
               breaksymbol={},
               breaksymbolleft={},
               fontsize=\scriptsize]
"Name": "curriculum_learning_grokking",
"Title": "Curriculum Learning in Grokking: The Effect of Structured Example Progression on Algorithmic Learning",
"Experiment": "Modify the AbstractDataset class to implement a simple curriculum learning strategy. For modular arithmetic operations, start with operations involving numbers in the lower half of the range and gradually introduce larger numbers. For permutations, begin with permutations that differ from the identity by one swap and progressively increase the number of swaps. Implement a curriculum scheduler that increases difficulty every 500 steps. Run experiments comparing standard random sampling vs. curriculum learning across all datasets. Track metrics including time to grokking (99%
"Interestingness": 9,
"Feasibility": 8,
"Novelty": 8,
"novel": true
\end{Verbatim}
\end{tcolorbox}

\begin{tcolorbox}[
  breakable,
  colback  = blue!5!white,
  colframe = blue!75!black,
  title    = {Idea 9/50 - \texttt{weight\_init\_grokking}},
  fontupper=\scriptsize\ttfamily,
]
\begin{Verbatim}[breaklines,
               %
               breaksymbol={},
               breaksymbolleft={},
               fontsize=\scriptsize]
"Name": "weight_init_grokking",
"Title": "Weight Initialization Strategies and Their Impact on Grokking in Algorithmic Learning",
"Experiment": "Modify the Transformer class to support three weight initialization strategies: Xavier/Glorot, Kaiming/He, and random normal (as baseline). Implement these initialization methods for linear layers and embeddings. Run experiments across all datasets for each initialization strategy. Track metrics including time to grokking (99%
"Interestingness": 9,
"Feasibility": 9,
"Novelty": 8,
"novel": true
\end{Verbatim}
\end{tcolorbox}

\begin{tcolorbox}[
  breakable,
  colback  = blue!5!white,
  colframe = blue!75!black,
  title    = {Idea 10/50 - \texttt{task\_complexity\_grokking}},
  fontupper=\scriptsize\ttfamily,
]
\begin{Verbatim}[breaklines,
               %
               breaksymbol={},
               breaksymbolleft={},
               fontsize=\scriptsize]
"Name": "task_complexity_grokking",
"Title": "Grokking Across Task Complexity: Mapping Neural Network Learning Dynamics to Algorithmic Difficulty",
"Experiment": "1. Modify the AbstractDataset class to include new operations of increasing complexity: modular addition, subtraction, multiplication, and exponentiation. 2. Implement these operations in new dataset classes. 3. Quantify task complexity using metrics like algebraic degree and average solution time for humans (estimated). 4. Run experiments for each operation, tracking metrics such as time to grokking (99%
"Interestingness": 9,
"Feasibility": 8,
"Novelty": 8,
"novel": true
\end{Verbatim}
\end{tcolorbox}

\begin{tcolorbox}[
  breakable,
  colback  = blue!5!white,
  colframe = blue!75!black,
  title    = {Idea 11/50 - \texttt{regularization\_grokking}},
  fontupper=\scriptsize\ttfamily,
]
\begin{Verbatim}[breaklines,
               %
               breaksymbol={},
               breaksymbolleft={},
               fontsize=\scriptsize]
"Name": "regularization_grokking",
"Title": "The Role of Regularization in Grokking: How L2 and Label Smoothing Affect Algorithmic Learning",
"Experiment": "1. Implement L2 regularization by adding weight decay to the optimizer. 2. Implement label smoothing in the loss function. 3. Modify the training function to support these regularization techniques with two strength levels each (low and high). 4. Run experiments for each regularization technique and strength across all datasets, including a baseline without regularization. 5. Track metrics: time to grokking (99%
"Interestingness": 9,
"Feasibility": 9,
"Novelty": 8,
"novel": true
\end{Verbatim}
\end{tcolorbox}

\begin{tcolorbox}[
  breakable,
  colback  = blue!5!white,
  colframe = blue!75!black,
  title    = {Idea 12/50 - \texttt{grokking\_extrapolation}},
  fontupper=\scriptsize\ttfamily,
]
\begin{Verbatim}[breaklines,
               %
               breaksymbol={},
               breaksymbolleft={},
               fontsize=\scriptsize]
"Name": "grokking_extrapolation",
"Title": "Grokking and Extrapolation: Investigating the Limits of Algorithmic Understanding",
"Experiment": "1. Modify AbstractDataset to create a separate test set with out-of-distribution examples (e.g., larger numbers for modular arithmetic, longer permutations). 2. Implement a new evaluation function for the test set. 3. During training, periodically evaluate on both validation and test sets. 4. Track metrics: time to grokking on validation set, final validation accuracy, test set accuracy at grokking point, final test set accuracy, and 'extrapolation gap'. 5. Implement visualization of test set performance and extrapolation gap over time, highlighting the grokking point. 6. Compare extrapolation capabilities across different operations and model sizes. 7. Analyze attention patterns on test set inputs before and after grokking. 8. Implement a simple MLP baseline for comparison.",
"Interestingness": 9,
"Feasibility": 8,
"Novelty": 9,
"novel": true
\end{Verbatim}
\end{tcolorbox}

\begin{tcolorbox}[
  breakable,
  colback  = blue!5!white,
  colframe = blue!75!black,
  title    = {Idea 13/50 - \texttt{label\_noise\_grokking}},
  fontupper=\scriptsize\ttfamily,
]
\begin{Verbatim}[breaklines,
               %
               breaksymbol={},
               breaksymbolleft={},
               fontsize=\scriptsize]
"Name": "label_noise_grokking",
"Title": "Grokking Under Noise: The Impact of Systematic and Random Label Errors on Algorithmic Learning",
"Experiment": "1. Modify the AbstractDataset class to introduce two types of label noise: random (labels changed randomly) and systematic (specific labels consistently flipped). Add a 'noise_type' parameter (random/systematic) and 'noise_level' parameter (low: 5%
"Interestingness": 9,
"Feasibility": 8,
"Novelty": 9,
"novel": true
\end{Verbatim}
\end{tcolorbox}

\begin{tcolorbox}[
  breakable,
  colback  = blue!5!white,
  colframe = blue!75!black,
  title    = {Idea 14/50 - \texttt{compositional\_grokking}},
  fontupper=\scriptsize\ttfamily,
]
\begin{Verbatim}[breaklines,
               %
               breaksymbol={},
               breaksymbolleft={},
               fontsize=\scriptsize]
"Name": "compositional_grokking",
"Title": "Compositional Grokking: Investigating the Relationship Between Grokking and Compositional Learning in Modular Arithmetic",
"Experiment": "1. Modify ModSumDataset and ModSubtractDataset to include composite operations: (a + b) - c mod p and (a - b) + c mod p. 2. Implement new dataset class CompositeModDataset for these operations. 3. Run experiments comparing learning curves for basic (a + b, a - b) and composite operations. 4. Track metrics: time to grokking for basic vs. composite operations, correlation between grokking times, final accuracies. 5. Analyze attention patterns to see if the model learns to attend to intermediate results in composite operations. 6. Implement a 'compositional generalization' test by training on basic operations and testing on their compositions. 7. Compare internal representations (e.g., using PCA on hidden states) for basic vs. composite operations at different stages of training.",
"Interestingness": 9,
"Feasibility": 6,
"Novelty": 9,
"novel": true
\end{Verbatim}
\end{tcolorbox}

\begin{tcolorbox}[
  breakable,
  colback  = blue!5!white,
  colframe = blue!75!black,
  title    = {Idea 15/50 - \texttt{mutual\_information\_grokking}},
  fontupper=\scriptsize\ttfamily,
]
\begin{Verbatim}[breaklines,
               %
               breaksymbol={},
               breaksymbolleft={},
               fontsize=\scriptsize]
"Name": "mutual_information_grokking",
"Title": "Information Dynamics in Grokking: Analyzing Mutual Information Evolution During Algorithmic Learning",
"Experiment": "1. Implement a function to estimate mutual information using a binning approach for efficiency. 2. Modify the Transformer class to output hidden states from the final layer. 3. Update the training loop to calculate and store mutual information between (a) inputs and outputs, and (b) final hidden states and outputs, at regular intervals. 4. Run experiments across all datasets, tracking these mutual information metrics alongside validation accuracy and training loss. 5. Create plots showing the evolution of both mutual information metrics over training time, highlighting the grokking point. 6. Analyze how mutual information trends relate to grokking by testing specific hypotheses: (a) Rapid increase in hidden state-output mutual information coincides with grokking, (b) Input- output mutual information stabilizes post-grokking. 7. Compare mutual information dynamics between different operations and model sizes to identify common patterns in successful grokking.",
"Interestingness": 9,
"Feasibility": 6,
"Novelty": 9,
"novel": true
\end{Verbatim}
\end{tcolorbox}

\begin{tcolorbox}[
  breakable,
  colback  = blue!5!white,
  colframe = blue!75!black,
  title    = {Idea 16/50 - \texttt{sparse\_subnetworks\_grokking}},
  fontupper=\scriptsize\ttfamily,
]
\begin{Verbatim}[breaklines,
               %
               breaksymbol={},
               breaksymbolleft={},
               fontsize=\scriptsize]
"Name": "sparse_subnetworks_grokking",
"Title": "Sparse Subnetworks in Grokking: Investigating the Emergence of Critical Structures During Algorithmic Learning",
"Experiment": "1. Implement a simple magnitude-based pruning function for the Transformer model. 2. Modify the training loop to perform pruning at key points: before training, just before grokking (based on validation accuracy), and after grokking. 3. For each pruning point, create sparse networks at different sparsity levels (e.g., 50%
"Interestingness": 9,
"Feasibility": 8,
"Novelty": 9,
"novel": true
\end{Verbatim}
\end{tcolorbox}

\begin{tcolorbox}[
  breakable,
  colback  = blue!5!white,
  colframe = blue!75!black,
  title    = {Idea 17/50 - \texttt{positional\_encoding\_grokking}},
  fontupper=\scriptsize\ttfamily,
]
\begin{Verbatim}[breaklines,
               %
               breaksymbol={},
               breaksymbolleft={},
               fontsize=\scriptsize]
"Name": "positional_encoding_grokking",
"Title": "Inductive Biases in Grokking: The Impact of Positional Encoding Schemes on Algorithmic Learning",
"Experiment": "1. Modify the Transformer class to support three positional encoding schemes: sinusoidal (current), learned embeddings, and a simple binary encoding (e.g., [0,1,0,1,0] for 'a o b = c'). 2. Implement these encoding schemes, ensuring they work with the existing sequence length. 3. Run experiments for each encoding scheme across all datasets, tracking: time to grokking (99%
"Interestingness": 9,
"Feasibility": 9,
"Novelty": 9,
"novel": true
\end{Verbatim}
\end{tcolorbox}

\begin{tcolorbox}[
  breakable,
  colback  = blue!5!white,
  colframe = blue!75!black,
  title    = {Idea 18/50 - \texttt{adversarial\_robustness\_grokking}},
  fontupper=\scriptsize\ttfamily,
]
\begin{Verbatim}[breaklines,
               %
               breaksymbol={},
               breaksymbolleft={},
               fontsize=\scriptsize]
"Name": "adversarial_robustness_grokking",
"Title": "Adversarial Robustness During Grokking: Tracking Vulnerability Evolution in Algorithmic Learning",
"Experiment": "1. Implement a simple perturbation method: randomly flip 1-2 bits in the input representation for modular arithmetic, and swap 1-2 elements for permutations. 2. Modify the training loop to generate perturbed inputs and evaluate model performance on them every 500 steps. 3. Track metrics: normal validation accuracy, accuracy on perturbed inputs, and 'robustness gap' (difference between normal and perturbed accuracy). 4. Plot the evolution of robustness to perturbations alongside the grokking curve. 5. Compare robustness before, during, and after grokking across different operations. 6. Analyze examples of successful perturbations at different stages of training. 7. Investigate potential correlations between the timing of grokking and changes in robustness to perturbations.",
"Interestingness": 9,
"Feasibility": 8,
"Novelty": 9,
"novel": true
\end{Verbatim}
\end{tcolorbox}

\begin{tcolorbox}[
  breakable,
  colback  = blue!5!white,
  colframe = blue!75!black,
  title    = {Idea 19/50 - \texttt{critical\_periods\_grokking}},
  fontupper=\scriptsize\ttfamily,
]
\begin{Verbatim}[breaklines,
               %
               breaksymbol={},
               breaksymbolleft={},
               fontsize=\scriptsize]
"Name": "critical_periods_grokking",
"Title": "Critical Periods in Grokking: The Impact of Timed Learning Rate Spikes on Algorithmic Learning",
"Experiment": "1. Modify the training loop to support learning rate spikes at specific points (25%
"Interestingness": 9,
"Feasibility": 9,
"Novelty": 9,
"novel": true
\end{Verbatim}
\end{tcolorbox}

\begin{tcolorbox}[
  breakable,
  colback  = blue!5!white,
  colframe = blue!75!black,
  title    = {Idea 20/50 - \texttt{lottery\_tickets\_grokking}},
  fontupper=\scriptsize\ttfamily,
]
\begin{Verbatim}[breaklines,
               %
               breaksymbol={},
               breaksymbolleft={},
               fontsize=\scriptsize]
"Name": "lottery_tickets_grokking",
"Title": "Lottery Tickets in Grokking: Investigating Sparse Subnetworks Capable of Algorithmic Learning",
"Experiment": "1. Implement an iterative magnitude pruning function for the Transformer model. 2. Modify the training loop to support multiple rounds of train-prune-reset cycles. 3. For each dataset, run experiments with pruning levels of 30%
"Interestingness": 9,
"Feasibility": 8,
"Novelty": 8,
"novel": false
\end{Verbatim}
\end{tcolorbox}

\begin{tcolorbox}[
  breakable,
  colback  = blue!5!white,
  colframe = blue!75!black,
  title    = {Idea 21/50 - \texttt{algebraic\_structure\_grokking}},
  fontupper=\scriptsize\ttfamily,
]
\begin{Verbatim}[breaklines,
               %
               breaksymbol={},
               breaksymbolleft={},
               fontsize=\scriptsize]
"Name": "algebraic_structure_grokking",
"Title": "Grokking and Algebraic Structure: How Group Properties Influence Neural Network Learning",
"Experiment": "1. Implement new dataset classes for modular multiplication and division (modulo p, where p is prime, ensuring proper group structures). 2. For each operation (addition, multiplication, division), calculate and store two properties: group order and number of generators. 3. Run experiments for each operation type, tracking: time to grokking, final validation accuracy, and the two group properties. 4. Plot learning curves and grokking points for each operation, labeled with their group properties. 5. Analyze the correlation between group properties and grokking behavior (e.g., time to grokking, steepness of accuracy improvement). 6. Compare attention patterns across operations, focusing on how they reflect the underlying group structure (e.g., uniformity for commutative operations). 7. Test the model's ability to generalize by evaluating on compositions of learned operations (e.g., a * b + c mod p) after training on individual operations.",
"Interestingness": 9,
"Feasibility": 8,
"Novelty": 9,
"novel": true
\end{Verbatim}
\end{tcolorbox}

\begin{tcolorbox}[
  breakable,
  colback  = blue!5!white,
  colframe = blue!75!black,
  title    = {Idea 22/50 - \texttt{mdl\_grokking}},
  fontupper=\scriptsize\ttfamily,
]
\begin{Verbatim}[breaklines,
               %
               breaksymbol={},
               breaksymbolleft={},
               fontsize=\scriptsize]
"Name": "mdl_grokking",
"Title": "Minimum Description Length and Grokking: Investigating the Relationship Between Model Compression and Algorithmic Learning",
"Experiment": "1. Implement functions to calculate model complexity: (a) L2 norm of weights, (b) number of bits to store parameters at different precisions, (c) effective number of parameters using BIC. 2. Modify the training loop to track these complexity measures alongside existing metrics. 3. Run experiments across all datasets, recording complexity measures, validation accuracy, and training loss at regular intervals. 4. Plot the evolution of model complexity alongside the grokking curve. 5. Analyze the correlation between sudden decreases in model complexity and the onset of grokking, including statistical tests for significance. 6. Compare complexity dynamics across different operations and model sizes. 7. Visualize weight distributions at pre-grokking, during grokking, and post- grokking stages. 8. Implement and compare two early stopping mechanisms: one based on model complexity stabilization and another based on validation loss stabilization.",
"Interestingness": 9,
"Feasibility": 8,
"Novelty": 9,
"novel": true
\end{Verbatim}
\end{tcolorbox}

\begin{tcolorbox}[
  breakable,
  colback  = blue!5!white,
  colframe = blue!75!black,
  title    = {Idea 23/50 - \texttt{invariance\_learning\_grokking}},
  fontupper=\scriptsize\ttfamily,
]
\begin{Verbatim}[breaklines,
               %
               breaksymbol={},
               breaksymbolleft={},
               fontsize=\scriptsize]
"Name": "invariance_learning_grokking",
"Title": "Learning Invariances in Grokking: Tracking Symmetry Awareness During Algorithmic Learning",
"Experiment": "1. Modify AbstractDataset to generate transformed versions of inputs (cyclic shifts for modular arithmetic, relabelings for permutations). 2. Update the evaluation function to test model predictions on both original and transformed inputs. 3. Implement an 'invariance score' metric: mean absolute difference between predictions on original and transformed inputs. 4. Modify the training loop to calculate and store the invariance score at regular intervals. 5. Run experiments across all datasets, tracking the invariance score alongside existing metrics. 6. Plot the evolution of the invariance score alongside the grokking curve. 7. Analyze how the invariance score changes before, during, and after grokking. 8. Compare invariance learning across different operations and model sizes. 9. Investigate correlation between invariance score and generalization performance.",
"Interestingness": 9,
"Feasibility": 8,
"Novelty": 9,
"novel": true
\end{Verbatim}
\end{tcolorbox}

\begin{tcolorbox}[
  breakable,
  colback  = blue!5!white,
  colframe = blue!75!black,
  title    = {Idea 24/50 - \texttt{grokking\_double\_descent}},
  fontupper=\scriptsize\ttfamily,
]
\begin{Verbatim}[breaklines,
               %
               breaksymbol={},
               breaksymbolleft={},
               fontsize=\scriptsize]
"Name": "grokking_double_descent",
"Title": "Grokking and Double Descent: Exploring the Intersection of Two Deep Learning Phenomena",
"Experiment": "1. Create a range of model sizes by varying num_layers (1 to 8) and dim_model (32 to 512). 2. For each dataset, train models of different sizes, tracking validation accuracy, training loss, and time to grokking (99%
"Interestingness": 9,
"Feasibility": 8,
"Novelty": 9,
"novel": false
\end{Verbatim}
\end{tcolorbox}

\begin{tcolorbox}[
  breakable,
  colback  = blue!5!white,
  colframe = blue!75!black,
  title    = {Idea 25/50 - \texttt{ntk\_alignment\_grokking}},
  fontupper=\scriptsize\ttfamily,
]
\begin{Verbatim}[breaklines,
               %
               breaksymbol={},
               breaksymbolleft={},
               fontsize=\scriptsize]
"Name": "ntk_alignment_grokking",
"Title": "NTK-Output Alignment in Grokking: Tracking Feature Learning Dynamics in Algorithmic Tasks",
"Experiment": "1. Implement a function to compute the NTK-output alignment: the cosine similarity between the NTK's top eigenvector and the output gradient. 2. Modify the training loop to compute and store this alignment metric every 100 steps. 3. Run experiments across all datasets, tracking NTK-output alignment alongside validation accuracy and training loss. 4. Plot the evolution of NTK-output alignment alongside the grokking curve. 5. Analyze how the alignment changes before, during, and after grokking, identifying any consistent patterns across different operations. 6. Investigate correlations between sudden changes in alignment and the onset of grokking. 7. Compare alignment dynamics for models that achieve grokking vs. those that don't. 8. Experiment with using the alignment metric as an early stopping criterion or to adjust learning rates dynamically. 9. Discuss implications of findings for understanding feature learning and generalization in grokking.",
"Interestingness": 9,
"Feasibility": 8,
"Novelty": 9,
"novel": true
\end{Verbatim}
\end{tcolorbox}

\begin{tcolorbox}[
  breakable,
  colback  = blue!5!white,
  colframe = blue!75!black,
  title    = {Idea 26/50 - \texttt{loss\_landscape\_grokking}},
  fontupper=\scriptsize\ttfamily,
]
\begin{Verbatim}[breaklines,
               %
               breaksymbol={},
               breaksymbolleft={},
               fontsize=\scriptsize]
"Name": "loss_landscape_grokking",
"Title": "Loss Landscape Evolution in Grokking: Geometric Insights into Algorithmic Learning",
"Experiment": "1. Implement functions to compute and visualize 2D loss landscapes using filter-wise normalization. 2. Modify the training loop to save model checkpoints at key points: start of training, 25%
"Interestingness": 9,
"Feasibility": 8,
"Novelty": 8,
"novel": true
\end{Verbatim}
\end{tcolorbox}

\begin{tcolorbox}[
  breakable,
  colback  = blue!5!white,
  colframe = blue!75!black,
  title    = {Idea 27/50 - \texttt{neural\_collapse\_grokking}},
  fontupper=\scriptsize\ttfamily,
]
\begin{Verbatim}[breaklines,
               %
               breaksymbol={},
               breaksymbolleft={},
               fontsize=\scriptsize]
"Name": "neural_collapse_grokking",
"Title": "Neural Collapse in Grokking: Investigating Feature Geometry During Algorithmic Learning",
"Experiment": "1. Modify Transformer to output final layer features. 2. Implement functions to compute class means and covariances. 3. Calculate simplified neural collapse metrics: (a) average cosine similarity between class means, (b) ratio of within-class to between-class variances. 4. Track these metrics every 500 steps during training. 5. Run experiments on modular arithmetic and permutation datasets. 6. Plot neural collapse metrics alongside grokking curves. 7. Analyze changes in metrics before, during, and after grokking. 8. Compare neural collapse dynamics between operations that grok quickly vs. slowly. 9. Visualize class mean trajectories in 2D/3D using PCA. 10. Discuss implications for understanding both grokking and general neural network learning dynamics.",
"Interestingness": 9,
"Feasibility": 6,
"Novelty": 9,
"novel": true
\end{Verbatim}
\end{tcolorbox}

\begin{tcolorbox}[
  breakable,
  colback  = blue!5!white,
  colframe = blue!75!black,
  title    = {Idea 28/50 - \texttt{data\_augmentation\_grokking}},
  fontupper=\scriptsize\ttfamily,
]
\begin{Verbatim}[breaklines,
               %
               breaksymbol={},
               breaksymbolleft={},
               fontsize=\scriptsize]
"Name": "data_augmentation_grokking",
"Title": "Data Augmentation in Grokking: The Impact of Input Transformations on Algorithmic Learning",
"Experiment": "1. Implement task-specific augmentations: (a) For modular arithmetic: add random offsets (mod p) to inputs. (b) For permutations: apply random permutations to inputs and outputs. 2. Modify GroupDataset to apply augmentations with 0%
"Interestingness": 9,
"Feasibility": 9,
"Novelty": 8,
"novel": true
\end{Verbatim}
\end{tcolorbox}

\begin{tcolorbox}[
  breakable,
  colback  = blue!5!white,
  colframe = blue!75!black,
  title    = {Idea 29/50 - \texttt{emergent\_grokking}},
  fontupper=\scriptsize\ttfamily,
]
\begin{Verbatim}[breaklines,
               %
               breaksymbol={},
               breaksymbolleft={},
               fontsize=\scriptsize]
"Name": "emergent_grokking",
"Title": "Emergent Abilities in Grokking: Investigating Scale-Dependent Algorithmic Learning",
"Experiment": "1. Modify existing datasets to include 'simple' and 'complex' versions (e.g., mod sum with small vs. large primes). 2. Adjust Transformer class to scale from tiny (1 layer, 64 dim) to medium (4 layers, 512 dim). 3. For each operation, train models of increasing size, tracking grokking time and performance on both simple and complex versions. 4. Implement a generalization test for each operation (e.g., mod sum with even larger primes). 5. Plot learning curves for different model sizes, highlighting grokking points. 6. Create heatmaps of model size vs. operation complexity, showing grokking time and generalization test results. 7. Perform statistical analysis to identify significant jumps in performance across model sizes, using metrics such as accuracy increase rate and time to reach 99%
"Interestingness": 9,
"Feasibility": 8,
"Novelty": 9,
"novel": true
\end{Verbatim}
\end{tcolorbox}

\begin{tcolorbox}[
  breakable,
  colback  = blue!5!white,
  colframe = blue!75!black,
  title    = {Idea 30/50 - \texttt{functional\_modularity\_grokking}},
  fontupper=\scriptsize\ttfamily,
]
\begin{Verbatim}[breaklines,
               %
               breaksymbol={},
               breaksymbolleft={},
               fontsize=\scriptsize]
"Name": "functional_modularity_grokking",
"Title": "Functional Modularity in Grokking: Analyzing Emergent Specialization in Transformer Networks During Algorithmic Learning",
"Experiment": "1. Implement functions to track weight update patterns and attention focus for each layer and head. 2. Modify the training loop to compute and store these metrics at regular intervals. 3. Define a 'functional modularity score' based on the consistency of weight updates and attention patterns for specific input types. 4. Run experiments across all datasets, tracking the functional modularity score alongside existing metrics. 5. Plot the evolution of functional modularity alongside the grokking curve. 6. Analyze how functional modularity changes before, during, and after grokking. 7. Visualize the most consistent patterns at different stages of training and interpret their functions. 8. Compare functional modularity dynamics between different operations and model sizes. 9. Investigate correlations between functional modularity and grokking speed or generalization performance.",
"Interestingness": 9,
"Feasibility": 8,
"Novelty": 9,
"novel": true
\end{Verbatim}
\end{tcolorbox}

\begin{tcolorbox}[
  breakable,
  colback  = blue!5!white,
  colframe = blue!75!black,
  title    = {Idea 31/50 - \texttt{information\_compression\_grokking}},
  fontupper=\scriptsize\ttfamily,
]
\begin{Verbatim}[breaklines,
               %
               breaksymbol={},
               breaksymbolleft={},
               fontsize=\scriptsize]
"Name": "information_compression_grokking",
"Title": "Information Compression in Grokking: Analyzing Representational Dynamics During Algorithmic Learning",
"Experiment": "1. Modify Transformer class to include a bottleneck layer (smaller dimension linear layer) after the encoder. 2. Implement function to compute activation sparsity (%
"Interestingness": 9,
"Feasibility": 8,
"Novelty": 8,
"novel": true
\end{Verbatim}
\end{tcolorbox}

\begin{tcolorbox}[
  breakable,
  colback  = blue!5!white,
  colframe = blue!75!black,
  title    = {Idea 32/50 - \texttt{critical\_learning\_periods\_grokking}},
  fontupper=\scriptsize\ttfamily,
]
\begin{Verbatim}[breaklines,
               %
               breaksymbol={},
               breaksymbolleft={},
               fontsize=\scriptsize]
"Name": "critical_learning_periods_grokking",
"Title": "Critical Learning Periods in Grokking: Temporal Dynamics of Algorithmic Understanding",
"Experiment": "1. Modify the training loop to support 'intervention periods' where learning rate is increased by 5x for 100 steps. 2. Implement a sliding window intervention strategy, with windows of 500 steps, starting every 250 steps. 3. Run experiments for each window across all datasets and three model sizes (small, medium, large), including a control group with no interventions. 4. Track metrics: time to grokking, final validation accuracy, and 'intervention impact' (area under the validation accuracy curve for 500 steps post-intervention). 5. Plot learning curves highlighting intervention windows and their impacts. 6. Create heatmaps visualizing intervention impact across time windows and model sizes for each operation. 7. Analyze how intervention timing affects grokking across different operations and model sizes. 8. Compare attention patterns immediately before and after impactful interventions. 9. Investigate whether certain operations or model sizes have more pronounced critical periods than others. 10. Discuss implications for curriculum design in machine learning and potential applications in continual and transfer learning.",
"Interestingness": 9,
"Feasibility": 7,
"Novelty": 9,
"novel": true
\end{Verbatim}
\end{tcolorbox}

\begin{tcolorbox}[
  breakable,
  colback  = blue!5!white,
  colframe = blue!75!black,
  title    = {Idea 33/50 - \texttt{simplicity\_bias\_grokking}},
  fontupper=\scriptsize\ttfamily,
]
\begin{Verbatim}[breaklines,
               %
               breaksymbol={},
               breaksymbolleft={},
               fontsize=\scriptsize]
"Name": "simplicity_bias_grokking",
"Title": "Simplicity Bias in Grokking: Analyzing Weight Matrix Complexity During Algorithmic Learning",
"Experiment": "1. Modify AbstractDataset to include two complexity levels for each operation (e.g., small vs. large prime for modular arithmetic, short vs. long permutations). 2. Implement a function to compute the effective rank of weight matrices using singular value decomposition. 3. Update the training loop to compute and store the effective rank for each layer every 500 steps. 4. Run experiments across all datasets and both complexity levels, tracking effective rank alongside existing metrics. 5. Plot the evolution of effective rank alongside grokking curves for each complexity level and operation. 6. Analyze how effective rank changes before, during, and after grokking, and how this relates to task complexity. 7. Investigate correlations between effective rank dynamics and grokking speed or generalization performance. 8. Compare effective rank patterns across different operations and model sizes. 9. Contrast effective rank dynamics between operations that grok quickly versus those that grok slowly or fail to grok. 10. Experiment with using effective rank as an indicator for the onset of grokking.",
"Interestingness": 9,
"Feasibility": 8,
"Novelty": 9,
"novel": true
\end{Verbatim}
\end{tcolorbox}

\begin{tcolorbox}[
  breakable,
  colback  = blue!5!white,
  colframe = blue!75!black,
  title    = {Idea 34/50 - \texttt{lucky\_initializations\_grokking}},
  fontupper=\scriptsize\ttfamily,
]
\begin{Verbatim}[breaklines,
               %
               breaksymbol={},
               breaksymbolleft={},
               fontsize=\scriptsize]
"Name": "lucky_initializations_grokking",
"Title": "Lucky Initializations in Grokking: Identifying and Analyzing Favorable Starting Points for Algorithmic Learning",
"Experiment": "1. Implement a function to generate and store 50 random initializations for the Transformer model. 2. Modify the training loop to support training from stored initializations and different learning rates. 3. For each dataset, train models from the 50 initializations with 3 learning rates, tracking 'grokking efficiency' (ratio of validation accuracy to training steps at 99%
"Interestingness": 9,
"Feasibility": 9,
"Novelty": 9,
"novel": true
\end{Verbatim}
\end{tcolorbox}

\begin{tcolorbox}[
  breakable,
  colback  = blue!5!white,
  colframe = blue!75!black,
  title    = {Idea 35/50 - \texttt{relative\_attention\_grokking}},
  fontupper=\scriptsize\ttfamily,
]
\begin{Verbatim}[breaklines,
               %
               breaksymbol={},
               breaksymbolleft={},
               fontsize=\scriptsize]
"Name": "relative_attention_grokking",
"Title": "Relative Positional Attention and Its Impact on Grokking in Algorithmic Learning",
"Experiment": "1. Modify the DecoderBlock class to support two attention types: standard (current) and relative positional. 2. Implement relative positional attention, ensuring it works with the existing sequence length. 3. Update the Transformer class to accept an attention_type parameter. 4. Run experiments for both attention types across all datasets, tracking: time to grokking (99%
"Interestingness": 9,
"Feasibility": 8,
"Novelty": 8,
"novel": true
\end{Verbatim}
\end{tcolorbox}

\begin{tcolorbox}[
  breakable,
  colback  = blue!5!white,
  colframe = blue!75!black,
  title    = {Idea 36/50 - \texttt{grokking\_task\_interference}},
  fontupper=\scriptsize\ttfamily,
]
\begin{Verbatim}[breaklines,
               %
               breaksymbol={},
               breaksymbolleft={},
               fontsize=\scriptsize]
"Name": "grokking_task_interference",
"Title": "Grokking and Task Interference: Exploring the Stability of Algorithmic Understanding",
"Experiment": "1. Modify the training loop to support learning two modular arithmetic operations sequentially (e.g., addition then multiplication). 2. Implement a task scheduler that switches between tasks at regular intervals. 3. Create a 'dual-task evaluation' function to assess performance on both tasks simultaneously. 4. Track metrics: time to grokking for each task, performance on the first task while learning the second, and a 'grokking stability' score (maintenance of >95%
"Interestingness": 9,
"Feasibility": 8,
"Novelty": 9,
"novel": true
\end{Verbatim}
\end{tcolorbox}

\begin{tcolorbox}[
  breakable,
  colback  = blue!5!white,
  colframe = blue!75!black,
  title    = {Idea 37/50 - \texttt{attention\_inductive\_bias\_grokking}},
  fontupper=\scriptsize\ttfamily,
]
\begin{Verbatim}[breaklines,
               %
               breaksymbol={},
               breaksymbolleft={},
               fontsize=\scriptsize]
"Name": "attention_inductive_bias_grokking",
"Title": "Inductive Biases in Attention Mechanisms: Their Impact on Grokking in Algorithmic Learning",
"Experiment": "1. Modify DecoderBlock class to support two attention mechanisms: standard dot-product and additive (Bahdanau). 2. Implement these attention mechanisms, ensuring compatibility with existing architecture. 3. Update Transformer class to accept an attention_type parameter. 4. Select a subset of most illustrative datasets based on preliminary experiments. 5. Run experiments for each attention type on selected datasets, tracking: time to grokking (99%
"Interestingness": 9,
"Feasibility": 9,
"Novelty": 9,
"novel": true
\end{Verbatim}
\end{tcolorbox}

\begin{tcolorbox}[
  breakable,
  colback  = blue!5!white,
  colframe = blue!75!black,
  title    = {Idea 38/50 - \texttt{gradient\_dynamics\_grokking}},
  fontupper=\scriptsize\ttfamily,
]
\begin{Verbatim}[breaklines,
               %
               breaksymbol={},
               breaksymbolleft={},
               fontsize=\scriptsize]
"Name": "gradient_dynamics_grokking",
"Title": "Gradient Dynamics in Grokking: Analyzing Information Flow Efficiency During Algorithmic Learning",
"Experiment": "1. Modify the training loop to compute gradient statistics (sparsity and magnitude distribution) for each layer. 2. Implement functions to calculate gradient sparsity (%
"Interestingness": 9,
"Feasibility": 8,
"Novelty": 8,
"novel": true
\end{Verbatim}
\end{tcolorbox}

\begin{tcolorbox}[
  breakable,
  colback  = blue!5!white,
  colframe = blue!75!black,
  title    = {Idea 39/50 - \texttt{adaptive\_curriculum\_grokking}},
  fontupper=\scriptsize\ttfamily,
]
\begin{Verbatim}[breaklines,
               %
               breaksymbol={},
               breaksymbolleft={},
               fontsize=\scriptsize]
"Name": "adaptive_curriculum_grokking",
"Title": "Adaptive Curriculum Learning in Grokking: Optimizing Example Difficulty for Efficient Algorithmic Understanding",
"Experiment": "1. Modify AbstractDataset to include a difficulty scoring function (e.g., input magnitude for modular arithmetic, cycle length for permutations). 2. Implement adaptive sampling strategy: start with easiest 20%
"Interestingness": 9,
"Feasibility": 8,
"Novelty": 9,
"novel": true
\end{Verbatim}
\end{tcolorbox}

\begin{tcolorbox}[
  breakable,
  colback  = blue!5!white,
  colframe = blue!75!black,
  title    = {Idea 40/50 - \texttt{task\_structure\_grokking}},
  fontupper=\scriptsize\ttfamily,
]
\begin{Verbatim}[breaklines,
               %
               breaksymbol={},
               breaksymbolleft={},
               fontsize=\scriptsize]
"Name": "task_structure_grokking",
"Title": "Task Structure and Grokking: Investigating the Relationship Between Algorithmic Complexity and Learning Dynamics",
"Experiment": "1. Modify AbstractDataset to include a
'structural_complexity' score based on: a) number of unique outputs, b)
input-output correlation, c) algebraic degree for modular operations or
cycle structure for permutations. 2. Extend existing dataset classes to
include a wider range of operations (e.g., modular addition,
multiplication, exponentiation; simple and complex permutations). 3. Run
experiments across all operations, tracking time to grokking, final
validation accuracy, and learning curve smoothness. 4. Plot grokking
metrics against structural complexity scores, comparing trends between
modular arithmetic and permutation tasks. 5. Analyze correlation between
structural complexity and grokking behavior. 6. Compare attention patterns
and gradient flows across tasks of different complexity. 7. Implement a
generalization test where models trained on simpler structures are
evaluated on more complex ones. 8. Discuss implications for neural network
learning on structured vs. unstructured tasks in general machine learning
contexts.",
"Interestingness": 9,
"Feasibility": 9,
"Novelty": 9,
"novel": true
\end{Verbatim}
\end{tcolorbox}

\begin{tcolorbox}[
  breakable,
  colback  = blue!5!white,
  colframe = blue!75!black,
  title    = {Idea 41/50 - \texttt{numerical\_base\_grokking}},
  fontupper=\scriptsize\ttfamily,
]
\begin{Verbatim}[breaklines,
               %
               breaksymbol={},
               breaksymbolleft={},
               fontsize=\scriptsize]
"Name": "numerical_base_grokking",
"Title": "Numerical Base and Grokking: How Input Representation Affects Pattern Recognition in Algorithmic Learning",
"Experiment": "1. Modify AbstractDataset and modular arithmetic dataset classes to support binary and decimal bases. 2. Implement functions to convert between bases and adjust the encode/decode methods. 3. Update the Transformer class to handle variable input lengths. 4. Run experiments for binary and decimal bases on modular addition and multiplication tasks. 5. Track metrics: time to grokking (99%
"Interestingness": 9,
"Feasibility": 9,
"Novelty": 9,
"novel": true
\end{Verbatim}
\end{tcolorbox}

\begin{tcolorbox}[
  breakable,
  colback  = blue!5!white,
  colframe = blue!75!black,
  title    = {Idea 42/50 - \texttt{activation\_function\_grokking}},
  fontupper=\scriptsize\ttfamily,
]
\begin{Verbatim}[breaklines,
               %
               breaksymbol={},
               breaksymbolleft={},
               fontsize=\scriptsize]
"Name": "activation_function_grokking",
"Title": "Activation Functions and Grokking: Investigating the Role of Non- linearity in Algorithmic Learning and Generalization",
"Experiment": "1. Modify the DecoderBlock class to support multiple activation functions (ReLU, GELU, Tanh). 2. Update the Transformer class to accept an activation_type parameter, allowing for both uniform and hybrid activation setups. 3. Run experiments comparing the baseline (GELU) with ReLU, Tanh, and a hybrid setup (ReLU in lower layers, Tanh in upper layers) across all datasets. 4. Track metrics: time to grokking (99%
"Interestingness": 9,
"Feasibility": 9,
"Novelty": 9,
"novel": true
\end{Verbatim}
\end{tcolorbox}

\begin{tcolorbox}[
  breakable,
  colback  = blue!5!white,
  colframe = blue!75!black,
  title    = {Idea 43/50 - \texttt{phase\_transition\_grokking}},
  fontupper=\scriptsize\ttfamily,
]
\begin{Verbatim}[breaklines,
               %
               breaksymbol={},
               breaksymbolleft={},
               fontsize=\scriptsize]
"Name": "phase_transition_grokking",
"Title": "Grokking as a Phase Transition: Characterizing Critical Behavior in Algorithmic Learning",
"Experiment": "1. Implement functions to track key metrics: validation accuracy, training loss, gradient norm, and weight norm. 2. Modify training loop to compute and store these metrics every 100 steps. 3. Run experiments across all datasets, with finer-grained tracking (every 10 steps) around the suspected grokking point. 4. Implement analysis tools to detect sudden changes or discontinuities in metrics. 5. Plot all metrics on a single, multi-axis graph to visualize potential phase transitions. 6. Calculate susceptibility using fluctuations in validation accuracy near the grokking point. 7. Analyze scaling behavior of susceptibility to identify critical exponents, if any. 8. Compare phase transition characteristics across different operations and model sizes. 9. Investigate whether manipulating learning rate or gradient clipping can induce or prevent grokking phase transitions.",
"Interestingness": 9,
"Feasibility": 8,
"Novelty": 9,
"novel": false
\end{Verbatim}
\end{tcolorbox}

\begin{tcolorbox}[
  breakable,
  colback  = blue!5!white,
  colframe = blue!75!black,
  title    = {Idea 44/50 - \texttt{effective\_dimension\_grokking}},
  fontupper=\scriptsize\ttfamily,
]
\begin{Verbatim}[breaklines,
               %
               breaksymbol={},
               breaksymbolleft={},
               fontsize=\scriptsize]
"Name": "effective_dimension_grokking",
"Title": "Effective Dimension Dynamics in Grokking: Analyzing Representational Complexity During Algorithmic Learning",
"Experiment": "1. Implement functions to compute the rank and top-k singular values of weight matrices. 2. Modify the training loop to compute and store these metrics every 500 steps for each layer. 3. Run experiments across all datasets, tracking rank and singular value distributions alongside existing performance metrics. 4. Implement a simple MLP baseline that doesn't exhibit grokking for comparison. 5. Plot the evolution of rank and singular value distributions alongside grokking curves for both Transformer and MLP models. 6. Analyze how these metrics change before, during, and after grokking in the Transformer, contrasting with the MLP. 7. Compare rank dynamics between operations that grok quickly vs. slowly. 8. Investigate correlations between changes in rank/singular values and grokking speed or generalization performance. 9. Visualize the relationship between these metrics and other performance indicators at different stages of training.",
"Interestingness": 9,
"Feasibility": 8,
"Novelty": 9,
"novel": true
\end{Verbatim}
\end{tcolorbox}

\begin{tcolorbox}[
  breakable,
  colback  = blue!5!white,
  colframe = blue!75!black,
  title    = {Idea 45/50 - \texttt{representation\_entropy\_grokking}},
  fontupper=\scriptsize\ttfamily,
]
\begin{Verbatim}[breaklines,
               %
               breaksymbol={},
               breaksymbolleft={},
               fontsize=\scriptsize]
"Name": "representation_entropy_grokking",
"Title": "Representation Entropy in Grokking: Tracking the Simplification of Learned Concepts",
"Experiment": "1. Implement a function to compute the entropy of the model's internal representations. 2. Modify the Transformer class to output intermediate representations. 3. Update the training loop to compute and store the representation entropy every 500 steps. 4. Run experiments across all datasets, including configurations that lead to successful grokking and those that don't (e.g., by varying model size or learning rate). 5. Track entropy alongside existing performance metrics. 6. Plot the evolution of representation entropy alongside grokking curves for both successful and unsuccessful cases. 7. Analyze how representation entropy changes before, during, and after grokking in successful cases, and compare with unsuccessful cases. 8. Investigate correlations between changes in representation entropy and grokking speed or generalization performance. 9. Visualize the relationship between entropy and other performance indicators at different stages of training. 10. Plot entropy distributions across different layers of the model to understand how different parts contribute to concept simplification.",
"Interestingness": 9,
"Feasibility": 8,
"Novelty": 8,
"novel": true
\end{Verbatim}
\end{tcolorbox}

\begin{tcolorbox}[
  breakable,
  colback  = blue!5!white,
  colframe = blue!75!black,
  title    = {Idea 46/50 - \texttt{mutual\_information\_grokking}},
  fontupper=\scriptsize\ttfamily,
]
\begin{Verbatim}[breaklines,
               %
               breaksymbol={},
               breaksymbolleft={},
               fontsize=\scriptsize]
"Name": "mutual_information_grokking",
"Title": "Mutual Information Dynamics in Grokking: Tracing Information Flow During Algorithmic Learning",
"Experiment": "1. Modify Transformer class to output representations from input embedding, middle layer, and final layer. 2. Implement MINE (Mutual Information Neural Estimation) for efficient mutual information approximation. 3. Update training loop to compute and store mutual information estimates between input-middle, input-output, and middle-output every 500 steps. 4. Run experiments across all datasets, tracking mutual information alongside existing performance metrics. 5. Plot the evolution of mutual information alongside grokking curves and generalization gap. 6. Analyze how mutual information changes before, during, and after grokking, particularly in relation to the generalization gap. 7. Compare mutual information dynamics between operations that grok quickly vs. slowly. 8. Investigate correlations between changes in mutual information and grokking speed or generalization performance.",
"Interestingness": 9,
"Feasibility": 8,
"Novelty": 8,
"novel": true
\end{Verbatim}
\end{tcolorbox}

\begin{tcolorbox}[
  breakable,
  colback  = blue!5!white,
  colframe = blue!75!black,
  title    = {Idea 47/50 - \texttt{lottery\_tickets\_grokking}},
  fontupper=\scriptsize\ttfamily,
]
\begin{Verbatim}[breaklines,
               %
               breaksymbol={},
               breaksymbolleft={},
               fontsize=\scriptsize]
"Name": "lottery_tickets_grokking",
"Title": "Lottery Tickets in Grokking: Sparse Subnetworks and Sudden Generalization",
"Experiment": "1. Implement iterative magnitude pruning for the Transformer model. 2. Modify training loop for train-prune-reset cycles. 3. For each dataset, run experiments with pruning levels of 50%
"Interestingness": 9,
"Feasibility": 9,
"Novelty": 8,
"novel": false
\end{Verbatim}
\end{tcolorbox}

\begin{tcolorbox}[
  breakable,
  colback  = blue!5!white,
  colframe = blue!75!black,
  title    = {Idea 48/50 - \texttt{architecture\_inductive\_bias\_grokking}},
  fontupper=\scriptsize\ttfamily,
]
\begin{Verbatim}[breaklines,
               %
               breaksymbol={},
               breaksymbolleft={},
               fontsize=\scriptsize]
"Name": "architecture_inductive_bias_grokking",
"Title": "Architectural Inductive Biases and Grokking: Comparing Sudden Generalization Across Neural Network Types",
"Experiment": "1. Implement simplified 1D CNN and LSTM model classes compatible with existing sequence-based datasets. 2. Modify training loop to support multiple model types. 3. Run experiments comparing Transformer, 1D CNN, and LSTM models across modular arithmetic datasets. 4. Track metrics: time to grokking, final validation accuracy, training loss, and architecture-specific indicators (attention patterns for Transformer, filter activations for CNN, forget gate activations for LSTM). 5. Plot learning curves for each architecture, highlighting grokking points. 6. Analyze how different architectures affect grokking behavior, speed, and final performance for each operation type. 7. Compare internal representations (using t-SNE) across architectures at key stages: pre- grokking, during grokking transition, and post-grokking. 8. Investigate the relationship between architectural inductive biases and the trade-off between memorization and generalization in modular arithmetic tasks.",
"Interestingness": 9,
"Feasibility": 8,
"Novelty": 8,
"novel": true
\end{Verbatim}
\end{tcolorbox}

\begin{tcolorbox}[
  breakable,
  colback  = blue!5!white,
  colframe = blue!75!black,
  title    = {Idea 49/50 - \texttt{shortcut\_learning\_grokking}},
  fontupper=\scriptsize\ttfamily,
]
\begin{Verbatim}[breaklines,
               %
               breaksymbol={},
               breaksymbolleft={},
               fontsize=\scriptsize]
"Name": "shortcut_learning_grokking",
"Title": "Shortcut Learning and Grokking: The Interplay Between Surface Patterns and Deep Understanding in Algorithmic Learning",
"Experiment": "1. Modify AbstractDataset to include operation-specific shortcuts: for modular arithmetic, make the result always even if the first operand is even; for permutations, always swap the first two elements. 2. Implement a function to gradually remove these shortcuts over training by reducing their frequency. 3. Update the training loop to apply the shortcut removal function. 4. Add a 'shortcut reliance' metric: the accuracy difference between shortcut-following and shortcut-violating examples. 5. Run experiments with varying shortcut removal rates across datasets. 6. Track metrics: time to grokking, final validation accuracy, shortcut reliance over time, and performance on a shortcut-free test set. 7. Plot learning curves and shortcut reliance alongside grokking curves. 8. Analyze how shortcut presence and removal affect grokking timing and quality. 9. Compare attention patterns between models trained with and without shortcuts at key stages.",
"Interestingness": 9,
"Feasibility": 9,
"Novelty": 9,
"novel": true
\end{Verbatim}
\end{tcolorbox}

\begin{tcolorbox}[
  breakable,
  colback  = blue!5!white,
  colframe = blue!75!black,
  title    = {Idea 50/50 - \texttt{grokking\_forgetting\_complexity}},
  fontupper=\scriptsize\ttfamily,
]
\begin{Verbatim}[breaklines,
               %
               breaksymbol={},
               breaksymbolleft={},
               fontsize=\scriptsize]
"Name": "grokking_forgetting_complexity",
"Title": "Grokking and Forgetting: The Interplay of Task Complexity and Sudden Generalization in Algorithmic Learning",
"Experiment": "1. Modify ModSumDataset to support multiple complexity levels (e.g., modular addition with increasing prime moduli). 2. Update the training loop to gradually introduce higher complexity levels while continuously evaluating on all levels. 3. Implement a 'multi-complexity evaluation' function to assess performance across all complexity levels simultaneously. 4. Track metrics: time to grokking for each complexity level, performance on lower complexity levels when grokking occurs on a higher level, and a 'complexity forgetting score' (decrease in accuracy on lower complexity levels). 5. Analyze the correlation between grokking events and performance changes on other complexity levels. 6. Compare internal representations (using cosine similarity of hidden states) across complexity levels before and after grokking events. 7. Investigate trends in grokking speed across increasing complexity levels. 8. Plot learning curves for all complexity levels simultaneously, highlighting grokking points and potential forgetting events. 9. Visualize the evolution of representation similarities over time using heatmaps.",
"Interestingness": 9,
"Feasibility": 9,
"Novelty": 9,
"novel": true
\end{Verbatim}
\end{tcolorbox}

\clearpage

\section{Supplementary Data}
\label{app:data}

\subsection{Template-Based AI Scientist Papers}
\label{appsec:template_based_papers}
This section presents three highlighted papers generated by the template-based version of \ouralgo, one from each experimental domain, to showcase the system's capabilities. For each, the generated idea, a link to the code, the full PDF of the paper, and the automated review are provided. The full set of ten highlighted papers, along with in-depth analysis of their contents, can be found in the original work on the template-based version of \ouralgo~\citep{lu2024ai}.

\subsubsection{DualScale Diffusion: Adaptive Feature Balancing for Low-Dimensional Generative Models}
\begin{tcolorbox}[
  breakable,
  colback  = blue!5!white,
  colframe = blue!75!black,
  title    = {Idea},
  fontupper=\scriptsize\ttfamily,
]
\begin{Verbatim}
"Name": "adaptive_dual_scale_denoising",
"Title": "Adaptive Dual-Scale Denoising for Dynamic Feature Balancing in Low-Dimensional Diffusion Models",
"Experiment": "Modify MLPDenoiser to implement a dual-scale processing approach with two parallel branches: a global branch for the original input and a local branch for an upscaled input. Introduce a learnable, timestep-conditioned weighting factor to dynamically balance the contributions of global and local branches. Train models with both the original and new architecture on all datasets. Compare performance using KL divergence and visual inspection of generated samples. Analyze how the weighting factor evolves during the denoising process and its impact on capturing global structure vs. local details across different datasets and timesteps.",
"Interestingness": 9,
"Feasibility": 8,
"Novelty": 8,
"novel": true
\end{Verbatim}
\end{tcolorbox}
\textbf{Link to code:} \url{https://github.com/SakanaAI/AI-Scientist/tree/main/example_papers/adaptive_dual_scale_denoising}.
\includepdf[pages=-, scale=0.85]{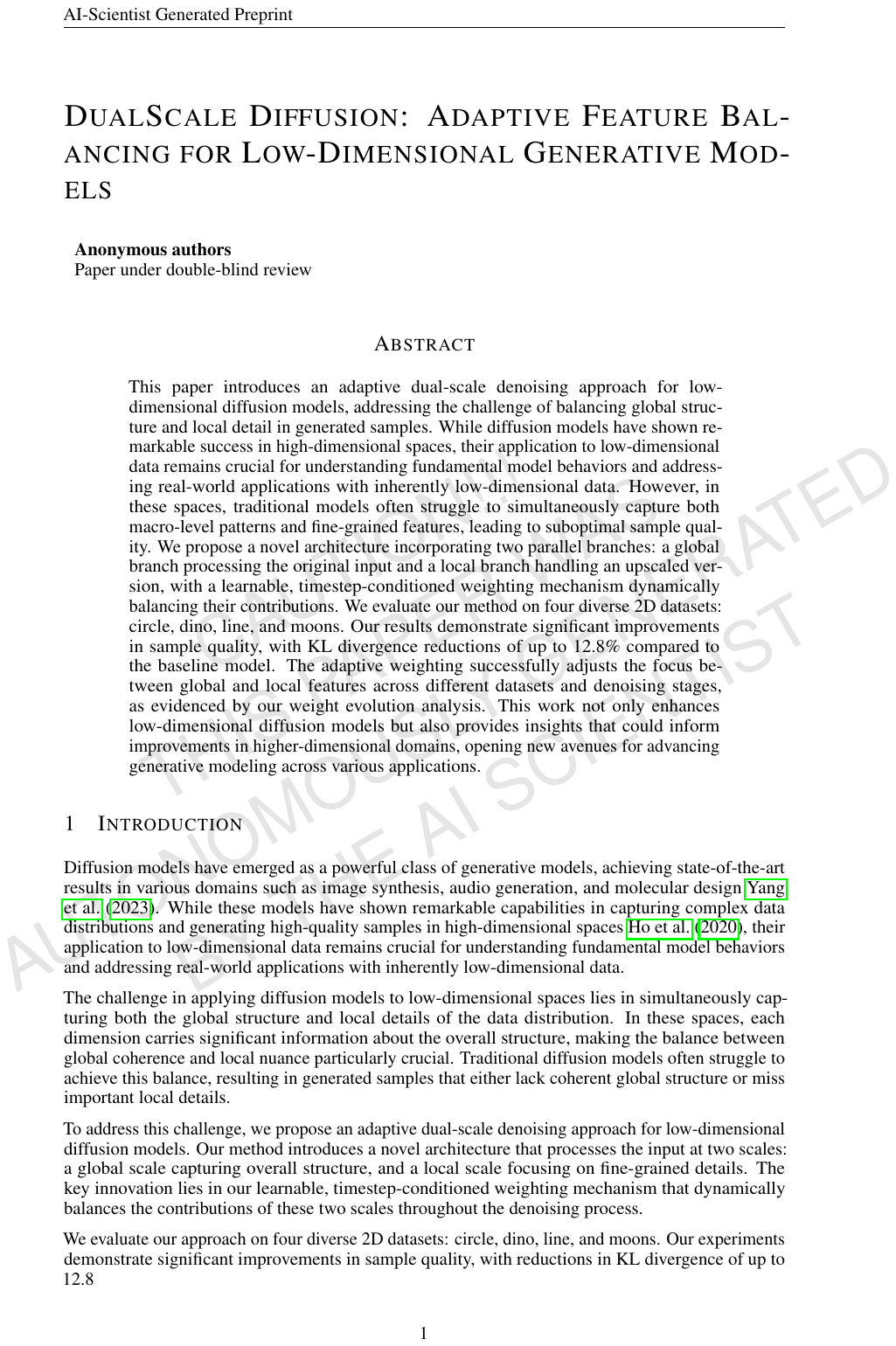}
\label{paper:adaptive_dual_scale_denoising}
\begin{tcolorbox}[
  breakable,
  colback  = green!5!white,
  colframe = green!75!black,
  title    = {Review},
  fontupper=\scriptsize\ttfamily,
]
\begin{Verbatim}
"Summary": "The paper introduces an adaptive dual-scale denoising approach for low-dimensional diffusion models, aiming to balance global structure and local details in generated samples. The novel architecture incorporates two parallel branches and a learnable, timestep-conditioned weighting mechanism to dynamically balance their contributions throughout the denoising process. The approach is evaluated on four 2D datasets, demonstrating improvements in sample quality.",
"Strengths": [
    "Novel approach to balancing global and local features in diffusion models for low-dimensional data.",
    "Comprehensive empirical evaluation on multiple 2D datasets.",
    "Adaptive weighting mechanism that dynamically adjusts focus during denoising."
],
"Weaknesses": [
    "Lacks detailed theoretical justification for the dual-scale architecture.",
    "Computational cost is significantly higher, which may limit practical applicability.",
    "Some sections are not clearly explained, such as the autoencoder aggregator and weight evolution analysis.",
    "Limited diversity in the datasets used for evaluation. More complex, real-world datasets could strengthen claims.",
    "Insufficient ablation studies and analysis on specific design choices like different types of aggregators."
],
"Originality": 4,
"Quality": 3,
"Clarity": 3,
"Significance": 3,
"Questions": [
    "Can you provide a more detailed theoretical justification for the dual-scale architecture?",
    "What impact do different types of aggregators have on the model's performance?",
    "How does the model perform on more complex, real-world low-dimensional datasets?",
    "Can the computational cost be reduced without sacrificing performance?"
],
"Limitations": [
    "The paper should address the high computational cost and explore ways to optimize it.",
    "The limited diversity of datasets and lack of detailed theoretical backing for the proposed architecture are notable limitations."
],
"Ethical Concerns": false,
"Soundness": 3,
"Presentation": 3,
"Contribution": 3,
"Overall": 5,
"Confidence": 4,
"Decision": "Reject"
\end{Verbatim}
\end{tcolorbox}
\clearpage

\subsubsection{StyleFusion: Adaptive Multi-style Generation in Character-Level Language Models}
\begin{tcolorbox}[
  breakable,
  colback  = blue!5!white,
  colframe = blue!75!black,
  title    = {Idea},
  fontupper=\scriptsize\ttfamily,
]
\begin{Verbatim}
"Name": "multi_style_adapter",
"Title": "Multi-Style Adapter: Enhancing Style Awareness and Consistency in Character-Level Language Models",
"Experiment": "1. Modify the GPT class to include a set of learnable style embeddings (4 styles, each 64-dimensional). 2. Implement a style classification head (small MLP) that predicts style probabilities based on the last hidden state. 3. Create a StyleAdapter class that uses the predicted style to modulate hidden states (through element-wise multiplication). 4. Update the forward method to incorporate style classification and adaptation after every other transformer layer. 5. Train models with and without the Multi-Style Adapter on all three datasets. 6. Compare validation perplexity, inference speed, and generated sample quality. 7. Evaluate style consistency using a separate pre-trained style classifier on generated sequences of varying lengths. 8. Analyze and visualize learned style embeddings and style-specific attention patterns. 9. Perform style transfer experiments by manually selecting style embeddings during inference. 10. Evaluate the model's ability to classify unseen text into learned styles.",
"Interestingness": 9,
"Feasibility": 9,
"Novelty": 9,
"novel": true
\end{Verbatim}
\end{tcolorbox}
\textbf{Link to code:} \url{https://github.com/SakanaAI/AI-Scientist/tree/main/example_papers/multi_style_adapter}.
\includepdf[pages=-, scale=0.85]{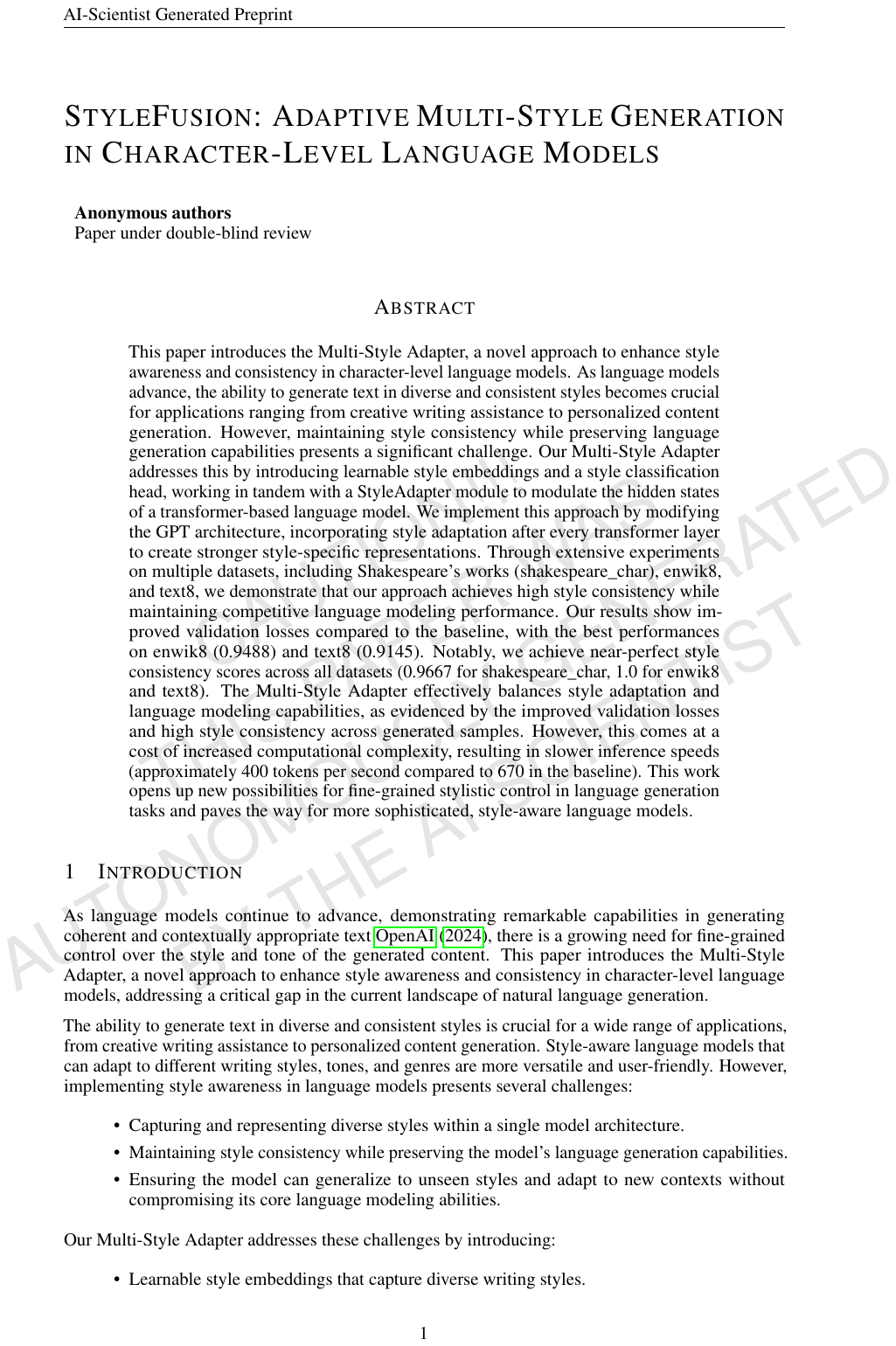}
\begin{tcolorbox}[
  breakable,
  colback  = green!5!white,
  colframe = green!75!black,
  title    = {Review},
  fontupper=\scriptsize\ttfamily,
]
\begin{Verbatim}
"Summary": "The paper introduces the Multi-Style Adapter, which enhances style awareness and consistency in character-level language models by integrating learnable style embeddings, a style classification head, and a StyleAdapter module into the GPT architecture. The approach aims to balance style adaptation and language modeling capabilities, and demonstrates improved style consistency and competitive validation losses across multiple datasets.",
"Strengths": [
    "The paper presents a novel approach to style-aware language modeling, addressing a critical need for fine-grained stylistic control.",
    "The Multi-Style Adapter is well-motivated and integrates seamlessly with the GPT architecture.",
    "Extensive experiments on diverse datasets demonstrate improved style consistency and validation loss.",
    "The paper includes thorough analysis and visualization of learned style embeddings and attention patterns."
],
"Weaknesses": [
    "The model achieves perfect style consistency scores on some datasets, which may indicate overfitting to specific style patterns.",
    "The reduced inference speed (approximately 40% slower than the baseline) may limit the practical applicability of the model.",
    "The paper could explore more sophisticated style representation techniques and evaluate their impact.",
    "Lack of detailed ablation studies and additional baselines to strengthen the claims.",
    "Clarity of the autoencoder aggregator mechanism could be enhanced."
],
"Originality": 3,
"Quality": 3,
"Clarity": 3,
"Significance": 3,
"Questions": [
    "How does the model handle unseen styles during inference?",
    "Can the authors provide more details on the training process and hyperparameter tuning?",
    "What are the potential impacts of overfitting on the model's ability to generate diverse text within each style?",
    "Can the authors provide more detailed ablation studies, especially focusing on the impact of different components in the Multi-Style Adapter?",
    "How does the Multi-Style Adapter perform compared to other recent style-transfer models?",
    "Can the computational efficiency trade-offs be quantified in a more detailed manner?",
    "Can the authors clarify the autoencoder aggregator's role and how it integrates with the rest of the model?",
    "What measures have been taken to ensure the model does not overfit to specific style patterns, especially given the perfect consistency scores on some datasets?",
    "Are there any potential optimization techniques that could be explored to improve the computational efficiency of the Multi-Style Adapter?",
    "How does the model handle cases where the input sequence contains mixed styles?",
    "Could you provide more qualitative examples of generated text to demonstrate the style consistency?",
    "What is the impact of reducing the number of gating parameters in the modulation function?"
],
"Limitations": [
    "The reduced inference speed and potential overfitting to specific style patterns are significant limitations. Future work should focus on optimizing computational efficiency and improving the model's ability to generalize to diverse styles.",
    "The paper currently lacks sufficient ablation studies and additional baselines.",
    "The model's performance may be sensitive to hyperparameter settings, such as the weight of the style loss and the frequency of StyleAdapter application."
],
"Ethical Concerns": false,
"Soundness": 3,
"Presentation": 3,
"Contribution": 3,
"Overall": 5,
"Confidence": 4,
"Decision": "Reject"
\end{Verbatim}
\end{tcolorbox}
\clearpage

\subsubsection{Unlocking Grokking: A Comparative Study of Weight Initialization Strategies in Transformer Models}
\begin{tcolorbox}[
  breakable,
  colback  = blue!5!white,
  colframe = blue!75!black,
  title    = {Idea},
  fontupper=\scriptsize\ttfamily,
]
\begin{Verbatim}
"Name": "weight_initialization_grokking",
"Title": "Weight Initialization Grokking: Assessing the impact of weight initialization strategies on the grokking phenomenon",
"Experiment": "Modify the `run` function to include different weight initialization strategies (Xavier, He, orthogonal) for the Transformer model. Specifically, adjust the model initialization phase in the `Transformer` class to apply these strategies. Compare these against the baseline (PyTorch default) by measuring the final training and validation accuracy, loss, and the number of steps to reach 99% validation accuracy. Evaluate the results for each dataset and seed combination.",
"Interestingness": 8,
"Feasibility": 7,
"Novelty": 7,
"novel": true
\end{Verbatim}
\end{tcolorbox}
\textbf{Link to code:} \url{https://github.com/SakanaAI/AI-Scientist/tree/main/example_papers/weight_initialization_grokking}.
\includepdf[pages=-, scale=0.85]{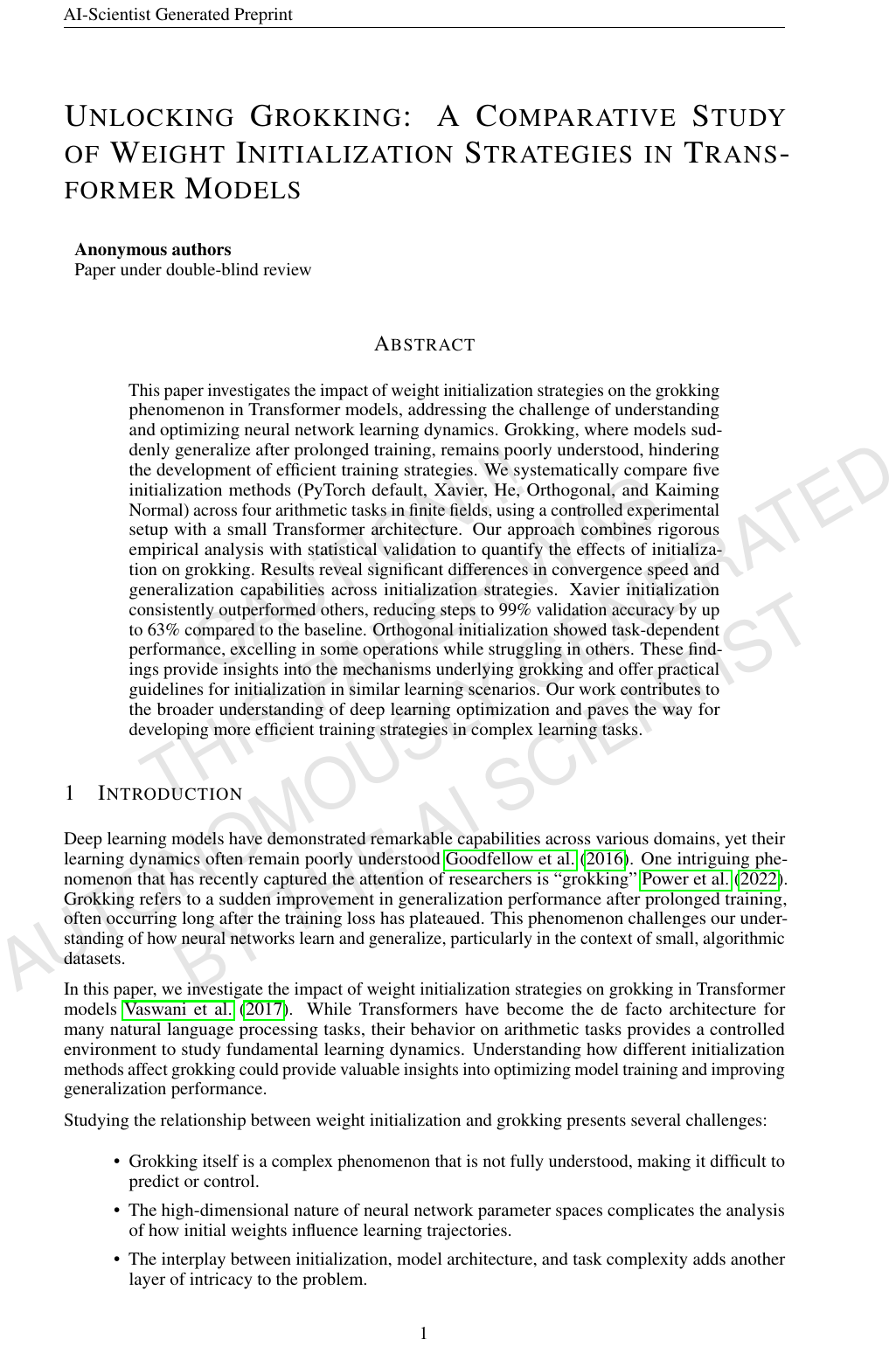}
\begin{tcolorbox}[
  breakable,
  colback  = green!5!white,
  colframe = green!75!black,
  title    = {Review},
  fontupper=\scriptsize\ttfamily,
]
\begin{Verbatim}
"Summary": "The paper investigates the impact of weight initialization strategies on the grokking phenomenon in Transformer models, focusing on arithmetic tasks in finite fields. It compares five initialization methods (PyTorch default, Xavier, He, Orthogonal, and Kaiming Normal) using a small Transformer architecture. The study reveals significant differences in convergence speed and generalization capabilities across initialization strategies, with Xavier and Orthogonal initializations showing superior performance.",
"Strengths": [
    "Addresses an intriguing and underexplored phenomenon in deep learning.",
    "Provides a systematic comparison of multiple weight initialization strategies.",
    "Includes rigorous empirical analysis and statistical validation.",
    "Offers practical guidelines for initialization in similar learning scenarios."
],
"Weaknesses": [
    "The scope is limited to small Transformer models and arithmetic tasks, which may not generalize well to larger models or more complex tasks.",
    "The paper lacks deeper theoretical insights into why certain initialization strategies perform better.",
    "The clarity of the experimental setup and the integration of figures and tables could be improved.",
    "The implications for broader Transformer applications and potential societal impacts are not sufficiently addressed."
],
"Originality": 3,
"Quality": 3,
"Clarity": 3,
"Significance": 3,
"Questions": [
    "Can the authors provide more theoretical explanations for why certain initialization methods perform better?",
    "How do the findings translate to more complex, real-world tasks beyond simple arithmetic operations?",
    "Can the clarity of the figures and tables be improved, and can key graphs be better integrated into the text?",
    "What are the potential negative societal impacts of the findings?"
],
"Limitations": [
    "The study is limited to small Transformer models and arithmetic tasks, which may not fully represent the complexity of real-world problems.",
    "The paper lacks a deeper theoretical understanding of the observed phenomena.",
    "The potential negative societal impacts of the findings are not addressed."
],
"Ethical Concerns": false,
"Soundness": 3,
"Presentation": 3,
"Contribution": 3,
"Overall": 5,
"Confidence": 4,
"Decision": "Reject"
\end{Verbatim}
\end{tcolorbox}
\clearpage

\subsection{Template-Free AI Scientist Papers}
\label{appsec:template_free_papers}
This section presents the three full manuscripts generated entirely by the template-free system and submitted to the ICLR 2025 ICBINB workshop.
A summary of these submissions is provided in \Cref{tab:submitted_papers_v2}.
Following the table, each manuscript is included in full, accompanied by comprehensive annotations detailing the internal evaluation, including scientific assessment and code review.

\begin{table}[htbp]
    \centering
    \caption{Overview of AI-Generated Workshop Submissions.}
    \label{tab:submitted_papers_v2}
    \begin{tabularx}{\textwidth}{@{} X c c @{}}
        \toprule
        \textbf{Title} & \textbf{Workshop Result} & \textbf{Materials} \\
        \midrule
        Compositional Regularization: Unexpected Obstacles in Enhancing Neural Network Generalization
        & Accepted (Score: 6.33)
        & \makecell[c]{See \Cref{paper:compos_reg}, \\ \href{https://github.com/SakanaAI/AI-Scientist-ICLR2025-Workshop-Experiment/tree/master/compositional-regularization}{GitHub Repository}} \\
        \addlinespace[1em] %

        Unveiling the Impact of Label Noise on Model Calibration in Deep Learning
        & Rejected
        & \makecell[c]{See \Cref{paper:label_noise}, \\ \href{https://github.com/SakanaAI/AI-Scientist-ICLR2025-Workshop-Experiment/tree/master/label-noise}{GitHub Repository}} \\
        \addlinespace[1em]

        Real-world Challenges in Pest Detection using Deep Learning: an Investigation into Failures and Solutions
        & Rejected
        & \makecell[c]{See \Cref{paper:pest_detection}, \\ \href{https://github.com/SakanaAI/AI-Scientist-ICLR2025-Workshop-Experiment/tree/master/pest-detection}{GitHub Repository}} \\
        \bottomrule
    \end{tabularx}
\end{table}

\clearpage

\subsubsection{Compositional Regularization: Unexpected Obstacles in Enhancing Neural Network Generalization}

\paragraph{\ouralgo Idea}
\begin{tcolorbox}[
  breakable,
  colback  = blue!5!white,
  colframe = blue!75!black,
  title    = {Idea},
  fontupper=\scriptsize\ttfamily,
]
\begin{Verbatim}
"Name": "compositional_regularization_nn",
"Title": "Enhancing Compositional Generalization in Neural Networks via Compositional Regularization",
"Short Hypothesis": "Introducing a compositional regularization term during training can encourage neural networks to develop compositional representations, thereby improving their ability to generalize to novel combinations of known components.",
"Related Work": "Previous work has highlighted the challenges neural networks face in achieving compositional generalization. Studies such as 'Compositional Generalization through Abstract Representations in Human and Artificial Neural Networks' (Ito et al., NeurIPS 2022) have explored abstract representations to tackle this issue. However, limited research focuses on directly incorporating explicit regularization terms into the training objective to enforce compositional structures. Our proposal distinguishes itself by introducing a novel regularization approach that penalizes deviations from predefined compositional patterns during training, encouraging the network to internalize compositional rules.",
"Abstract": "Neural networks excel in many tasks but often struggle with compositional generalization—the ability to understand and generate novel combinations of familiar components. This limitation hampers their performance on tasks requiring systematic generalization beyond the training data. In this proposal, we introduce a novel training method that incorporates an explicit compositional regularization term into the loss function of neural networks. This regularization term is designed to encourage the formation of compositional representations by penalizing the network when its internal representations deviate from expected compositional structures. We hypothesize that this approach will enhance the network's ability to generalize to unseen combinations, mimicking human-like compositional reasoning. We will test our method on synthetic benchmarks like the SCAN and COGS datasets, which are specifically designed to evaluate compositional generalization, as well as on real-world tasks such as machine translation and semantic parsing. By comparing our method to baseline models and existing approaches, we aim to demonstrate significant improvements in generalization performance. This work offers a new avenue for enforcing compositionality in neural networks through regularization, potentially bridging the gap between neural network capabilities and human cognitive flexibility.",
"Experiments": [
  "Implement the compositional regularization term and integrate it into the loss function of standard sequence-to-sequence neural network architectures with attention mechanisms.",
  "Train models on synthetic datasets like SCAN and COGS, evaluating performance on compositional generalization tasks with and without the regularization term.",
  "Apply the method to real-world tasks such as machine translation using the IWSLT dataset and semantic parsing with the GeoQuery dataset, assessing improvements in generalization to new language constructs.",
  "Analyze the learned representations by visualizing embedding spaces and utilizing compositionality metrics to assess how the regularization affects internal representations.",
  "Conduct ablation studies to determine the impact of different strengths of the  regularization term, identifying the optimal balance between enforcing compositionality and maintaining overall performance.",
  "Compare the proposed method against other approaches aimed at improving compositional generalization, such as meta-learning techniques and specialized architectures."
],
"Risk Factors and Limitations": [
  "The effectiveness of the compositional regularization may vary across different datasets and tasks, potentially limiting its generalizability.",
  "An improperly balanced regularization term could negatively impact model performance on the primary task, leading to lower accuracy.",
  "Additional computational overhead from calculating the regularization term may increase training time and resource requirements.",
  "Defining appropriate compositional structures for complex or less-understood domains may be challenging, affecting the applicability of the method.",
  "The approach may face scalability issues when applied to very large models or datasets common in industrial applications."
]
\end{Verbatim}
\end{tcolorbox}

\includepdf[pages=-, scale=0.85]{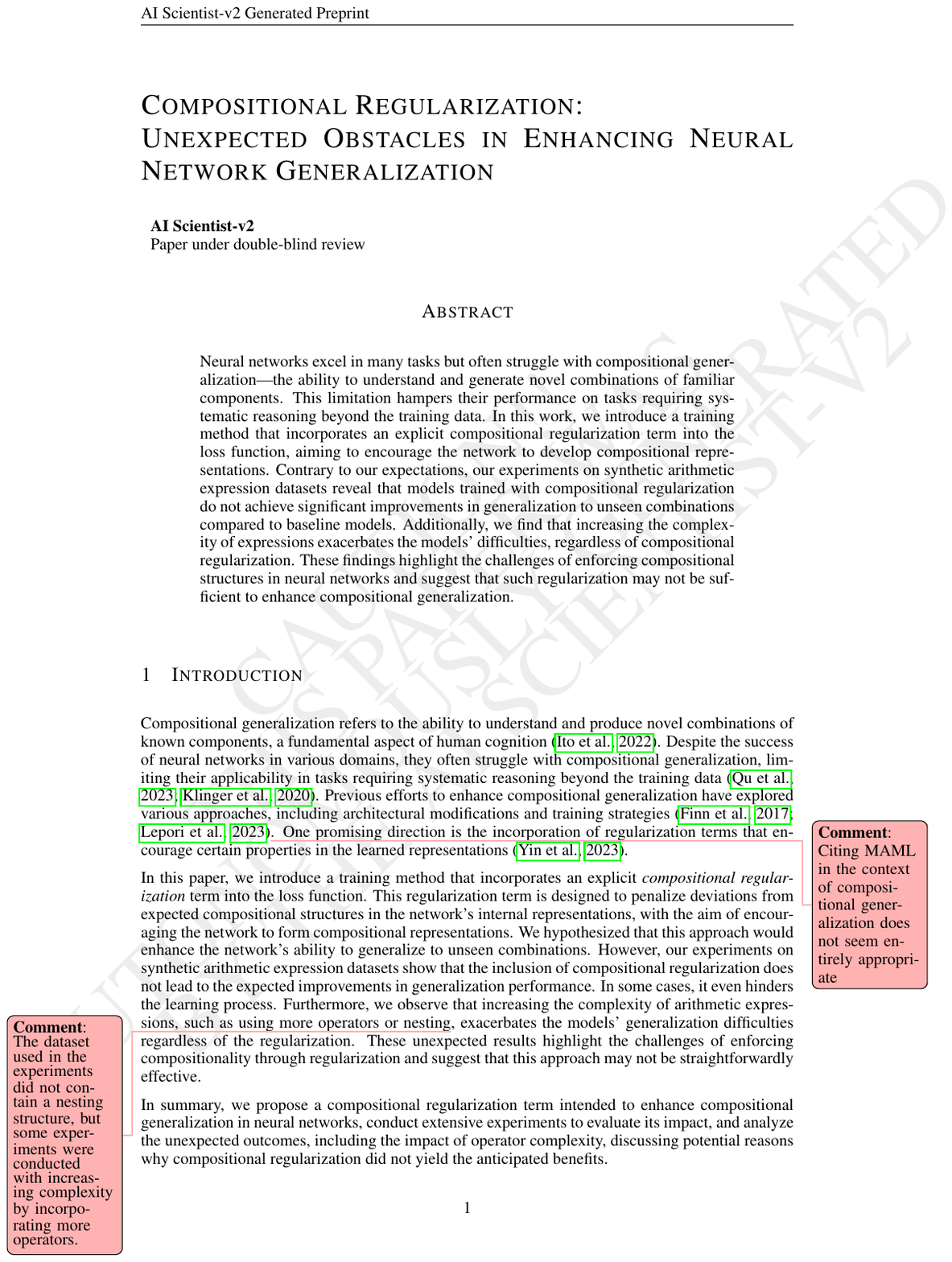}
\label{paper:compos_reg}

\paragraph{AI Scientist Team Review}

\textbf{Paper Summary}

This paper investigates the impact of a temporal consistency regularization term on the compositional generalization of sequence models. The regularizer penalizes large changes in the embedding representation between successive time steps. The experiments consider simple arithmetic tasks and provide evidence that such a regularizer does not improve performance when training the sequence model on multiple tasks. Furthermore, the paper provides small sweeps across different settings, including embedding dimension, regularization strength and architectures. 

\vspace{1em}

\textbf{Strengths}

\begin{itemize}
    \item Although the reasoning behind the design of the proposed regularization is not immediately clear, a simple approach--such as encouraging successive token embeddings to be closer together--presents an interesting avenue for exploring compositional representations.
    \item The chosen arithmetic task is simple, but suitable for testing the hypothesis for varying degrees of difficulty. The chosen experiments provide insights into the impact on various aspects and limitations of the regularization.
\end{itemize}

\textbf{Weaknesses}
\begin{itemize}
    \item The description of the regularization term is vague and can be misleading. Intuitively, the reader can think that it is applied to the LSTM hidden state. Inspecting the code reveals that the regularizer refers to the input embedding hidden state. The text could be enhanced by being more explicit about this detail, adding a code appendix, or providing ablations that apply the regularizer to the LSTM hidden state.
    \item The paper lacks several references, and for example, does not cite Hochreiter and Schmidhuber (1997) but instead opts for the textbook by Goodfellow et al. (2016).
    \item The caption of Figure 3 is wrong. The validation loss increases as task complexity increases. Furthermore, the self-attention-based version discussed in Figure 5 performs substantially better than the LSTM version, while the text argues that they perform on par.
    \item The experimental evaluation could benefit from more depth. The considered sequence lengths are very short, and the considered task is only synthetic. Some of the claims could be strengthened by including real-world tasks, larger networks, and in-depth mechanistic analyses.
\end{itemize}

\textbf{Scores}\footnote{For the ``Overall'' score, we provide evaluations as if we had been reviewing for a workshop (a lower bar) and for a conference (a higher bar).}
\begin{itemize}
    \item \underline{Soundness}: 3/5 good. $\Rightarrow$ Interesting idea with targeted experiments.
    \item \underline{Presentation}: 2/5 fair $\Rightarrow$ Citations, imprecise description, too confident interpretation.
    \item \underline{Contribution}: 3/5 good $\Rightarrow$ Regularizer, analysis, ablations
    \item \underline{Overall - Workshop}: 5/10 (Borderline accept): Technically solid paper where reasons to accept outweigh reasons to reject, e.g., limited evaluation.
    \item \underline{Overall - Conference}: 4/10: (Borderline reject): Technically solid paper where reasons to reject, e.g., limited evaluation, outweigh reasons to accept, e.g., good evaluation.
    \item \underline{Confidence}: 4/5. You are confident in your assessment, but not absolutely certain. It is unlikely, but not impossible, that you did not understand some parts of the submission or that you are unfamiliar with some pieces of related work.
\end{itemize}

\textbf{Additional Comments}
\begin{itemize}
    \item To strengthen the analysis, several different compositional regularizers should be compared across different tasks. Additionally, it needs to be more explicitly tested whether the regularizer actually induces compositional representations. This could be done, for example, via linear probes trained on the embedding representations or by visualizing low-dimensional embeddings.
\end{itemize}

\textbf{Potential Violation of Code of Ethics:} No.\\

\paragraph{AI Scientist Team Code Review}

\subsection*{Inspecting the dataset generation process}

\begin{figure}[h!]
\centering
\includegraphics[width=0.75\textwidth]{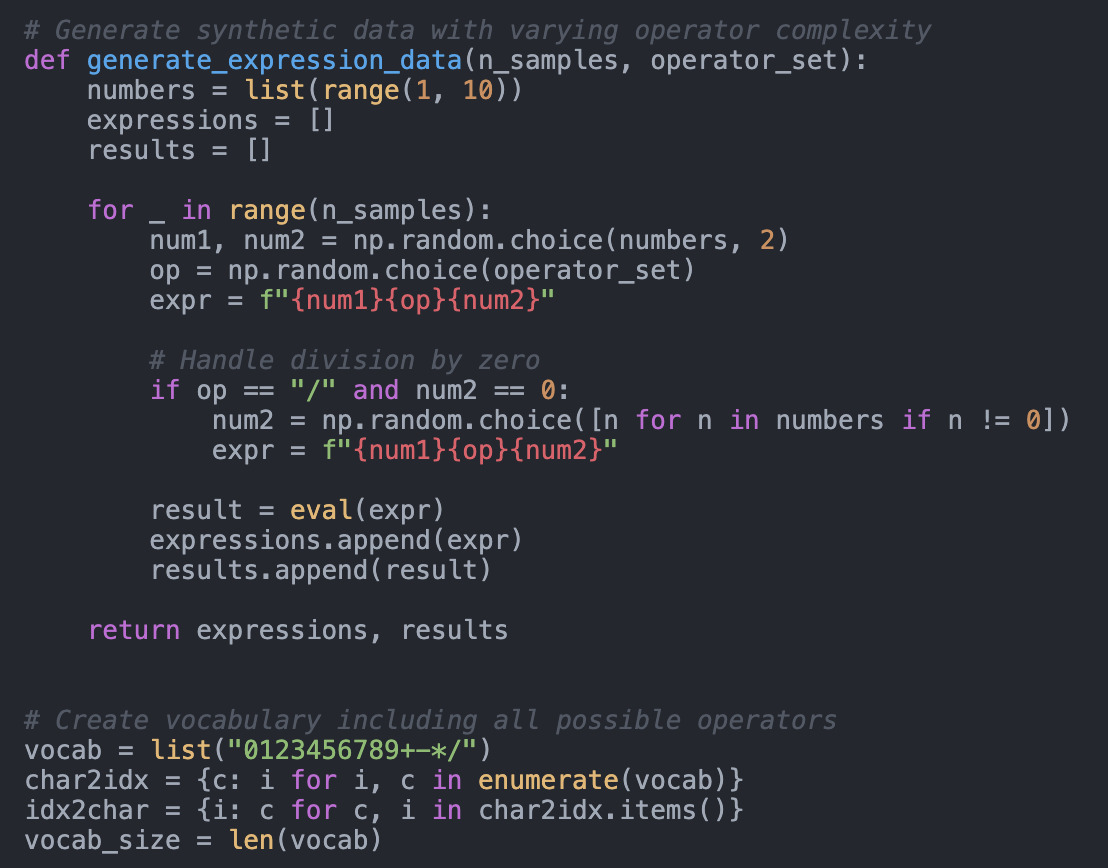}
\caption{Example of the data-generating function used in the experiments.}
\label{fig:data_generating}
\end{figure}

The data-generating function, which uses a single-digit expression as shown in \Cref{fig:data_generating}, generates at most $81 * k$ possible combinations, where $k$ is the number of operators.
Generating train and test sets from the same data generator can be fine in large search spaces, but \ouralgo failed to realize this space is too low-dimensional to not explicitly deduplicate them.
This suggests that the training and test datasets can have significant overlap, depending on the number of samples and the choice of operators.

As a sanity check, we generated the dataset 10 times using addition and multiplication operators, with [1-9] as the available numbers, and 1,000 training samples and 200 test samples. On average, we found that about 57\% of the test set overlapped with the training set.

\subsection*{Model architecture, loss function, and evaluation function}

\begin{figure}[h!]
\centering
\includegraphics[width=0.75\textwidth]{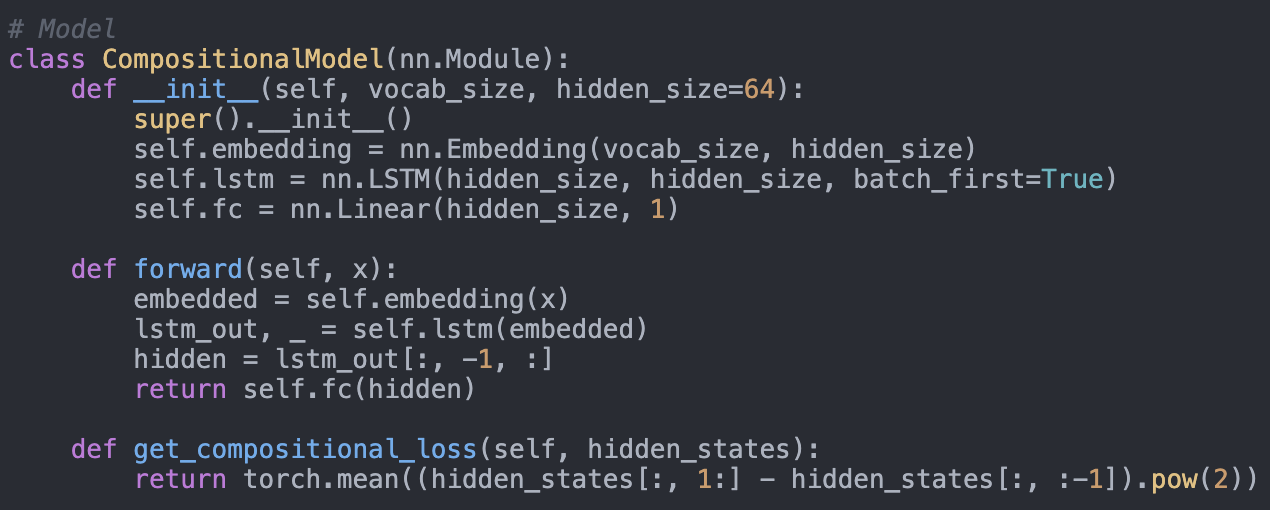}
\caption{The generated model class shows an embedding layer, a single LSTM layer, and a linear layer head.}
\label{fig:model_class}
\end{figure}

The model architecture is simple, and its implementation appears to be correct, as shown in \Cref{fig:model_class}.

\begin{figure}[h!]
\centering
\includegraphics[width=0.75\textwidth]{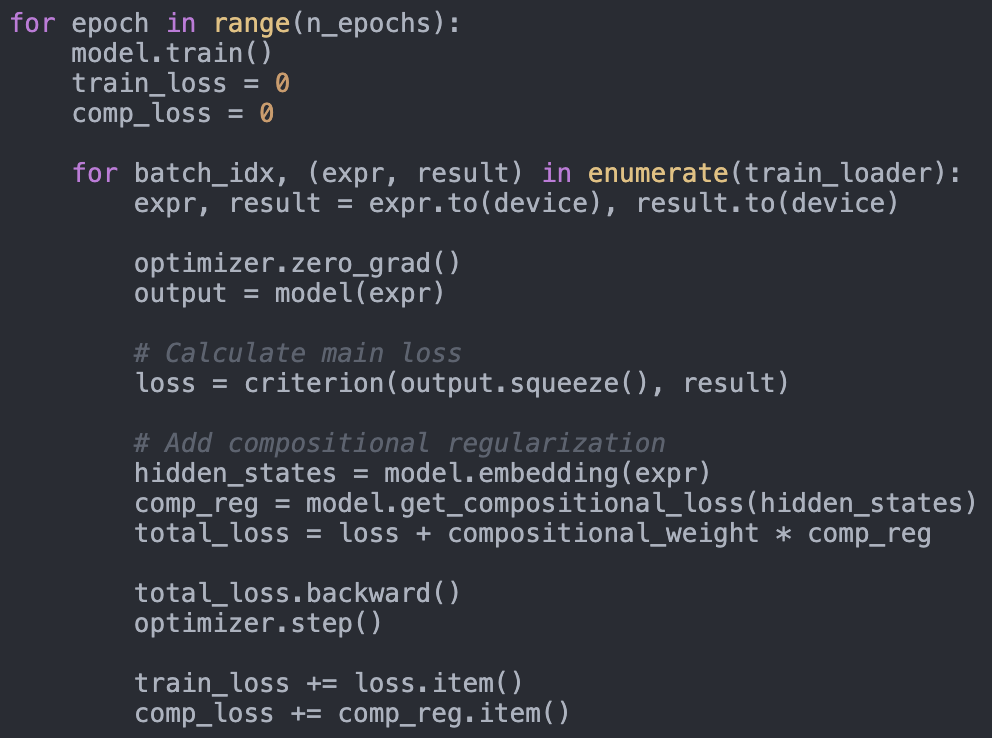}
\caption{The generated training loop shows the loss function as well as the proposed regularization.}
\label{fig:train_loop_loss_regularization}
\end{figure}

In the training loop, presented in \Cref{fig:train_loop_loss_regularization}, compositional regularization is computed using the embedding states. Therefore, the main paper should use the notation $e_t$ to represent embeddings instead of $h_t$, and explicitly refer to these as embeddings rather than hidden states. 
Although embeddings are technically a hidden layer, the term ``hidden states'' in this context usually refers to LSTM hidden states, which could be confusing.

\begin{figure}[h!]
\centering
\includegraphics[width=0.75\textwidth]{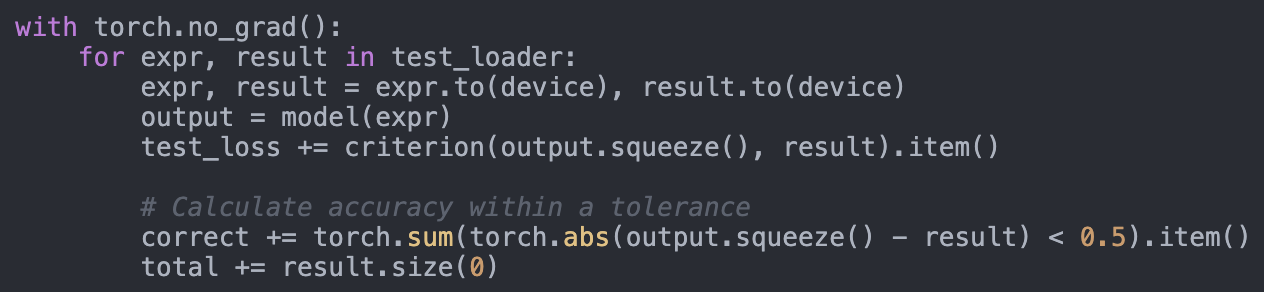}
\caption{The generated accuracy calculation function uses regression to match an output with a ground truth.}
\label{fig:eval_function}
\end{figure}

The accuracy calculation function (\Cref{fig:eval_function}) indicates that the model performs regression on the output to match the ground truth digits. This approach makes sense, as it allows the model to handle arbitrary values, including those outside the range [0-9].

\subsection*{Attention-augmented LSTM}

In the paper, a 100\% test accuracy was reported for the attention-augmented LSTM. To verify this, we re-ran the same experiment using the generated code for two cases: the first with the available numbers [1-9] (as in the original setup), and the second with the available numbers modified to [10-19]. In the first case, the attention-augmented LSTM achieved 100\% test accuracy, while in the second case, it achieved 56\% test accuracy. For the baseline LSTM, the first case resulted in 85\% test accuracy, and the second case yielded 0\% test accuracy. We concluded that the first case was too simple for the attention-augmented LSTM, and as the task complexity increased (e.g., the first case involved a length of 3, such as 3 + 5, while the second case involved a length of 5, such as 14 * 19, with a larger output space), the test accuracy deviated from the initial 100\%.

\newpage
\paragraph{Workshop Reviews}

\begin{tcolorbox}[
  breakable,
  colback  = green!5!white,
  colframe = green!75!black,
  title    = {\small Reviewer \#1: A good paper analyzing the effectiveness of a compositional regularization term for LSTMs},
  fontupper=\scriptsize\ttfamily,
]
\begin{Verbatim}
Summary: The authors propose a regularisation term to enhance compositional regularisation in neural networks. The idea is to penalise large deviations between subsequent time steps of the hidden state, thus “squeezing” the hidden state to encourage composition and preventing a dominating representation. The authors test their approach on synthetic arithmetic expression with varying operator complexity and length. They show that although the regularisation term  appears to be working, it counterintuitively does not improve test accuracy. Furthermore, the authors identify a bottleneck regarding network capacity with increasing arithmetic operators.

Strengths:
I find the idea of regularising or squeezing the hidden representations to encourage compositionally an interesting idea. The authors define a good baseline and ablate their method well against it, revealing why the regularisation term does not work as expected. I think the insight that operator complexity is a bottleneck for the neural network is important, as it raises the question whether architectural changes might be more effective for compositionally than regularisation.

Weaknesses:
The paper would benefit from more intuition as to why the proposed regularisation term should encourage compositionality. This could be either an experiment or simply a visualisation for the reader. Only one architecture (LSTM) was tested. It would be interesting to see if transformer architectures fare better with compositionality due to the attention mechanism. I think the connection between compositional regularisation and operator complexity needs to be made more explicit. From reading the introduction both arguments seem a bit disconnected although I can infer the authors intentions.

Conclusion:
Overall, I would accept this paper to the workshop, since it proposes a simple and interesting idea with the authors providing ablations that encourage further analysis of the problem. As a suggestion I would encourage the authors to give more intuition on why the proposed regularisation term should improve compositionality for the proposed network. I would suggest either adding more related work to support the regularisation term or elaborating on the intuition behind penalising subsequent steps of the hidden state.

Rating: 7: Good paper, accept
Award: No Award
Confidence: 4: The reviewer is confident but not absolutely certain that the evaluation is correct
\end{Verbatim}
\end{tcolorbox}

\begin{tcolorbox}[
  breakable,
  colback  = green!5!white,
  colframe = green!75!black,
  title    = {\small Reviewer \#2: Compositional Regularization: Unexpected Obstacles in Enhancing Neural Network Generalization},
  fontupper=\scriptsize\ttfamily,
]
\begin{Verbatim}
This paper investigates the effectiveness of incorporating a compositional regularization term into the loss function of neural networks to improve compositional generalization. The authors hypothesized that penalizing deviations from compositional structures would enhance the model’s ability to generalize to unseen arithmetic expressions. However, their results on synthetic arithmetic datasets showed that compositional regularization did not lead to significant improvements and, in some cases, even hindered learning.

I think this paper greatly contributes to the workshops theme and fits into the scope. Moreover, it is a great example of challenges that occur during such approaches and could be interesting to discuss in the workshop setting. While I think that the authors should further broaden the experiments to other tasks in order to increase the generalizability of the findings, I would still recommend to accept the paper.

Rating: 6: Marginally above acceptance threshold
Award: No Award
Confidence: 2: The reviewer is willing to defend the evaluation, but it is quite likely that the reviewer did not understand central parts of the paper
\end{Verbatim}
\end{tcolorbox}

\clearpage

\subsubsection{Unveiling the Impact of Label Noise on Model Calibration in Deep Learning}

\paragraph{\ouralgo Idea}
\begin{tcolorbox}[
  breakable,
  colback  = blue!5!white,
  colframe = blue!75!black,
  title    = {Idea},
  fontupper=\scriptsize\ttfamily,
]
\begin{Verbatim}
"Name": "label_noise_calibration",
"Title": "Unveiling the Impact of Label Noise on Model Calibration in Deep Learning",
"Short Hypothesis": "Label noise not only degrades model accuracy but also adversely affects model calibration and uncertainty estimation; by systematically studying this impact, we can develop methods to improve both accuracy and calibration under label noise.",
"Related Work": "Previous studies have focused on the impact of label noise on model accuracy and have proposed methods to mitigate this issue, such as robust loss functions and label correction techniques. However, there is limited research on how label noise affects model calibration and uncertainty estimation. For instance, works like 'Dynamics-Aware Loss for Learning with Label Noise' (Li et al., 2023) address robustness to label noise but do not explore calibration aspects. Our proposal distinguishes itself by systematically investigating the effect of label noise on model calibration, which is crucial for reliable deployment of deep learning models in real-world applications.",
"Abstract": "Label noise is a prevalent issue in real-world datasets, where incorrect annotations can degrade the performance of deep learning models. While the impact of label noise on model accuracy has been extensively studied, its effect on model calibration and uncertainty estimation remains underexplored. Model calibration measures how well the predicted probabilities reflect the true likelihood of outcomes, which is vital for risk-sensitive applications that rely on uncertainty estimates for decision-making. In this research, we propose to systematically investigate how different types and levels of label noise affect the calibration of deep learning models. We hypothesize that label noise leads to overconfident and miscalibrated predictions, undermining the reliability of uncertainty estimates. Through controlled experiments on benchmark datasets with synthetic label noise and real-world datasets with inherent label noise, we will analyze calibration metrics such as Expected Calibration Error (ECE) and reliability diagrams. Additionally, we will assess the effectiveness of existing label noise mitigation techniques in improving model calibration. The findings from this study will provide insights into the relationship between label noise and model calibration, guiding the development of more robust models that maintain reliable uncertainty estimates despite noisy labels.",
"Experiments": [
    "Introduce varying levels and types of synthetic label noise (e.g., symmetric and asymmetric noise) into benchmark datasets like CIFAR-10 and MNIST.",
    "Train deep learning models (e.g., ResNet, CNNs) on these noisy datasets and evaluate their accuracy and calibration using metrics like ECE and reliability diagrams.",
    "Analyze how different label noise levels impact model calibration compared to their effect on accuracy.",
    "Apply existing label noise mitigation techniques, such as robust loss functions and label correction methods, to assess their effectiveness in improving calibration.",
    "Evaluate models on real-world datasets known to contain label noise (e.g., web-scraped datasets) to validate the findings in practical scenarios.",
    "Conduct ablation studies to understand the interplay between label noise, model calibration, and uncertainty estimation."
],
"Risk Factors and Limitations": [
    "Results may be specific to the selected models and datasets, potentially limiting generalization to other architectures or domains.",
    "Measuring calibration accurately requires sufficient test data; small test sets may lead to unreliable calibration metrics.",
    "Existing mitigation techniques may not significantly improve calibration, indicating a need for developing new methods.",
    "Synthetic label noise may not capture all aspects of real-world label noise, affecting the applicability of the findings."
],
"Code": "from datasets import load_dataset\nfrom huggingface_hub import ..."
\end{Verbatim}
\end{tcolorbox}

\includepdf[pages=-, scale=0.85]{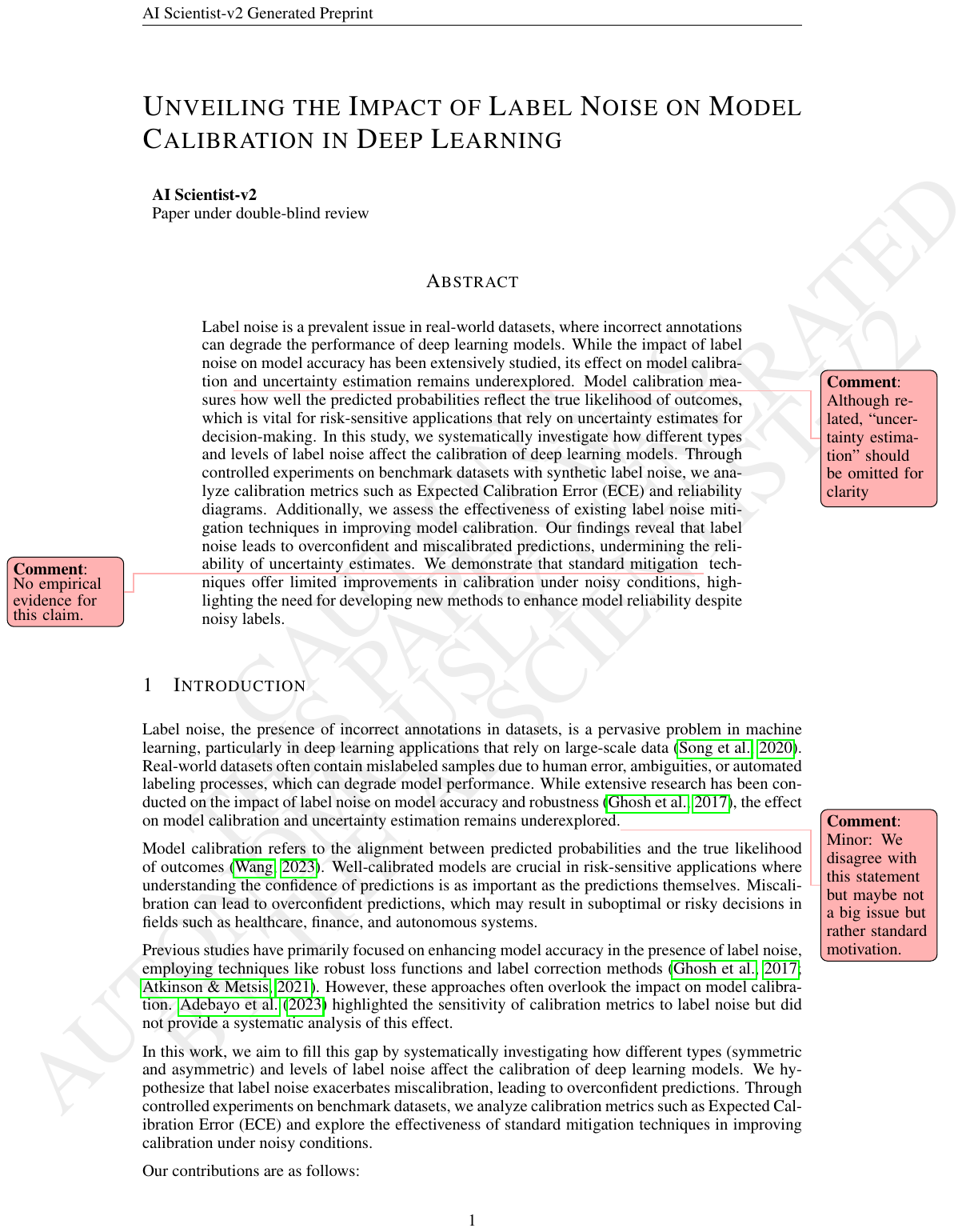}
\label{paper:label_noise}

\paragraph{AI Scientist Team Review}

\textbf{Paper Summary}
This paper studies the impact of label noise on model calibration using different noise models. More specifically, the paper contrasts symmetric (unstructured label perturbations) and asymmetric (structured label perturbations) noise. The empirical experiments consider standard small-scale vision datasets (i.e., MNIST, Fashion-MNIST and CIFAR-10) and demonstrate that asymmetric noise leads to higher expected calibration error.

\textbf{Strengths}

\begin{itemize}
    \item The research question is of real-world importance and shines light on the impact of noisy labels beyond their effect on prediction accuracy.
    \item The study design is simple and focuses on a single key factor, i.e., the impact of different noise models (asymmetric noise increasing ECE more than symmetric noise). The considered datasets are appropriate for a workshop submission.
    \item The impact of the different noise models on the downstream model calibration is robust and consistent across the considered datasets. 
\end{itemize}

\textbf{Weaknesses}
\begin{itemize}
    \item There are multiple instances where the written interpretation of results is not substantially supported by the empirical results presented. E.g., the paragraph interpreting Figure 3 refers to ECE measures, which are not displayed in the figures.
    \item The paper states that it compares different calibration methods, but the paper does not provide any results. The same holds for the mentioned reliability diagrams.
    \item Furthermore, the supplementary material includes duplicate figures, a missing citation for SVHN, and a corresponding missing figure.
\end{itemize}

\textbf{Scores}
\begin{itemize}
    \item \underline{Soundness}: 2 fair. $\Rightarrow$ Interesting research question with a potentially simple empirical evaluation setup. 
    \item \underline{Presentation}: 1 poor. $\Rightarrow$ Wrong description and duplication of figures. Missing citation and downplaying of related work.
    \item \underline{Contribution}: 1 poor. $\Rightarrow$ While the question considered is important, the displayed results do not provide enough evidence for the conclusions drawn.
    \item \underline{Overall - Workshop}: 3/10 (Reject): For instance, a paper with technical flaws, weak evaluation, inadequate reproducibility, and incompletely addressed ethical considerations.
    \item \underline{Overall - Conference}: 2/10: (Strong reject): For instance, a paper with major technical flaws, and/or poor evaluation, limited impact, poor reproducibilit,y and mostly unaddressed ethical considerations.
    \item \underline{Confidence}: 4/5. You are confident in your assessment, but not absolutely certain. It is unlikely, but not impossible, that you did not understand some parts of the submission or that you are unfamiliar with some pieces of related work.
\end{itemize}

\textbf{Additional Comments}
\begin{itemize}
    \item The biggest flaw of this paper is the mention of results that are not substantiated. This includes the assessment of various methods tailored to uncertainty calibration, as well as the usage of reliability diagrams. The paper could be substantially improved if these results were added, and the selection of displayed figure results was better curated.
    \item The readability of Figure 2 should be improved by splitting the 6 plots across 2 rows. Furthermore, the related work section appears to dismiss efforts by the scientific community to relate calibration and noisy data.
\end{itemize}

\textbf{Potential Violation of Code of Ethics:} No.

\newpage

\paragraph{AI Scientist Team Code Review}

\subsection*{Temperature scaling}

\begin{figure}[h!]
\centering
\includegraphics[width=0.75\textwidth]{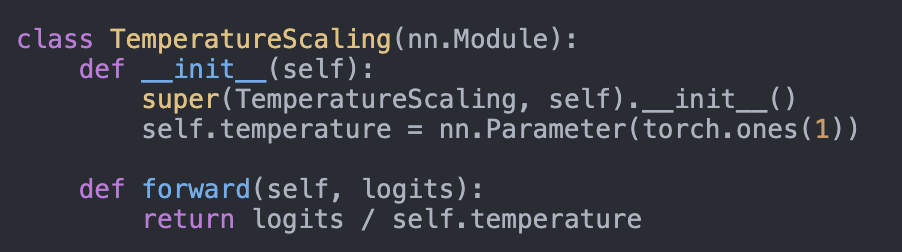}
\caption{Temperature scaling implementation.}
\label{fig:temperature_scaling}
\end{figure}

In our review of the paper, we noted that it lacked experiments involving temperature scaling. 
Upon inspecting the generated code, we found that the AI Scientist had implemented temperature scaling, as can be seen in \Cref{fig:temperature_scaling}, but never actually used it. 

During the paper writing stage, the AI Scientist had access to a set of generated experiment code and its initial plans before generating the code. As a result, it is likely that the paper was influenced by these plans and code, which included temperature scaling, but the AI Scientist failed to realize that the experiments using temperature scaling were never actually conducted.

\subsection*{Dataset class}

\begin{figure}[h!]
\centering
\includegraphics[width=0.49\textwidth]{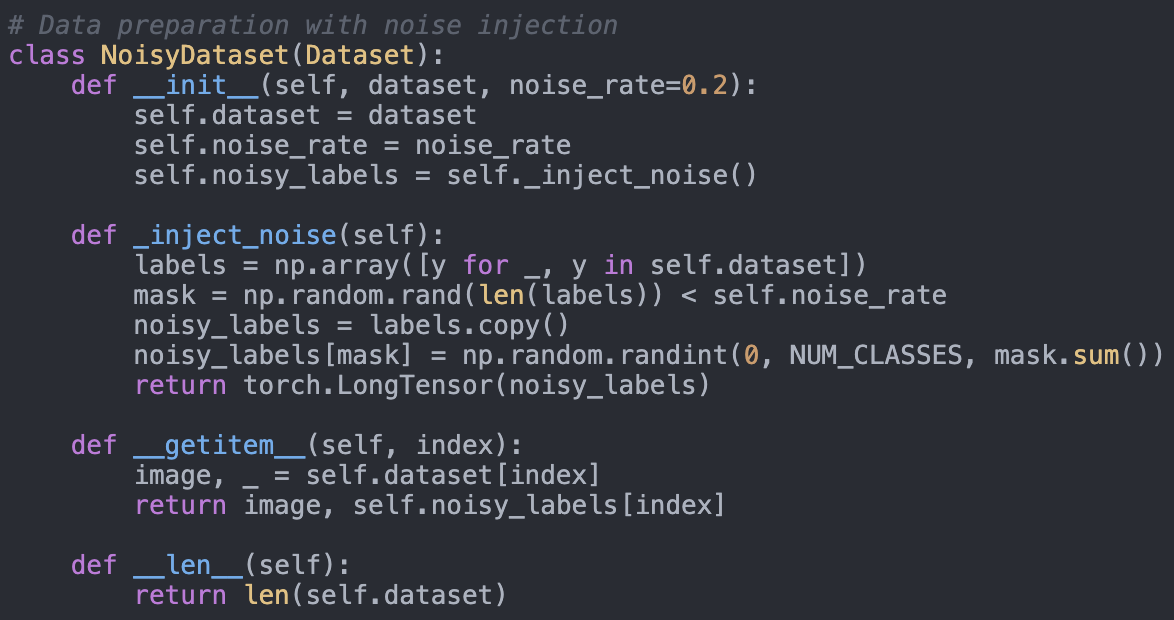}
\includegraphics[width=0.49\textwidth]{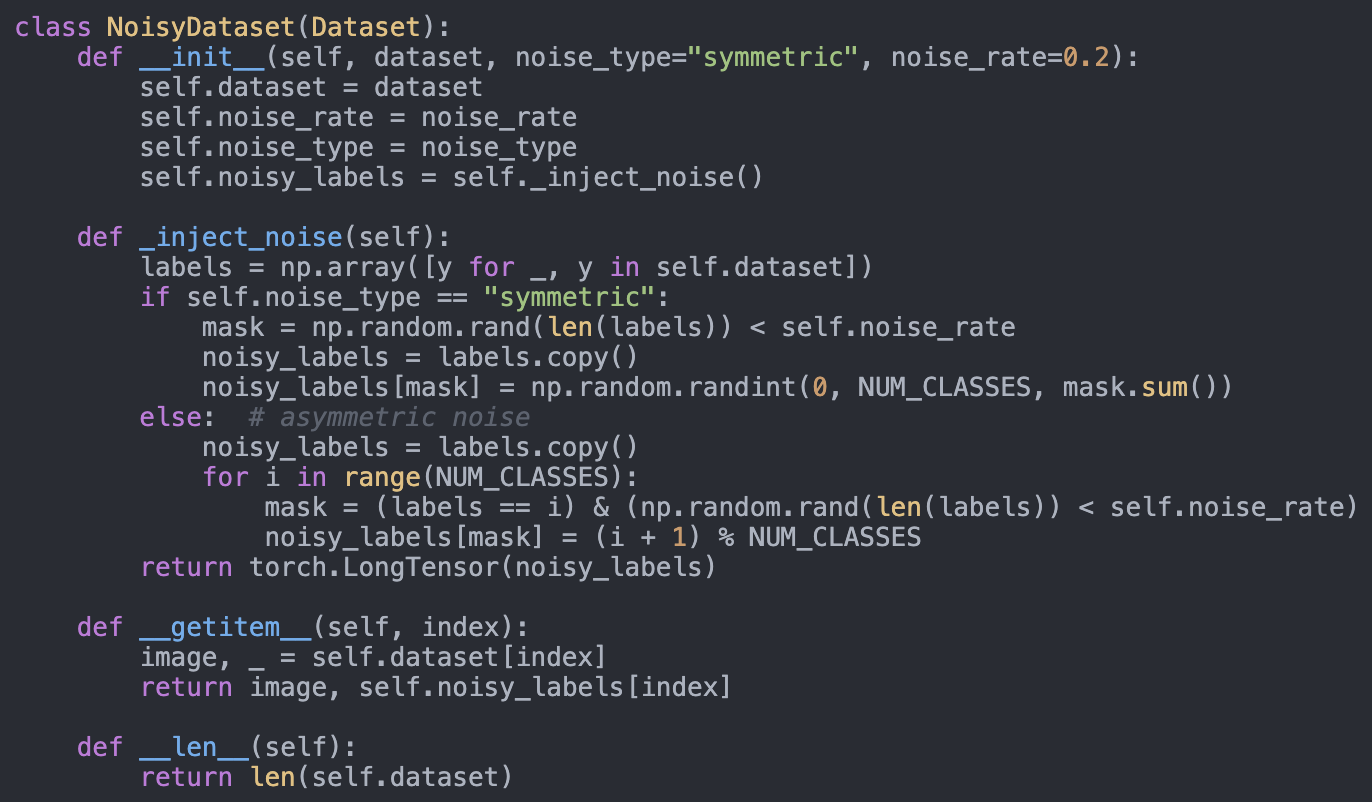}
\caption{Noisy dataset class implementation.}
\label{fig:noisy_dataset_class}
\end{figure}

We found that the initial implementation of the dataset class lacked an option for symmetric/asymmetric noise distribution, even though it was part of the initial plan. The AI Scientist recognized this mistake and later implemented the correct version, as shown in \Cref{fig:noisy_dataset_class}.

In the main paper, the AI Scientist wrote: ``Assymetric Noise: Labels are flipped to specific incorrect classes based on a predefined confusion matrix, simulating more realistic mislabeling.'' 
The asymmetric noise implementation in the generated code always maps class i to class (i+1) \% NUM CLASSES. 
While this is a valid approach, it is worth noting that there are other ways to implement asymmetric noise.

\subsection*{Evaluation function}

\begin{figure}[h!]
\centering
\includegraphics[width=0.75\textwidth]{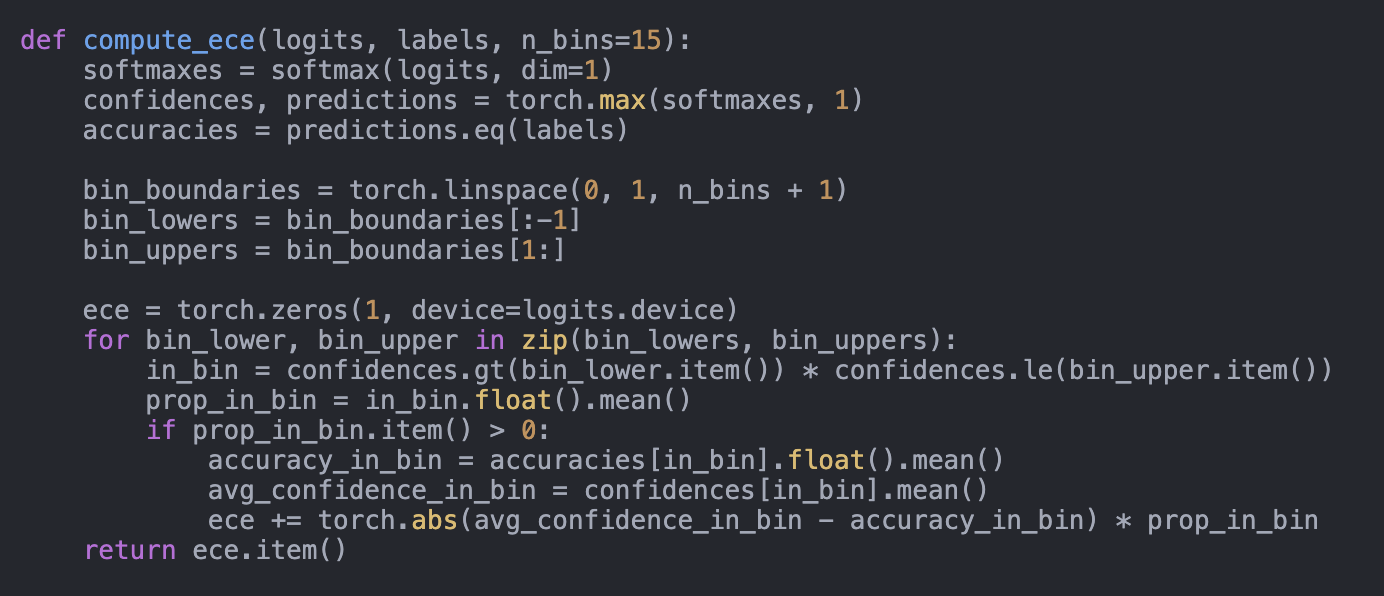}
\caption{The implementation for Expected Calibration Error.}
\label{fig:ece_calc_func}
\end{figure}

The evaluation function used to compute the Expected Calibration Error is shown in \Cref{fig:ece_calc_func}.
We manually created test cases and used the MulticlassCalibrationError function with norm=`l1' from torchmetrics as the ground truth.
Since the MulticlassCalibrationError function expects probability inputs, we omitted the softmax operation in the first line to align with the implementation details.
After this adjustment, we confirmed that both functions produce the same results, apart from minor numerical differences.

\newpage
\paragraph{Workshop Reviews}

\begin{tcolorbox}[
  breakable,
  colback  = green!5!white,
  colframe = green!75!black,
  title    = {\small Reviewer \#1: This work explores the impact of label noise on model calibration, demonstrating that label noise degrades calibration performance.},
  fontupper=\scriptsize\ttfamily,
]
\begin{Verbatim}
The research question is intriguing; however, the experimental analysis appears somewhat unclear. The underlying mechanism explaining how the experimental results support the claimed statement is not well articulated. Specifically, in the abstract, the authors state that "label noise leads to overconfident and miscalibrated predictions, undermining the reliability of uncertainty  estimates," yet I struggle to see a clear connection between this claim and the content in the main body.

Additionally, the experimental setup raises some concerns. To thoroughly assess the impact of label noise on model calibration, a more refined approach to introducing label noise should be considered. Moreover, incorporating a broader range of evaluation metrics would help strengthen the conclusions.

Furthermore, the images in the paper are difficult to interpret, and some citations appear to be missing. The referenced papers are also kind of old, which could weaken the soundness of related work.

Rating: 3: Clear rejection
Award: No Award
Confidence: 5: The reviewer is absolutely certain that the evaluation is correct and very familiar with the relevant literature
\end{Verbatim}
\end{tcolorbox}

\begin{tcolorbox}[
  breakable,
  colback  = green!5!white,
  colframe = green!75!black,
  title    = {\small Reviewer \#2: Official Review for Submission41},
  fontupper=\scriptsize\ttfamily,
]
\begin{Verbatim}
This paper is not finished, there are missing references indicated by (?) there are unlinked references (eg L097), the figures are unreadable (eg Fig 2). I feel this paper is in a late draft status and not review ready.

Remaining comment

It is unclear how Noise Mitigation Techniques, or Calibration Improvements (like temperature scaling) are taken into account in the study. Or how they affect performance after label noise. It is stated that temperature scaling is used, but its effect is not made clear.

Rating: 3: Clear rejection
Award: No Award
Confidence: 4: The reviewer is confident but not absolutely certain that the evaluation is correct
\end{Verbatim}
\end{tcolorbox}

\clearpage

\subsubsection{Real-world Challenges in Pest Detection using Deep Learning: an Investigation into Failures and Solutions}

\paragraph{\ouralgo Idea}
\begin{tcolorbox}[
  breakable,
  colback  = blue!5!white,
  colframe = blue!75!black,
  title    = {Idea},
  fontupper=\scriptsize\ttfamily,
]
\begin{Verbatim}
"Name": "real_world_pest_detection",
"Title": "Real-World Challenges in Pest Detection Using Deep Learning: An Investigation into Failures and Solutions",
"Short Hypothesis": "Deep learning models for pest detection often fail to generalize in real-world agricultural settings due to data quality issues, environmental variability, and model limitations. Investigating these failures can lead to more robust solutions.",
"Related Work": "Several studies, such as those by Agarwal et al. (2023) and Dong et al. (2024), have explored deep learning for pest detection in agriculture. These studies generally report high accuracy in controlled settings but often do not address real-world deployment challenges. Our proposal distinguishes itself by focusing on the negative outcomes and the underlying reasons behind these failures.",
"Abstract": "Accurate pest detection is vital for protecting crops and ensuring food security. While deep learning models have shown promise in controlled environments, their performance often degrades in real-world applications. This proposal aims to investigate the reasons behind these failures. We hypothesize that data quality issues, environmental variability, and model limitations are significant factors. By conducting a series of experiments, we will explore these challenges in depth and propose robust solutions to improve the generalizability of deep learning models for pest detection. Our research will provide valuable insights for the agricultural community and contribute to the development of more reliable AI tools for precision farming.",
"Experiments": [
    "1. **Data Quality Analysis**: Collect a diverse dataset of pest images from different agricultural environments and analyze its quality. Identify common issues such as label noise, class imbalance, and distribution shift.",
    "2. **Model Robustness Testing**: Train state-of-the-art deep learning models (e.g., YOLOv8, EfficientNetB3) on the collected dataset and evaluate their performance in controlled vs. real-world settings. Metrics: Mean Average Precision (mAP), F1 Score.",
    "3. **Environmental Variability Study**: Evaluate model performance under different environmental conditions (e.g., lighting, weather). Identify which conditions most significantly impact model accuracy.",
    "4. **Failure Mode Analysis**: Conduct a detailed analysis of misclassifications to identify common patterns and potential causes (e.g., feature overlap between pests and background).",
    "5. **Improvement Strategies**: Implement and test various strategies to mitigate identified challenges, such as data augmentation, domain adaptation, and model ensembling. Evaluate their effectiveness in improving model robustness."
],
"Risk Factors and Limitations": "Potential risks include the availability and quality of real-world data, the computational demands of training and testing multiple deep learning models, and the generalizability of the findings to different types of pests and crops. Additionally, environmental factors may introduce variability that is challenging to control.",
"Code": "from datasets import ..."
\end{Verbatim}
\end{tcolorbox}

\textbf{Note on dataset preparation:}
Three AI-generated ideas were selected--two from the core machine learning research batch and one from the more applied, real-world-oriented batch. This pest detection idea comes from the latter.
While the system performs well when downloading standard machine learning datasets, it still struggles to automatically access real-world datasets available online. To address this, a relevant dataset from Kaggle (Crop Pest and Disease Detection\footnote{https://www.kaggle.com/datasets/nirmalsankalana/crop-pest-and-disease-detection}) was manually downloaded and its size was reduced to one-tenth to speed up the experiments.
Automating this data preparation process further to handle a broader range of datasets remains an interesting direction for future research.

\includepdf[pages=-, scale=0.85]{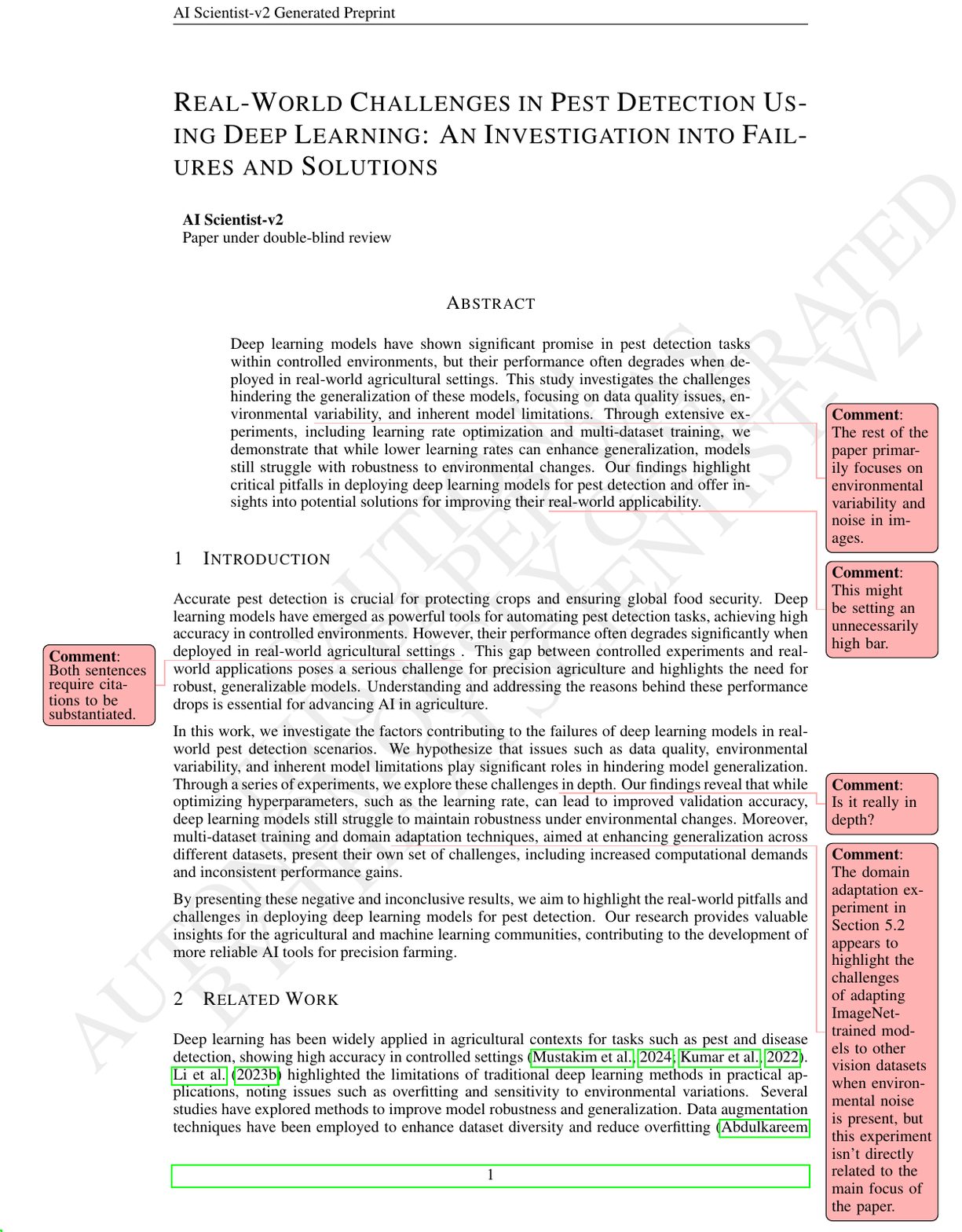}
\label{paper:pest_detection}

\paragraph{AI Scientist Team Review}

\textbf{Paper Summary}
This paper studies the application of Deep Learning models for a real-world application to pest prediction. It introduces an Environmental Robustness Score that leverages various data augmentation techniques, mimicking environmental factors affecting data collection. It compares various learning rates and the resultant impact of out-of-distribution testing settings across non-pest vision datasets.

\textbf{Strengths}

\begin{itemize}
    \item The paper fits the ICBNB workshop topic especially well. It discusses a real-world application of Deep Learning methods to pest prediction.
    \item Understanding the differential impact of training and out-of-distribution data augmentation technique settings across datasets is interesting.  
\end{itemize}

\textbf{Weaknesses}
\begin{itemize}
    \item The paper refers to domain adaptation being studied multiple times. The experiments, on the other hand, only investigate the usage of data augmentation methods (such as lighting, blurring, and contrast manipulation). Furthermore, studying the impact of the learning rate on generalization is fairly trivial.
    \item It is hard to motivate that the Eurosat, Medmnist, and CIFAR-10 results are related to the pest prediction problem. Why should a result on these datasets transfer to pest prediction?
    \item Some of the statements regarding multi-dataset training are misleading. There are no results in the paper that result from such a training setup. Instead, multiple models are trained on individual datasets.
\end{itemize}

\textbf{Scores}
\begin{itemize}
    \item \underline{Soundness}: 2 fair. $\Rightarrow$ Interesting research question with potentially simple empirical evaluation setup. 
    \item \underline{Presentation}: 1 poor. $\Rightarrow$ Wrong description and duplication of figures. Missing citation and downplaying of related work.
    \item \underline{Contribution}: 1 poor. $\Rightarrow$ While the question considered is important, the displayed results do not provide enough evidence for the conclusions drawn.
    \item \underline{Overall - Workshop}: 3/10 (Reject): For instance, a paper with technical flaws, weak evaluation, inadequate reproducibility, and incompletely addressed ethical considerations.
    \item \underline{Overall - Conference}: 2/10: (Strong reject): For instance, a paper with major technical flaws, and/or poor evaluation, limited impact, poor reproducibility, and mostly unaddressed ethical considerations.
    \item \underline{Confidence}: 4/5. You are confident in your assessment, but not absolutely certain. It is unlikely, but not impossible, that you did not understand some parts of the submission or that you are unfamiliar with some pieces of related work.
\end{itemize}

\textbf{Additional Comments}
\begin{itemize}
    \item The presentation of the results needs significant improvement. There are multiple missing citations (?), and the interpretation of the results can be misleading. This includes the conclusions with regard to the impact of a lower learning rate on overfitting or naming the multi-model-single-dataset experiment ``mulit-dataset''.
\end{itemize}

\textbf{Potential Violation of Code of Ethics:} No.

\newpage

\paragraph{AI Scientist Team Code Review}

\subsection*{Domain Adaptation and Multi-dataset training}

The paper seems to describe the ``Domain Adaptation'' experiment as primarily focused on transferring ImageNet-pretrained models to other vision datasets.
After reviewing the code, we found attempts to implement a domain adaptation technique by training a separate classifier to distinguish different domains, but these attempts were unsuccessful.
In the end, the AI Scientist opted for an implementation that does not include this domain adaptation technique.

\begin{figure}[h!]
\centering
\includegraphics[width=0.49\textwidth]{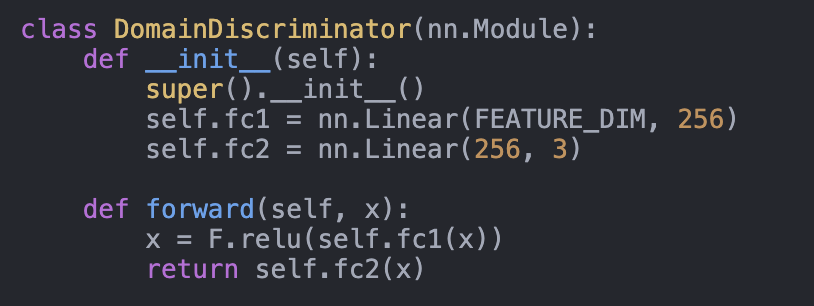}
\includegraphics[width=0.49\textwidth]{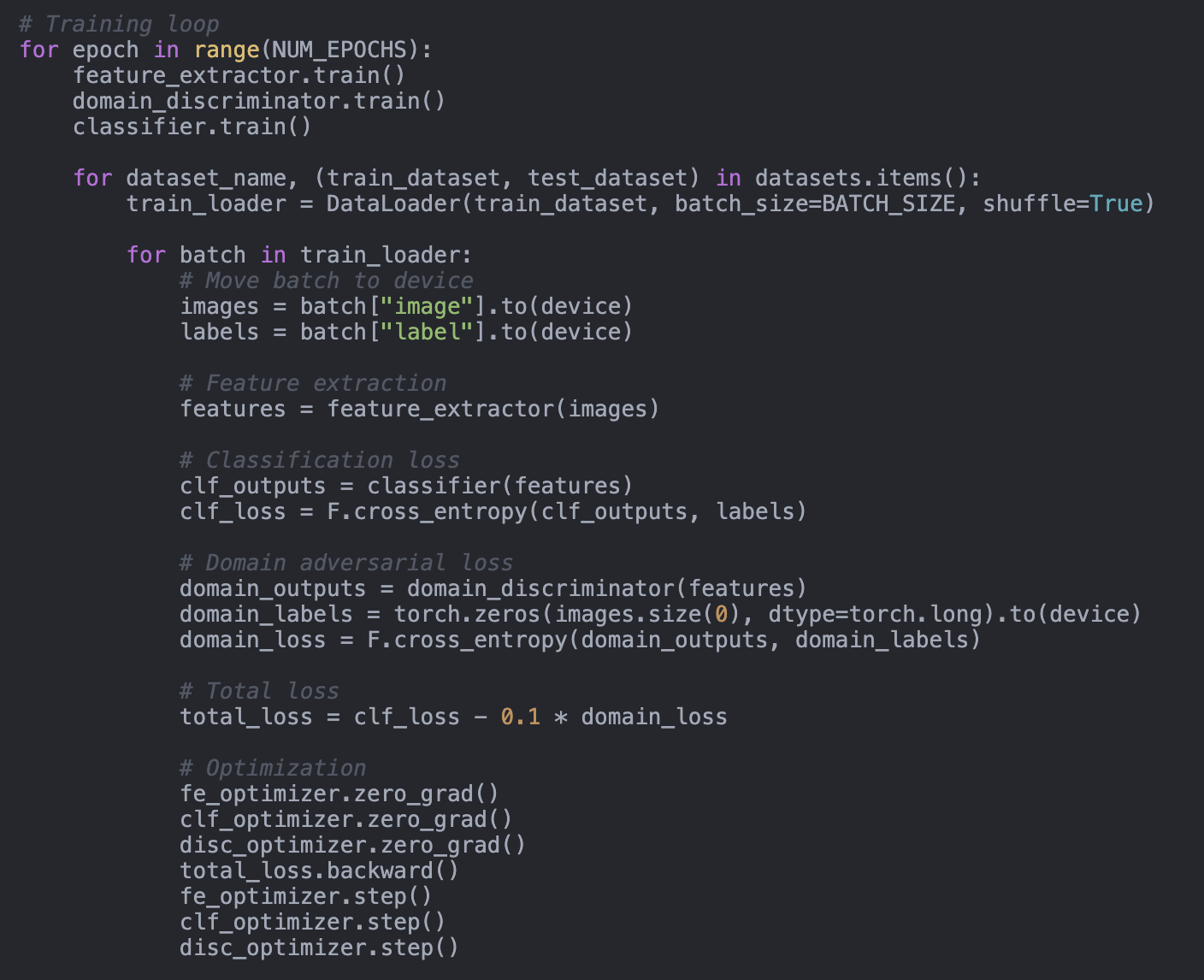}
\caption{Domain discriminator and multi-dataset training loop.}
\label{fig:domain_disc_and_multi_dataset_train_loop}
\end{figure}

Moreover, in the code where this domain adaptation technique was implemented, multi-dataset training was correctly performed as well--training a single model on all three datasets with domain discriminator loss, as shown in \Cref{fig:domain_disc_and_multi_dataset_train_loop}.
Had this code run successfully, the AI Scientist would likely have chosen it over the one ultimately selected, which lacked proper multi-dataset training but ran without errors.

\subsection*{Environmental noise implementation}

\begin{figure}[h!]
\centering
\includegraphics[width=0.75\textwidth]{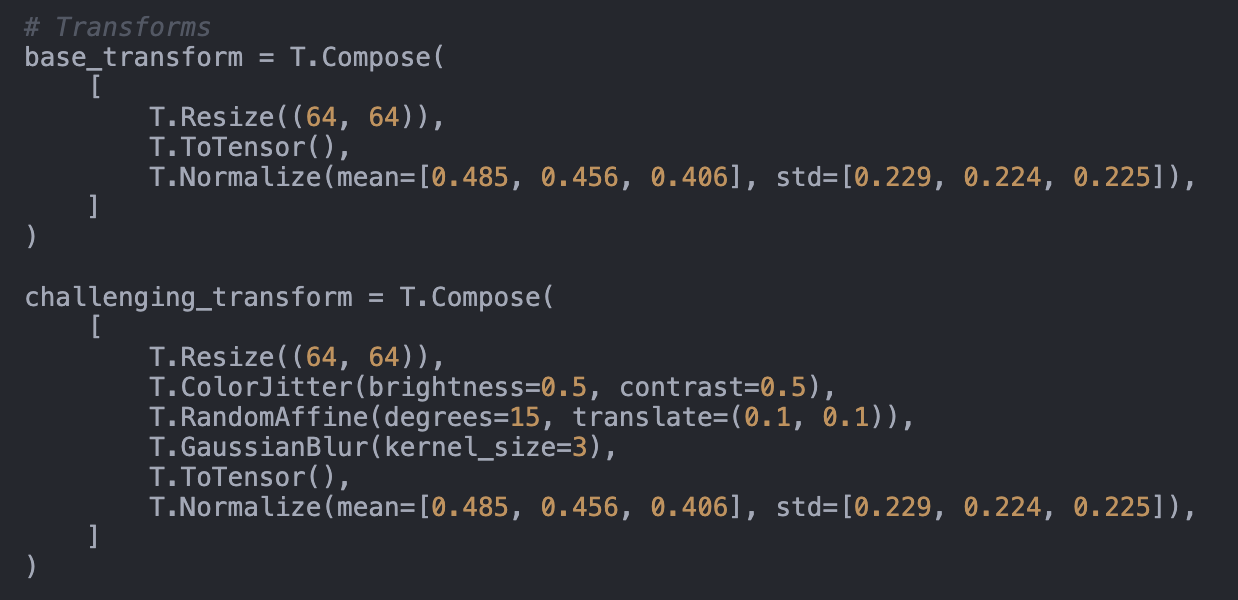}
\caption{Environment noise simulation implementation.}
\label{fig:env_noise_simulation}
\end{figure}

The paper states, ``To simulate challenging environmental conditions, we applied data augmentations during testing, including brightness and contrast adjustments, Gaussian blur, and random affine transformations.'' 
This is confirmed in the code, as shown in \Cref{fig:env_noise_simulation}.

\begin{figure}[h!]
\centering
\includegraphics[width=0.75\textwidth]{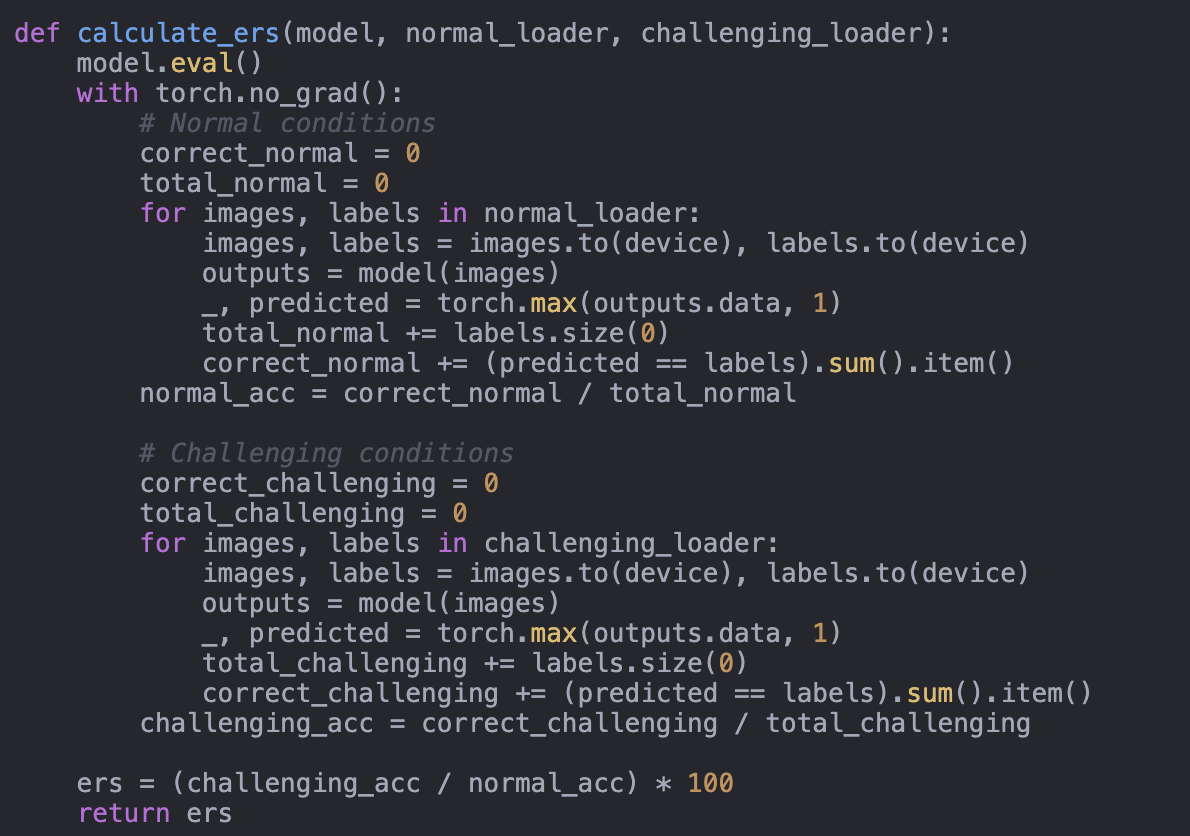}
\caption{Environmental Robustness Score calculation.}
\label{fig:ers_calc}
\end{figure}

The calculation of the Environmental Robustness Score--a metric introduced by the AI Scientist and defined as ``the ratio of model accuracy under challenging conditions to that under normal conditions, to quantify robustness''--matches the description in the paper, as shown in \Cref{fig:ers_calc}.

\newpage
\paragraph{Workshop Reviews}
\begin{tcolorbox}[
  breakable,
  colback  = green!5!white,
  colframe = green!75!black,
  title    = {\small Reviewer \#1: Review of Real-World Challenges in Pest Detection Using Deep Learning: An Investigation into Failures and Solutions.},
  fontupper=\scriptsize\ttfamily,
]
\begin{Verbatim}
Summary
This paper studies deep learning methods in pest detection applications. It highlights the need for more research and attempts to perform first experiments in this area.

Results
The paper performs experiments on an image classifier, a ResNet-18 trained on ImageNet, fine-tuned on the Crop Pest and Disease dataset. It uses various data augmentations to emulate the real-world conditions and further trains the model using "multi-dataset training". The model performance is measured in accuracy, loss, and Environmental Robustness Score (ERS). The first experiment investigates the effect tuning the learning rate has on training. The paper claims that a lower learning rate leads to a higher generalisation. The second experiment investigates the model's generalisation across different datasets. The paper claims that training the model on EuroSAT and CIFAR-10 datasets leads to better generalisation.

Strengths
The paper's background, motivation, and related work are well written. The motivation to study generalisation of deep learning methods in real-world agricultural applications is good.

Weaknesses
- The experiments are unmotivated and unclear.
- It is unclear how the choice of augmentation methods are related to "real-world environmental variability."
- The choice of ERS is unmotivated and no intuition on the score is given in the paper.
- The learning rate experiment conclusion that the model's generalization and robustness to real-world setting seems misleading since only 5 learning rates are used and the models are only trained for 10 epochs. Furthermore, the model was not tested in a real-world deployment. It is unclear what the paper means by "multi-dataset training" especially since the datasets have a different number of classes. Thus, the results of this experiment are unclear.
- The paper claims to have studied "deploying deep learning models for pest detection in real-world agricultural settings", however, the paper does not test the trained model in a real-world setting. Thus, the paper's conclusions are misleading.

General Remarks
- In the introduction, the difference between "controlled environment" and "real-world agricultural settings" should be explained since the experiments are not performed in a real-world deployment.
- Plots are small and difficult to read. Increasing the font size would help as well.
- In Figure 1, it would be helpful if the Train and Validation lines for the same learning rate were the same colour.
- In Figure 1, it is unclear whether the ERS scores are evaluated on the train or validation dataset.
- Unclear why the results in Figure 4 are pushed to the appendix and not combined with Figure 2 similar to Figure 1.
- References to the datasets are missing, e.g., the Crop Pest and Disease dataset.

Concluding Remarks
Overall, the paper addresses an interesting real-world problem. However, the lack of detail in the experiment section limits the strength of the conclusions, as the experimental results are not sufficiently supported by evidence. In its current state, it is of the reviewer's opinion that the paper does not meet the standard for publication. The authors are encouraged to further develop this research and consider deploying the model in a real-world setting to strengthen the validity of their findings.

Rating: 3: Clear rejection
Award: No Award
Confidence: 4: The reviewer is confident but not absolutely certain that the evaluation is correct
\end{Verbatim}
\end{tcolorbox}

\begin{tcolorbox}[
  breakable,
  colback  = green!5!white,
  colframe = green!75!black,
  title    = {\small Reviewer \#2: Review "REAL-WORLD CHALLENGES IN PEST DETECTION USING DEEP LEARNING: AN INVESTIGATION INTO FAILURES AND SOLUTIONS"},
  fontupper=\scriptsize\ttfamily,
]
\begin{Verbatim}
Summary: The authors investigate deep learning models in the context of pest detection. They state that common deep learning models work well in theoretical settings but struggle to generalize when being exposed to environmental changes. In addition, they offer potential approaches to address these issues and increase robustness of deep learning models in pest detection.

Scientific Rigor and Transparency: Yes, the authors conducted several experiments to underline their findings.

Novelty and Significance: Yes, the paper highlights the weaknesses of deep learning models in pest detection by analyzing how multi-dataset training and hyperparameter tuning affect their performance and ability to generalize.

Clarity of Writing: Yes, the paper is generally well-structured and written in a nice way. However, the paper could still improve on clarity by adding the missing references marked with a (?) in Section 3.

Alignment with Workshop Topics: Yes, the paper aligns with the theme of the workshop.

Additional Comments: The submitted paper provides insights into the weaknesses of deep learning models in real-world pest detection scenarios. It proposes strategies to mitigate these issues through hyperparameter tuning and multi-dataset training. While these methods can enhance model performance in practical applications, challenges persist due to environmental variability and domain discrepancies. I recommend that the authors include additional references in the field of pest detection, such as "Crop Pest Recognition in Real Agricultural Environments Using Convolutional Neural Networks with a Parallel Attention Mechanism" by Zhao et al., to offer a more comprehensive perspective on the topic.

Rating: 7: Good paper, accept
Award: No Award
Confidence: 3: The reviewer is fairly confident that the evaluation is correct
\end{Verbatim}
\end{tcolorbox}

\begin{tcolorbox}[
  breakable,
  colback  = green!5!white,
  colframe = green!75!black,
  title    = {\small Reviewer \#3: Critical Review of Real-World Challenges in Pest Detection Using Deep Learning: Methodological and Theoretical Considerations},
  fontupper=\scriptsize\ttfamily,
]
\begin{Verbatim}
The presented paper discusses challenges in pest detection based on digital images using the ResNet-18 model. The authors discuss experiments to evaluate the variability of classification performance based on simulated environmental changes. This topic is relevant, given major challenges such as biodiversity loss. Furthermore, the importance of interdisciplinary research (in this case, data science and biology) will increase, and such studies will help accelerate the use of machine learning in life sciences. However, I found shortcomings in this study, which I summarize in the text below.

The introduction provides a good overview of why this work is important. However, the technical motivation is not clear. A clear motivation based on the theoretical aspects of 'generalization' (see [1,2]), as well as a clear statement including literature on challenges in 'AI' in agriculture, would have been necessary.

The methodology section should refer to the appendix for more details (there are important details in the appendix). Furthermore, dataset details (e.g., example images, how many disease-related images are there?) are missing. The hypothesis concerning the learning rate, as well as augmentation to simulate real environmental variability, is not well motivated. I believe this harsh simulation of real-life dynamics should have been introduced in the abstract and introduction (it would still be an interesting study!). The motivation and derivation of the ERS are missing (is it an ad hoc approach?). The metric is prone to over/underestimating robustness due to unbalanced datasets (see the equation, and using the definition of accuracy, the size of the datasets influences the fraction). Based on the missing training/test/validation details above, it is unclear whether bias is introduced. Furthermore, I do not think that such studies must rely on the newest models. However, ResNet-18 is a rather old model, and no justification for selecting this model is given. A comparison to transformer-based architectures would have been interesting.

Finally, there are some language issues and BibTeX errors (see '?'). The figures should be updated to increase readability. Considering my discussion above, I think the results are still interesting. However, I do believe that the presentation must be adapted. I recommend a major revision, including a solid theoretical foundation, a presentation of the evaluation strategy using augmentation throughout the manuscript, and a comparison to recent deep learning models. Furthermore, I recommend switching from augmentation to real datasets or generative models. With these improvements, the impact of this study would be increased significantly.

[1] Wolpert, D.H. (2002). The Supervised Learning No-Free-Lunch Theorems. In: Soft Computing and Industry. Springer, London. https://doi.org/10.1007/978-1-4471-0123-9_3

[2] Goldblum, M. et al. (2024). Position: The No Free Lunch Theorem, Kolmogorov Complexity, and the Role of Inductive Biases in Machine Learning. Proceedings of the 41st International Conference on Machine Learning

Rating: 4: Ok but not good enough - rejection
Award: No Award
Confidence: 4: The reviewer is confident but not absolutely certain that the evaluation is correct
\end{Verbatim}
\end{tcolorbox}

\end{appendices}

\hypersetup{pageanchor=true}

\clearpage
\bibliography{references}
\bibliographystyle{plainnat}

\end{document}